\def\tnsenonblind{1}
\def\tnsebuildfull{1}

\documentclass[journal,comsoc]{IEEEtran}
\usepackage{amsmath,amsfonts,amssymb,amsthm}
\usepackage{thmtools}
\usepackage{algorithm}
\usepackage[commentColor=CommentDarkGray, italicComments=true, rightComments=true, beginComment=$\triangleright$~]{algpseudocodex}
\usepackage{array}
\usepackage[caption=false,font=normalsize,labelfont=sf,textfont=sf]{subfig}
\usepackage{textcomp}
\usepackage{stfloats}
\usepackage{url}
\usepackage{verbatim}
\usepackage{graphicx}
\usepackage{enumitem}
\usepackage{cite}
\usepackage{bm}
\usepackage[dvipsnames, x11names]{xcolor}
\usepackage{booktabs}
\usepackage{multirow}
\usepackage{multicol}
\usepackage{makecell}
\usepackage{longtable}
\usepackage{siunitx}
\allowdisplaybreaks
\newif\iftnseblindreview
\newif\iftnseincludesupplement
\tnseblindreviewfalse
\tnseincludesupplementfalse
\ifdefined\tnseanonymous
  \tnseblindreviewtrue
\fi
\ifdefined\tnsenonblind
  \tnseblindreviewfalse
\fi
\ifdefined\tnsebuildfull
  \tnseincludesupplementtrue
\fi

\definecolor{CommentDarkGray}{gray}{0.4}
\definecolor{CommentBlue}{RGB}{31,119,180}

\definecolor{StageOneFill}{HTML}{FEE0D2}   
\definecolor{StageOneDraw}{HTML}{DE2D26}   

\definecolor{StageTwoFill}{HTML}{E5F5E0}   
\definecolor{StageTwoDraw}{HTML}{31A354}   

\definecolor{StageThreeFill}{HTML}{DEEBF7}   
\definecolor{StageThreeDraw}{HTML}{3182BD}   

\definecolor{StageFiveFill}{HTML}{F5F5F5}   
\definecolor{StageFiveDraw}{HTML}{616161}   

\newtheorem{assumption}{Assumption}

\newtheorem{remark}{Remark}
\newtheorem{theorem}{Theorem}

\newenvironment{questionblock}{%
\noindent\textbf{Question: }\itshape
}{%
}

\begin{document}


\title{Communication-Efficient Federated Learning under Dynamic Device Arrival and Departure: Convergence Analysis and Algorithm Design}

\iftnseblindreview
\author{Anonymous Author(s)}
\else
\author{Zhan-Lun Chang,~\IEEEmembership{Student Member,~IEEE,} Dong-Jun Han,~\IEEEmembership{Member,~IEEE,} Seyyedali Hosseinalipour,~\IEEEmembership{Senior Member,~IEEE,} Mung Chiang,~\IEEEmembership{Fellow,~IEEE,} Christopher G. Brinton,~\IEEEmembership{Senior Member,~IEEE}
\thanks{Z.-L. Chang, M. Chiang, and C. G. Brinton are with the Elmore Family School of Electrical and Computer Engineering, Purdue University, West Lafayette, IN 47907, USA.}%
\thanks{D.-J. Han is with the Department of Computer Science and Engineering, Yonsei University, Seoul 03722, Republic of Korea.}%
\thanks{S. Hosseinalipour is with the Department of Electrical Engineering, University at Buffalo--SUNY, Buffalo, NY 14260, USA.}%
}
\fi



\maketitle

\begin{abstract}
Most federated learning (FL) approaches assume a fixed device set. However, real-world scenarios often involve devices dynamically joining or leaving the system, driven by, e.g., user mobility patterns or handovers across cell boundaries. This dynamic setting introduces unique challenges: (1) the optimization objective evolves with the active device set, unlike traditional FL's static objective; and (2) the current global model may no longer serve as an effective initialization for subsequent rounds, potentially hindering adaptation, delaying convergence, and reducing resource efficiency. To address these challenges, we first provide a convergence analysis for FL under a dynamic device set, accounting for factors such as gradient noise, local training iterations, and data heterogeneity in this practical setting. Motivated by this analysis, we propose a model initialization algorithm that enables rapid adaptation whenever devices join or leave the network. Our key idea is to compute a weighted average of previous global models, guided by gradient similarity, to prioritize models trained on data distributions that closely align with the current device set, thereby accelerating recovery from distribution shifts in fewer training rounds. This plug-and-play algorithm is designed to integrate seamlessly with existing FL methods, offering broad applicability. Experiments demonstrate that our approach achieves convergence speedups typically an order of magnitude or more compared to baselines, which we show drastically reduces energy consumption to reach a target accuracy.
\end{abstract}


\section{Introduction}
\IEEEPARstart{F}{ederated} learning (FL) facilitates collaborative model training across a set of devices connected to a central server. In FL, each device performs local training using its local data, and uploads model updates to the server periodically for aggregation. The aggregated models are then broadcast across the devices to initiate the next round. FL has found many applications in resource-constrained environments, including the Internet of Things (IoT)~\cite{wang2019adaptive, abdulrahman2023management}, autonomous driving~\cite{samarakoon2019distributed}, and smart cities~\cite{xing2021federated}.

In conventional FL frameworks~\cite{dinh2020federated,xiang2023federated, jhunjhunwala2022fedvarp, xu2020client}, the cohort of edge devices engaged in training is typically assumed to be static, implying the objective function is fixed. Nevertheless, in practice, the assumption of a static device set rarely holds in wireless edge networks, where user mobility and fluctuating coverage patterns drive continuous system dynamics. Devices frequently traverse cell boundaries, triggering handovers that result in their departure from one federated system and potential entry into another. Similarly, time-varying channel conditions and finite battery capacities can force devices to drop out of the system entirely. Unlike static scenarios where a fixed device set simply ``duty-cycles'' due to intermittent connectivity, these mobility-driven events fundamentally alter the active topology of the network: as devices physically enter or exit the server's coverage area, the system faces a structural evolution of the device set, necessitating a learning framework that can handle/track the available population in the network.

\begin{figure*}[t]
\vspace{-5.5mm}
    \centering
    \includegraphics[width=\linewidth]{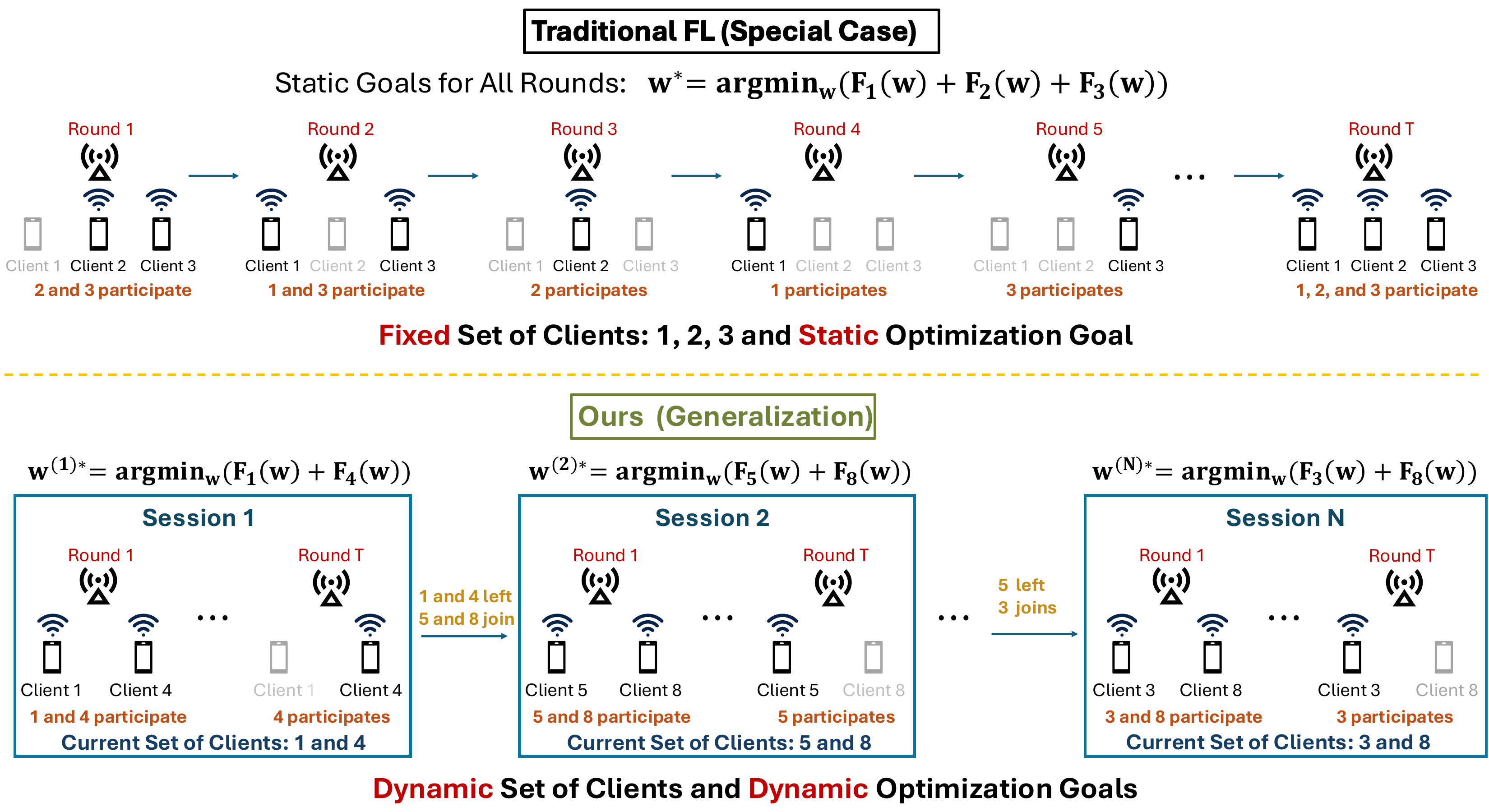}
    \vspace{-8.5mm}
    \caption{Comparison illustrating our proposed dynamic FL framework as a generalization of traditional FL. Traditional FL represents the single-session special case, limited to a fixed device set and a static optimization goal. Our framework generalizes this structure to accommodate dynamic device arrivals and departures across multiple sessions, allowing the optimization goal to evolve and minimize the loss for the currently active set of users.}
    \label{fig:system_model} 
    \vspace{-4.5mm}
\end{figure*}

\textbf{Illustrative Example:} Figure \ref{fig:system_model} illustrates our problem setting as a generalization of traditional FL to support dynamic device populations. Standard FL approaches typically handle temporary device unavailability by adjusting for partial participation, yet they assume the underlying set of users and the resulting optimization goal $\bm{w}^{*}$ remain fixed indefinitely. We generalize this by structuring model training into distinct \textit{sessions} to accommodate population changes in the system (traditional FL corresponds to one single session). As shown in the bottom panel of Figure \ref{fig:system_model}, we aim to retain the ability to handle partial device participation within a session while simultaneously adapting to device turnover between sessions. In doing so, \textit{the optimization goal will evolve from a single static target to a sequence of objectives specific to each session.} By carefully accounting for device arrivals and departures, our framework establishes a sequence of session-specific optimization targets; however, efficiently tracking these shifting goals is crucial to preventing model stagnation as well as minimizing time to (re-)convergence and the associated communication overhead. This leads to our key research question:

\begin{questionblock}
How can we design a plug-and-play method that (i) integrates with existing FL algorithms (e.g., FedProx \cite{li2020federated}, SCAFFOLD \cite{karimireddy2020scaffold}) originally developed for static device sets within a single session and (ii) provides fast adaptation to the changing optimization goal \(\bm{w}^{(s)\ast}\) as the system transitions between sessions \(s\)?
\end{questionblock}

Achieving this adaptation is non-trivial because the system operates blindly with respect to future device dynamics: since the server has no prior knowledge of which devices will join or leave the system, it cannot pre-calculate the optimal starting point for the next session. Furthermore, simply carrying over the global model from the previous session is often suboptimal as significant shifts in the underlying data distribution can render the old model stale, forcing the algorithm to waste communication rounds merely ``removing'' outdated information. In resource-constrained wireless networks, this slow re-convergence is unacceptable, as it directly translates to excessive latency and higher energy consumption from executing more training rounds. The goal, therefore, is enabling fast global model re-convergence to reduce the resource overhead of each session.

\textbf{Difference from Continual Learning:}   
 Unlike existing continual learning paradigms that aim to prevent catastrophic forgetting \cite{yoon2021federated}, which we consider in Section \ref{sec:experiments} as one of our baselines, our work focuses on enabling \textit{fast adaptation} to the current data distribution. This resembles a service provider aiming to build a global model that effectively serves the users who are currently active, motivated by wireless edge scenarios where devices frequently join and leave a local network. From this perspective, once a device leaves, the priority shifts to optimizing performance for the remaining users rather than preserving accuracy for those who have departed. We thus adopt a dynamic, session-centric view, emphasizing responsiveness to the active device population. 
\subsection*{Summary of Contributions}
To address the challenges introduced by dynamic device arrival and departure, this work makes the following key contributions:
\begin{itemize}[leftmargin=0.35cm]
    \item We conduct a convergence analysis for dynamic FL with session-specific objectives induced by device arrivals and departures. Unlike standard fixed-objective FL analyses, our bound averages the stationarity measure over a sequence of changing objectives \(F^{(s)}\) and makes explicit how the session-boundary initialization gap affects the descent term. The remaining terms recover standard FL effects, including mini-batch sampling noise, local-update drift, and inter-device data heterogeneity.
    \item Motivated by the convergence analysis, we propose a plug-and-play algorithm for dynamic optimization in FL, which constructs an effective initial model for each session. Aiming for low-complexity integration with existing FL methods (e.g., MOON \cite{li2021modelmoon}, FedProx \cite{li2020federated}, SCAFFOLD \cite{karimireddy2020scaffold}, FedACG \cite{kim2024communication}), our algorithm initializes the model as a gradient similarity-weighted average of previous models, prioritizing models trained on devices whose data characteristics closely align with those of the current session. This initialization strategy accelerates model adaptation and mitigates costly performance drops caused by dynamic device participation.
    \item  We evaluate system performance using two distinct metrics: \textit{Time-to-Accuracy} and \textit{Accumulated Accuracy Gain}. \textit{Time-to-Accuracy} measures the number of global rounds required to recover a specific target accuracy, which translates to adaptation latency and energy consumption as well. Our results show that our strategy reduces the rounds needed to reach a target accuracy typically by an order of magnitude or more compared to continual learning baselines. Through communication and computation modeling, we quantify the corresponding improvements in energy consumption as well. Complementing this, the \textit{Accumulated Accuracy Gain} metric confirms superior sustained performance throughout the session, yielding total accumulated accuracy gains of up to 8x in challenging zero-overlap session scenarios.
\end{itemize}

\vspace{-2mm}
\section{Related Works}

FL has emerged as a prominent paradigm for distributed learning, enabling the collaborative training of models across devices while preserving data privacy \cite{mcmahan2017communication, li2020federated, nguyen2020fast}. One of the foundational works introduced the Federated Averaging (FedAvg) algorithm, which remains a cornerstone of many FL systems \cite{mcmahan2017communication}. Subsequent studies have explored various aspects of FL, including enhancing resource efficiency \cite{wang2021quantized, xing2021federated, luo2021cost, xu2023federated, chen2020joint}, building robustness to adversarial attacks \cite{wei2021low, yang2025communication}, modeling energy consumption \cite{yang2020energy}, and more distributed (non-star) topologies \cite{lin2021semi, xing2021federated, lim2021dynamic, luo2020hfel, yang2023decentralized, ye2022decentralized, chen2022decentralized, zhou2023decentralized}. However, these approaches typically assume a fixed set of participating devices throughout the training process, where devices are expected to remain available for the entire training duration \cite{li2020federated}. This assumption simplifies the modeling of convergence and performance but does not adequately capture real-world scenarios characterized by dynamic device participation patterns. 

To address the limitations of static device sets in FL, research on dynamic device selection and flexible participation has gained momentum, particularly in response to the challenges posed by varying device availability \cite{xu2020client, jhunjhunwala2022fedvarp, sun2021dynamic, ribero2022federated}. These studies explore strategies such as optimizing device selection based on resource constraints, modeling participation patterns probabilistically, and employing adaptive algorithms to address the effects of non-IID (independent and identically distributed) data. In essence, the goal of these studies is to enhance overall model performance by strategically managing device participation during training, balancing computational efficiency, communication costs, and data representativeness. However, these works often assume a fixed device set, neglecting the dynamic nature of device arrival and departure. While \cite{ruan2021towards} propose a flexible FL framework that allows for inactive devices, incomplete model updates, or dynamic device participation, their analysis is limited to scenarios where the optimization goal changes only once, specifically when a single device joins during training. This restricts the applicability of their approach to more complex and realistic patterns of device availability. Also, their analysis relies on more restrictive assumptions such as strong convexity and universally bounded gradient.  There is also a line of work in continual FL \cite{dong2022federated}, where the goal is to build a model that preserves all previously learned knowledge by preventing catastrophic forgetting. In contrast, as previously mentioned, our work focuses on rapidly adapting the global model to the current data distribution of devices. Given this difference in objectives, our approach significantly outperforms continual FL in scenarios involving dynamic device arrivals and departures, as we will show in Section \ref{sec:experiments}.

\vspace{-2mm}
\section{System Model and Theoretical Analysis} 

In this section, we present our system model for dynamic FL (Sec.~\ref{sec:sysMod}) and conduct our convergence analysis (Sec.~\ref{section:theoretical_analysis}).

\subsection{System Model}
\label{sec:sysMod}
We discretize the FL process into a number of sessions, indexed by \( s \in \{1, \dots, S\} \). Each session consists of \( T \) global training rounds, indexed by \( t \in \{1, \dots, T\} \).  
To model device dynamics, \( \mathcal{K}^{(s)} = \{1, \dots, K^{(s)}\} \) denotes the population of devices available in session \( s \). Each device is connected to a central server and maintains its own ML model. Specifically, at session \( s \), each device \( k \in \mathcal{K}^{(s)} \) holds a dataset \( \mathcal{D}_k^{(s)} \) of size \( D_k^{(s)} \triangleq |\mathcal{D}_k^{(s)}| \), where each sample \( d \in \mathcal{D}_k^{(s)} \) consists of a feature-label pair. A loss function \( f(\bm{w}, d) \) evaluates model parameters \( \bm{w} \in \mathbb{R}^M \) over individual samples. The local loss at device \( k \) during session \( s \) is defined as follows:
\begin{equation}
    F_k^{(s)}(\bm{w}, \mathcal{D}_k^{(s)}) \triangleq  \sum_{d \in \mathcal{D}_k^{(s)}} f(\bm{w}, d)/D_k^{(s)},
\end{equation}
and the global loss function aggregates local losses over the current population:
\begin{equation}
    F^{(s)}(\bm{w}, \mathcal{D}^{(s)}) \triangleq \frac{1}{D^{(s)}} \sum_{k \in \mathcal{K}^{(s)}} D_k^{(s)} F_k^{(s)}(\bm{w}, \mathcal{D}_k^{(s)}),
\end{equation}
where \( D^{(s)} \triangleq |\mathcal{D}^{(s)}| \) denotes the total number of data samples across all available devices in session $s$.

In our scenario of interest, where devices may join or leave at any time, the FL training objective evolves per session. As a result, the reference stationary global models form a sequence \( \{\bm{w}^{(s)\ast}\}_{s=1}^{S} \), with each \( \bm{w}^{(s)\ast} \) corresponding to the current session's data:
\begin{equation}
\begin{aligned}
\bm{w}^{(s)\ast} &= \arg\min_{\bm{w} \in \mathbb{R}^{M}} F^{(s)}(\bm{w}, \mathcal{D}^{(s)}).
\end{aligned}
\end{equation}
This formulation is different from the traditional FL objectives (i.e., $\bm{w}^{\ast} = \arg\min_{\bm{w} \in \mathbb{R}^{M}} F(\bm{w}, \mathcal{D})$), which seeks a single model \( \bm{w}^* \) optimized over a fixed dataset \( \mathcal{D} \). In contrast, our formulation targets a time-varying objective, optimizing \( \bm{w}^{(s)\ast} \) based on the current/evolving data \( \mathcal{D}^{(s)} \). Since the considered losses are generally nonconvex, global optimality is not guaranteed; thus, \( \bm{w}^{\ast} \) and \( \bm{w}^{(s)\ast} \) are used as reference points for stationary solutions satisfying first-order optimality conditions (e.g., small gradient norm).

In our setting, each training round \(t\) in session \(s\) proceeds as follows. Devices autonomously join/leave over time, and the server only schedules a subset of the currently available devices as participants, denoted by \( \mathcal{A}^{(s,t)} \subseteq \mathcal{K}^{(s)} \). The server broadcasts the global model \( \bm{w}^{(s,t)} \in \mathbb{R}^{M} \) to them. Each participating device \( k \) then updates the model using its data \( \mathcal{D}_k^{(s)} \) over \( e_k^{(s,t)} \) steps of stochastic gradient descent (SGD). Specifically, across SGD iterations \( h \in \{0, \dots, e_k^{(s,t)} - 1\} \), the model update rule is as follows:
\begin{equation}
    \bm{w}_k^{(s,t), h+1} = \bm{w}_k^{(s,t), h} - \eta^{(s,t)} \nabla F_k^{(s)}(\bm{w}_k^{(s,t), h}, \mathcal{B}_k^{(s,t)}),
\end{equation}
where \( \eta^{(s,t)} \) is the learning rate and \( \mathcal{B}_k^{(s,t)} \subset \mathcal{D}_k^{(s)} \) is a mini-batch. The local training starts from the global model \( \bm{w}_k^{(s,t), 0} = \bm{w}^{(s,t)} \) and results in a final local model \( \bm{w}_k^{(s,t), \mathsf{F}} \), where \( \mathsf{F} \triangleq e_k^{(s,t)} \) denotes the final local SGD step index. Devices send their final models to the server, which aggregates them as follows:
\begin{equation}
    \bm{w}^{(s,t+1)} = \sum_{k \in \mathcal{A}^{(s,t)}} \frac{D_k^{(s)}}{\sum_{j \in \mathcal{A}^{(s,t)}} D_j^{(s)}} \bm{w}_k^{(s,t), \mathsf{F}}.
\end{equation}
The updated global model \( \bm{w}^{(s,t+1)} \) is then broadcast to devices for the next training round.
For notational convenience, when there is no ambiguity, we omit the second argument (dataset) and write \(F_k^{(s)}(\bm{w})\) and \(F^{(s)}(\bm{w})\).

\subsection{Theoretical Analysis} \label{section:theoretical_analysis}

To gain insights into how the FL algorithm performs in the complex system we consider, we first analyze its performance, which inspires the design of our proposed algorithm. We leverage the following assumptions \cite{wang2020tackling} to derive the analysis (henceforth, $\| \cdot \|$ denotes the two-norm).

\begin{assumption}[Smoothness of Loss Functions] \label{assumption:smoothness}
Local loss function \(F_k^{(s)}\) is \(\beta\)-smooth for all \(k \in \mathcal{K}^{(s)}\), \(\forall \bm{w}, \bm{w}' \in \mathbb{R}^M\), and \(s\):
\begin{equation}
\|\nabla F_k^{(s)}(\bm{w}) - \nabla F_k^{(s)}(\bm{w}')\| \leq \beta \|\bm{w} - \bm{w}'\|,
\end{equation}
which implies the \(\beta\)-smoothness of the global loss function \(F^{(s)}\) for all \(s \in \{1, \dots, S\}\).
\end{assumption}

\begin{assumption}[Bounded Dissimilarity of Local Loss Functions] \label{assumption:bounded_dissimilarity}
There exist finite constants \(\zeta_1 \geq 1\), \(\zeta_2 \geq 0\), such that for any set of coefficients \(\{a_k \geq 0\}_{k \in \mathcal{K}^{(s)}}\) with \(\sum_{k \in \mathcal{K}^{(s)}} a_k = 1\), the following holds for all \(s\) and \(\bm{w}\):
\begin{equation}
\sum_{k \in \mathcal{K}^{(s)}} a_k \|\nabla F_k^{(s)}(\bm{w})\|^2
\leq \zeta_1 \big\| \sum_{k \in \mathcal{K}^{(s)}} a_k \nabla F_k^{(s)}(\bm{w}) \big\|^2 + \zeta_2.
\end{equation}
\end{assumption}

\begin{assumption}[Bounded Iterates] \label{assumption:bounded_iterates}
There exists a compact set \(\mathcal{W}\subset\mathbb{R}^M\) such that all global and local iterates generated by the algorithm satisfy \(\bm{w}^{(s,t)}\in\mathcal{W}\) and \(\bm{w}_k^{(s,t),h}\in\mathcal{W}\) for all \(s,t,k,h\).
\end{assumption}

This bounded-domain restriction can be enforced, for example, by projection, and is standard in nonconvex stochastic optimization when Lipschitz-type conditions are only needed over the parameter region visited by the algorithm rather than globally over \(\mathbb{R}^M\).

To capture the heterogeneity within the local datasets of devices participating in a specific round, we assume bounded local data variability. For clarity, we write each sample as \(d=(\bm{x},y)\), where \(\bm{x}\) is the feature vector and \(y\) is the label. The data-distance notation \(\|d-d'\|\) refers only to the feature-vector distance. That is, for \(d=(\bm{x},y)\) and \(d'=(\bm{x}',y')\), we define \(\|d-d'\|\triangleq\|\bm{x}-\bm{x}'\|_2\). Also, we denote the gradient with respect to the model parameter \(\bm{w}\) as \(\nabla_{\bm{w}} f(\bm{w}, d)\).
\begin{assumption}[Bounded Local Data Variability] \label{assumption:local_data_variability}
The local data variability at each device \(k\) is measured via \(\Theta_k^{(s,t)} \geq 0\), which bounds the sensitivity of the sample-wise gradient with respect to changes in local data over the compact iterate region \(\mathcal{W}\).  For all \(\bm{w}\in\mathcal{W}\), \(k\), and \(d,d' \in \mathcal{D}_k^{(s)}\),
\begin{equation}
    \|\nabla_{\bm{w}} f(\bm{w}, d) - \nabla_{\bm{w}} f(\bm{w}, d')\| \leq \Theta_k^{(s,t)} \|d - d'\|.
\end{equation}
We further define \(\Theta^{(s,t)} = \max_{k \in \mathcal{K}^{(s)}} \{\Theta_k^{(s,t)}\}\). Intuitively, a larger \(\Theta_k^{(s,t)}\) means that different feature vectors on device \(k\) can induce more diverse gradient directions, which increases the mini-batch sampling noise. Thus, \(\Theta^{(s,t)}\) is not a tunable algorithm parameter; it is an analysis constant determined by the loss function, the bounded iterate region \(\mathcal{W}\), and the local feature domain. It converts feature dispersion into gradient dispersion. When the dependence on the sample is not central to the discussion, we write the loss or gradient only as a function of \(\bm{w}\).
\end{assumption}

\begin{assumption}[Uniform Mini-Batch Sampling] \label{assumption:uniform_minibatch}
At each local SGD step, conditioned on the current local model, device \(k\) samples the mini-batch \(\mathcal{B}_k^{(s,t),h}\) uniformly without replacement from \(\mathcal{D}_k^{(s)}\). We assume \(D_k^{(s)}\ge 2\) and \(1\le B_k^{(s,t)}\le D_k^{(s)}\) for all devices, sessions, and rounds, so that the finite-population variance and correction terms are well defined. Therefore, the mini-batch gradient is an unbiased estimator of the local full-batch gradient:
\begin{equation}
\mathbb{E}_{\mathcal{B}}\left[
\frac{1}{B_k^{(s,t)}}\sum_{d \in \mathcal{B}_k^{(s,t),h}}
\nabla_{\bm{w}} f(\bm{w},d)
\right]
= \nabla F_k^{(s)}(\bm{w}).
\end{equation}

\end{assumption}

For each device \(k\), we  write its local dataset as \(\mathcal{D}_k^{(s)}=\{(\bm{x}_{k,i}^{(s)},y_{k,i}^{(s)})\}_{i=1}^{D_k^{(s)}}\). Let
\[
\bm{\lambda}_k^{(s)}
\triangleq
\frac{1}{D_k^{(s)}}\sum_{i=1}^{D_k^{(s)}}\bm{x}_{k,i}^{(s)}
\]

denote the empirical mean of its local feature vectors. We define
\[
(\widetilde{\sigma}_k^{(s,t)})^2
\triangleq
\sum_{i=1}^{D_k^{(s)}}\|\bm{x}_{k,i}^{(s)}-\bm{\lambda}_k^{(s)}\|^2
\]

as the local feature dispersion. This quantity is independent of the model parameter and captures how spread out the feature vectors on device \(k\) are. The superscript \((s,t)\) is kept for consistency with the round-dependent convergence bound, although the underlying dataset is fixed within a session.

Assumption \ref{assumption:uniform_minibatch} supplies the usual stochastic-gradient unbiasedness property. It is worth noting we do not impose a separate generic bounded stochastic-gradient variance assumption. Instead, the finite-population variance induced by uniform sampling without replacement is bounded in the proof using the bounded local data variability in Assumption \ref{assumption:local_data_variability} and the local data dispersion \((\widetilde{\sigma}_k^{(s,t)})^2\).

We next obtain the convergence bound for the nonconvex global loss function in our dynamic system of interest. 

\begin{theorem}[Upper Bound on the Gradient Norm] \label{theorem:upper_bound}
Suppose each device performs local SGD updates and the server aggregates via weighted averaging. The bound is conditional on the session initial model \(\bm{w}^{(s,1)}\); For the multi-local-step case \(e^{(s,t)}\ge 2\), assume the learning rate \(\eta^{(s,t)}\) satisfies
{\small
\begin{align*}
\eta^{(s,t)} \leq
\frac{\sqrt{\Lambda^{(s,t)}}}{2\beta\sqrt{(\zeta_1+\Lambda^{(s,t)})e^{(s,t)}(e^{(s,t)}-1)}},
\end{align*}
}
$\!\!\!$ where \(\Lambda^{(s,t)}\in(0,1)\) is an analysis slack parameter used to keep the coefficient of \(\|\nabla F^{(s)}(\bm{w}^{(s,t)})\|^2\) strictly negative after bounding the local-update drift terms. It is not used by the algorithm; it only parameterizes the sufficient learning-rate condition. Then, for any pattern of device arrivals and departures, under full participation of the currently available devices in each round and for any bounded session initializations, the cumulative average of the gradient of the global loss over \(S\) sessions and \(T\) rounds satisfies \eqref{upper_bound}.

\begin{table*}[!t]
  \centering
  \footnotesize
  \begin{minipage}{1.0\textwidth}
  \vspace{.2mm}
  \begin{equation} \label{upper_bound}
  \hspace{-15mm}
  \begin{aligned}
  & \frac{1}{ST} \sum_{s=1}^{S} \sum_{t=1}^{T} \mathbb{E}\|\nabla F^{(s)}(\bm{w}^{(s,t)})\|^2  \leq \underbrace{\frac{1}{ST} \sum_{s=1}^{S} \sum_{t=1}^{T} \frac{2\mathbb{E}\big[F^{(s)}(\bm{w}^{(s,t)}) - F^{(s)}(\bm{w}^{(s,t+1)})\big]}{\eta^{(s,t)} (1 - \Lambda^{(s,t)})}}_{\text{(a)}} + \underbrace{\frac{1}{ST} \sum_{s=1}^{S} \sum_{t=1}^{T} \frac{4 \beta (\Theta^{(s,t)})^{2} \eta^{(s,t)}e^{(s,t)}}{1 - \Lambda^{(s,t)}} \sum_{k \in \mathcal{K}^{(s)}} \frac{(\widetilde{\sigma}_k^{(s,t)})^2}{B_k^{(s,t)}} \frac{D_k^{(s)} - B_k^{(s,t)}}{(D_k^{(s)})^2}}_{\text{(b)}}  \\
  & + \underbrace{\frac{1}{ST} \Bigg(\sum_{s=1}^{S} \sum_{t=1}^{T} \bigg(\frac{4 \beta^2 (\eta^{(s,t)})^2(e^{(s,t)} - 1)}{D^{(s)} (1 - \Lambda^{(s,t)})} \sum_{k \in \mathcal{K}^{(s)}} \left(1 - \frac{B_k^{(s,t)}}{D_k^{(s)}}\right) \frac{(\Theta^{(s,t)})^2 (\widetilde{\sigma}_k^{(s,t)})^2 }{B_k^{(s,t)}}\bigg) \Bigg)}_{\text{(c)}}  + \underbrace{\frac{1}{ST} \sum_{s=1}^{S} \sum_{t=1}^{T} 8 \beta^2 (\eta^{(s,t)})^2 e^{(s,t)}(e^{(s,t)}-1) \frac{\zeta_2}{1 - \Lambda^{(s,t)}}}_{\text{(d)}}.
  \end{aligned}\hspace{-14mm}
  \end{equation}
  \end{minipage}
  \hrule
  \end{table*}
\end{theorem}

\begin{proof}
We present a sketch proof here and provide the detailed proof in our online technical report \cite{chang2025federatedlearningdynamicclient}. The convergence analysis relies on the $\beta$-smoothness of the global loss function $F^{(s)}(\cdot)$ and the bounding of error terms introduced by stochastic sampling and local updates as follows:

\textit{Step 1. Smoothness and Error Decomposition:}
Let $\bm{w}^{(s,t)}$ be the global model at round $t$ of session $s$. Using the smoothness assumption, the expected descent in one global round is bounded by:
\begin{align}
    \mathbb{E} &[ F^{(s)}(\bm{w}^{(s,t+1)}) ] \leq F^{(s)}(\bm{w}^{(s,t)}) \nonumber \\
    &- \eta^{(s,t)} \mathbb{E} \left\langle \nabla F^{(s)}(\bm{w}^{(s,t)}), \bm{\Delta}^{(s,t)} \right\rangle + \frac{\beta (\eta^{(s,t)})^2}{2} \mathbb{E} \| \bm{\Delta}^{(s,t)} \|^2,
\end{align}
where under the full-participation theorem setting, $\bm{\Delta}^{(s,t)} = \sum_{k \in \mathcal{K}^{(s)}} \frac{D_k^{(s)}}{D^{(s)}} \nabla \overline{F}_k^{(s)}$ is the aggregated update direction derived from the currently available device set \(\mathcal{K}^{(s)}\). Using the algebraic identity $2\langle \bm{a}, \bm{b} \rangle = \|\bm{a}\|^2 + \|\bm{b}\|^2 - \|\bm{a} - \bm{b}\|^2$, we decompose the inner product to isolate the ``true'' gradient norm. This yields:
\begin{align} \label{eq:sketch_decomp}
    &\mathbb{E} [ F^{(s)}(\bm{w}^{(s,t+1)}) ] \leq F^{(s)}(\bm{w}^{(s,t)}) - \frac{\eta^{(s,t)}}{2} \|\nabla F^{(s)}(\bm{w}^{(s,t)})\|^2 \nonumber \\
    &+ \underbrace{\frac{\beta (\eta^{(s,t)})^2}{2} \mathbb{E} \| \bm{\Delta}^{(s,t)} \|^2}_{\text{(A) Update Second Moment}}
    + \underbrace{\frac{\eta^{(s,t)}}{2} \mathbb{E} \| \nabla F^{(s)}(\bm{w}^{(s,t)}) - \bm{\Delta}^{(s,t)} \|^2}_{\text{(B) Gradient Mismatch}}.
\end{align}
Equation \eqref{eq:sketch_decomp} highlights two error sources opposing convergence: the second moment of the aggregated stochastic update direction (Term A), whose bound contains the mini-batch sampling variance, and the mismatch between the true global gradient and the accumulated local update direction (Term B), which captures local-update drift.

\textit{Term A in \eqref{eq:sketch_decomp}:}
By Assumption \ref{assumption:uniform_minibatch}, the mini-batch gradient is unbiased. Unlike standard SGD analysis, our sampling is performed without replacement. Utilizing the properties of the hypergeometric distribution (finite population correction), we bound the variance of the mini-batch gradients directly rather than assuming a generic bounded stochastic-gradient variance. Specifically, we show that Term A has an upper bound:
\begin{align} \label{eq:sketch_variance}
     \beta (\eta^{(s,t)})^2 (\Theta^{(s,t)})^2 e^{(s,t)} \sum_{k \in \mathcal{K}^{(s)}} \frac{(\widetilde{\sigma}_k^{(s,t)})^2}{D_k^{(s)}} \frac{1 - \frac{B_k^{(s,t)}}{D_k^{(s)}}}{B_k^{(s,t)}}.
\end{align}
Crucially, the factor $(1 - B_k^{(s,t)}/D_k^{(s)})$ captures the variance reduction: as the local batch size $B_k^{(s,t)}$ approaches the local dataset size $D_k^{(s)}$, the variance contribution vanishes.

\textit{Term B in \eqref{eq:sketch_decomp}:}
Since devices perform the common number \(e^{(s,t)}\) of local steps in the theorem, the local model \(\bm{w}_k^{(s,t), e}\) drifts from \(\bm{w}^{(s,t)}\). We use an aggregate-first comparison between the global gradient and the weighted accumulated local direction, so the proof bounds only local-update drift and does not introduce a separate fixed-client heterogeneity term. The accumulated drift is bounded by:
\begin{align}
    \text{Term (B)} \leq \eta^{(s,t)} \mathcal{E}_{\text{var}} + \eta^{(s,t)} \mathcal{E}_{\text{grad}} \|\nabla F^{(s)}(\bm{w}^{(s,t)})\|^2,
\end{align}
where $\mathcal{E}_{\text{var}}$ represents drift caused by noise and $\mathcal{E}_{\text{grad}}$ represents drift caused by the gradient direction itself.

\textit{Step 4. Final Convergence Bound:}
Substituting the bounds for Terms (A) and (B) back into \eqref{eq:sketch_decomp}, we group the coefficients of $\|\nabla F^{(s)}(\bm{w}^{(s,t)})\|^2$. To guarantee convergence, we require the step size $\eta^{(s,t)}$ to be sufficiently small such that the net coefficient is negative:
\begin{equation}
    \eta^{(s,t)} \leq \frac{\sqrt{\Lambda^{(s,t)}}}{2 \beta \sqrt{\left( \zeta_1 + \Lambda^{(s,t)} \right) e^{(s,t)} \left(e^{(s,t)}-1\right)}},
\end{equation}
where \(\Lambda^{(s,t)}\in(0,1)\) is the proof slack parameter introduced in the theorem statement. Under this condition, telescoping the sum over $S$ sessions and $T$ rounds and rearranging yields the bound on $\frac{1}{ST} \sum_{s=1}^{S} \sum_{t=1}^{T} \|\nabla F^{(s)}(\bm{w}^{(s,t)})\|^2$ in Theorem \ref{theorem:upper_bound}.
\end{proof}

Examining~\eqref{upper_bound}, the theorem provides a nonconvex stationarity bound on the average gradient norm rather than a guarantee of exact convergence to a global minimizer. This is the standard convergence notion in nonconvex FL: the average stationarity measure becomes small when the descent term and the error terms are controlled. The bound is informative because it separates this stationarity measure into four interpretable terms. Term (a) is the normalized expected descent of the session-specific global loss. It decreases when the current initialization and subsequent updates produce larger loss reduction. Term (b) is the finite-population mini-batch sampling noise term. It depends on the correction \(1-B_k^{(s,t)}/D_k^{(s)}\), and therefore vanishes when full local batches are used. It also increases with the local data variability \(\Theta^{(s,t)}\), the local data dispersion \((\widetilde{\sigma}_k^{(s,t)})^2\), and the common number of local steps \(e^{(s,t)}\). This is because larger feature dispersion and larger sample-wise gradient sensitivity make mini-batch gradients more variable, while a larger \(e^{(s,t)}\) accumulates this sampling noise over more local SGD steps. Term (c) captures the interaction between mini-batch sampling noise and local model drift across multiple local SGD steps; it grows with \(e^{(s,t)}-1\), so it disappears when each device performs only one local step. Term (d) captures inter-device gradient dissimilarity through \(\zeta_2\) and also vanishes in the single-step case. Thus, larger mini-batches improve terms (b) and (c), more local steps can worsen the local-drift terms (c) and (d), larger data variability or local dispersion worsens the sampling-related terms, and larger inter-device heterogeneity worsens term (d). Finally, \(\Lambda^{(s,t)}\in(0,1)\) is only an analysis parameter: choosing a valid \(\Lambda^{(s,t)}\) and a learning rate satisfying the theorem's condition ensures that the descent term in the proof remains positive after accounting for local-drift errors. The factor \(1-\Lambda^{(s,t)}\) in the denominator of \eqref{upper_bound} reflects this reserved descent margin.

Theorem \ref{theorem:upper_bound} reduces to a standard nonconvex FedAvg-style full-participation bound when \(S=1\) and the device set is fixed, because the objective no longer changes across sessions. In that fixed-objective case, term (a) is the usual descent term, term (b) is the stochastic mini-batch sampling contribution with a finite-population correction, terms (c) and (d) are the standard local-update drift and heterogeneity contributions controlled by the number of local steps and bounded dissimilarity. What is specific to the dynamic-device setting is that the objective is session-dependent, \(F^{(s)}\), the bound averages gradients over changing objectives, and the quality of the session initial model, measured by the gap \(\mathbb{E}[F^{(s)}(\bm{w}^{(s,1)})-F^{(s)}(\bm{w}^{(s)\ast})]\), becomes meaningful because \(\bm{w}^{(s)\ast}\) changes with \(\mathcal{K}^{(s)}\). If \(S=1\), this dynamic-session motivation disappears and the result behaves like the classical fixed-objective FL case.

The common-local-step condition is used only for the convergence theorem to focus on dynamic device-set effects. If devices use arbitrary local step counts, the aggregate-first proof introduces an additional step-count imbalance term through \(\sum_k p_k^{(s,t)} e_k^{(s,t)}\nabla F_k^{(s)}(\bm{w}^{(s,t)})\), rather than \(e^{(s,t)}\sum_k p_k^{(s,t)}\nabla F_k^{(s)}(\bm{w}^{(s,t)})\). This term can be controlled by bounded dissimilarity, but it is orthogonal to the main contribution and is therefore left outside the main theorem.

\begin{remark}[Scope of Theorem~\ref{theorem:upper_bound}]
Theorem~\ref{theorem:upper_bound} is an analytical upper bound for the dynamic-device setting with session-specific objectives \(F^{(s)}\). It is proven under full participation of the currently available devices in each round, i.e., \(\mathcal{A}^{(s,t)}=\mathcal{K}^{(s)}\), though the broader system model and algorithm permit partial participation. This isolates the effects of changing device sets, mini-batch sampling, local updates, and data heterogeneity from the effects of partial client sampling; incorporating arbitrary client subsampling would add a separate client-sampling variance term. Further, the theorem holds for any bounded session initialization \(\bm{w}^{(s,1)}\) and treats this initialization as given, rather than analyzing the specific warm-start construction. Therefore, while term~(a) in~\eqref{upper_bound} will be useful in motivating our choice of design criterion in~\eqref{eq:implication_gap}, proving that Algorithm~\ref{alg:dynamic_initial_model_construction} satisfies this criterion would require a separate analysis of the warm-start rule itself. For the convergence result, all currently available devices are assumed to use the same number of local SGD steps in a round, i.e., \(e_k^{(s,t)}=e^{(s,t)}\) for all \(k\in\mathcal{K}^{(s)}\), which matches our experimental setting and isolates dynamic device-set effects from heterogeneous local computation budgets.
\end{remark}

\begin{remark}[Single-Local-Step Case]
When \(e^{(s,t)}=1\), the local-drift terms vanish and the second learning-rate restriction is unnecessary. In this single-local-step case, the same argument yields
\begin{align}
&\frac{1}{ST}\sum_{s=1}^{S}\sum_{t=1}^{T}
\mathbb{E}\|\nabla F^{(s)}(\bm{w}^{(s,t)})\|^2 \nonumber\\
&\leq
\frac{1}{ST}\sum_{s=1}^{S}\sum_{t=1}^{T}
\frac{2\mathbb{E}\big[F^{(s)}(\bm{w}^{(s,t)})-F^{(s)}(\bm{w}^{(s,t+1)})\big]}{\eta^{(s,t)}} \\
&+
\frac{1}{ST}\sum_{s=1}^{S}\sum_{t=1}^{T}
4\beta(\Theta^{(s,t)})^2\eta^{(s,t)}
\sum_{k\in\mathcal{K}^{(s)}}
\frac{(\widetilde{\sigma}_k^{(s,t)})^2}{B_k^{(s,t)}}
\frac{D_k^{(s)}-B_k^{(s,t)}}{(D_k^{(s)})^2}. \nonumber
\label{upper_bound_single_step}
\end{align}

\end{remark}

\textbf{Design Implications:} 
Term~(a) in \eqref{upper_bound} can be decomposed into two terms:
\(\mathbb{E}\bigl[F^{(s)}(\bm w^{(s,t)}) - F^{(s)}(\bm w^{(s)\ast})\bigr]\) and 
\(\mathbb{E}\bigl[F^{(s)}(\bm w^{(s)\ast}) - F^{(s)}(\bm w^{(s,t+1)})\bigr].
\)
The second term arises from the update following round~\(t\). At the beginning of a new session, the first term suggests an initialization-related design criterion: a smaller value of \(\mathbb{E}\bigl[F^{(s)}(\bm w^{(s,1)})-F^{(s)}(\bm w^{(s)\ast})\bigr]\) would reduce the term (a) in \eqref{upper_bound}, all else being equal. This observation motivates seeking an initial model \(\bm{w}_{\text{init}}^{(s,1)}\) satisfying
\begin{equation}\label{eq:implication_gap}
\begin{aligned}
&\mathbb{E}\bigl[F^{(s)}(\bm{w}_{\text{init}}^{(s,1)}) - F^{(s)}(\bm w^{(s)\ast})\bigr] \\
&\le\ \mathbb{E}\bigl[F^{(s)}(\bm w^{(s-1, T+1)}) - F^{(s)}(\bm w^{(s)\ast})\bigr],
\end{aligned}
\end{equation}
which implies that the chosen warm start should be closer to the new session optimum than simply carrying over the previous session's final model.
Motivated by this, our proposed initialization rule in Sec.~\ref{sec:algorithm} will form a warm start by weighting historical models according to the similarity between update directions estimated from a common reference model, aiming to achieve \eqref{eq:implication_gap}, a goal that we extensively evaluate in Section \ref{sec:experiments}. We next develop this strategy for constructing \(\bm{w}_{\text{init}}^{(s,1)}\).

\section{Dynamic Initial Model Construction}
\label{sec:algorithm}
We first summarize the main idea behind our method, the pseudo code of which is presented in Algorithm \ref{alg:dynamic_initial_model_construction}. In our method, the server uses the first \(P\) pilot sessions to save end-of-session global models and constructs a common pilot reference model \(\bm{w}_{\text{Pilot}}\). For each later session, the server starts from this common pilot reference and runs a small number \(V\) of auxiliary FL rounds on a sampled subset \(\mathcal{A}_{\text{grad}}\) to obtain a session signature \(G_s=\bm{w}_{\text{grad}}^{(s,V+1)}-\bm{w}_{\text{Pilot}}\). We refer to \(G_s\) as a Pseudo-Gradient because it captures the update direction induced by the currently available devices from a common reference point; it is not assumed to be an unbiased estimator of \(\nabla F^{(s)}\). The server then compares \(G_s\) with previously stored Pseudo-Gradients \(G_z\), assigns larger weights to historical models whose Pseudo-Gradients are closer to \(G_s\), and uses the resulting weighted average as the warm start for the new session.

The notation in Algorithm \ref{alg:dynamic_initial_model_construction} separates scenario-dictated quantities from design choices as follows: (i) Scenario-dictated quantities: The available set \(\mathcal{K}^{(s)}\) is dictated by device arrivals and departures. The participating set \(\mathcal{A}^{(s,t)}\subseteq\mathcal{K}^{(s)}\) is the ordinary FL client-selection set used during main training and may be determined by resource constraints, or any existing scheduler. The subset \(\mathcal{A}_{\text{grad}}\subseteq\mathcal{K}^{(s)}\) is used only for estimating the session Pseudo-Gradient \(G_s\). (ii) Design choices: The design parameters are \(P\), the number of pilot sessions; \(V\), the number of auxiliary Pseudo-Gradient computation rounds; and \(R\), the sensitivity of the similarity weights.

\begin{algorithm}[t]
    \caption{Dynamic Initial Model Construction}
    \label{alg:dynamic_initial_model_construction}
    \small
    \begin{algorithmic}[1]
        \Require $T$: global rounds per session, $P$: pilot preparation sessions, $S$: total sessions, $V$: Pseudo-Gradient computation global rounds.
        \State \textbf{Initialize:} Random global model $\bm{w}^{(1,1)}$, lists $Q_1$ and $Q_2$, similarity scaling factor $R$. \label{line:initialization}
        \For{each session $s = 1$ to $S$}:
            \For{each global round within session, $t = 1$ to $T$}
                \BeginBox[fill=StageFiveFill, draw=StageFiveDraw, opacity=.2, rounded corners=4pt, line width=2pt]
            \State \textbf{Pilot Preparation} 
            \If{$s\leq P$} \label{line:pilot_session_start}
                    \For{each device $k \in \mathcal{A}^{(s,t)}$} 
                    \State Local training on $\bm{w}^{(s,t)}$ for $e_k^{(s,t)}$ steps.  
                    \State Send final local model to the server.
                    \EndFor 
                    \State The server aggregates local models: $\bm{w}^{(s,t+1)}$.
                    \If{$t == T$} Save $\bm{w}^{(s,T+1)}$ to $Q_1$. \label{line:save_trained_model}\EndIf
                    \If{$s == P$ \textbf{and} $t==T$} \label{line:when_compute_pilot}
                        \State Compute pilot model $\bm{w}_{\text{Pilot}}=\sum_{s = 1}^{P} \bm{w}^{(s, T+1)}/P$ and delete first $P$ saved global models \(\{\bm{w}^{(s, T+1)}\}_{s=1}^{P}\). \label{line:pilot_model_actual} 
                    \EndBox
                    \EndIf
            \EndIf
            \BeginBox[fill=StageOneFill, draw=StageOneDraw, opacity=.13, rounded corners=4pt, line width=2pt]
            \State {\bfseries
                    Pseudo-Gradient Computation (When A Session Starts)%
            } 
            
            \If{$s > P$ \textbf{and} $t==1$} \label{line:gradient_computation_start} 
                        \State $\bm{w}_{\text{grad}}^{(s,1)} \gets \bm{w}_{\text{Pilot}}$
                    \State The server samples subset $\mathcal{A}_{\text{grad}} \subseteq \mathcal{K}^{(s)}$.
                    \For{$v = 1$ to $V$} \label{line:gradient_computation_round}
                        \For{each device $k \in \mathcal{A}_{\text{grad}}$}
                            \State Local training on $\bm{w}_{\text{grad}}^{(s,1)}$ for $e_k^{(s,1)}$ steps. 
                            \State Send final local model to the server.
                        \EndFor
                        \State The server aggregates local models: $\bm{w}_{\text{grad}}^{(s,v+1)}$.
                    \EndFor
                    \State Save \(G_{s}  = \bm{w}_{\text{grad}}^{(s, V+1)} - \bm{w}_{\text{Pilot}}\) to $Q_2$.  \label{line:gradient_compute_actual}
                    \EndBox
                \EndIf
            \BeginBox[fill=StageThreeFill, draw=StageThreeDraw, opacity=.13, rounded corners=4pt, line width=2pt]
            \State {\bfseries
                    Initial Model Construction (When A Session Starts)%
            } 
                \If{$s > P + 1$ \textbf{and} $t==1$} \label{line:initial_model_construction_start}
                            \State Compute $\bm{w}_{\text{init}}^{(s,1)}$ using~\eqref{initial_model_construction} \label{line:initial_model_construction_end}
                        \EndBox
            \EndIf 
            \BeginBox[fill=StageTwoFill, draw=StageTwoDraw, opacity=.13, rounded corners=4pt, line width=2pt]
            \State \textbf{Main Training} 
            \If{$s>P$} \label{line:regular_training_start}
                    \State The server samples participating set $\mathcal{A}^{(s,t)} \subseteq \mathcal{K}^{(s)}$. 
                    \For{each device $k \in \mathcal{A}^{(s,t)}$}
                        \State \textbf{Training with Constructed Model} 
                        \If{$s > P+1$ \textbf{and} $t == 1$} \label{line:using_initial_model_start}
                            \State \(\bm{w}^{(s,1)} \gets \bm{w}_{\text{init}}^{(s,1)}\)\label{line:using_initial_model}
                        \EndIf
                        \State Local training on $\bm{w}^{(s,t)}$ for $e_k^{(s,t)}$ steps.   
                        \State Send final local model to the server.
                    \EndFor 
                    \State The server aggregates local models: $\bm{w}^{(s,t+1)}$. 
                    \If{$t == T$}: Save $\bm{w}^{(s,T+1)}$ to $Q_1$. \label{line:save_trained_model_after_pilot}
                    \EndBox
                    \EndIf
            \EndIf
            \EndFor
        \EndFor
    \end{algorithmic}
\end{algorithm}

\textbf{Pilot Preparation (Lines \ref{line:pilot_session_start}--\ref{line:pilot_model_actual}):} The algorithm begins with a pilot preparation stage spanning the first $P$ sessions. During this phase ($s \le P$), the system performs standard FL training to build a history of model states. In each round $t$, a participating set of devices $\mathcal{A}^{(s,t)}$ trains locally on the current global model $\bm{w}^{(s,t)}$, and their updates are aggregated to form $\bm{w}^{(s,t+1)}$. The global model obtained at the end of each pilot session, $\bm{w}^{(s, T+1)}$, is saved to the repository $Q_1$ (Line \ref{line:save_trained_model}). Upon completion of the $P$-th session (Line \ref{line:when_compute_pilot}), the pilot model is derived by averaging these saved models (Line \ref{line:pilot_model_actual}):
\begin{equation} \label{pilot_model}
\bm{w}_{\text{Pilot}} = \frac{\sum_{s = 1}^{P} \bm{w}^{(s, T+1)}}{P}.
\end{equation}
These $P$ saved global models are subsequently discarded to free memory. The pilot model's primary function is not to be an optimal model but to serve as an efficient, consistent baseline. While its simple average could be influenced by early-stage data heterogeneity, its value lies in providing a reference point for computing Pseudo-Gradients, which capture the unique data distribution of each subsequent session. As shown in Section \ref{sec:experiments} and Table \ref{tab:pilot_comparison_EMA}, simple averaging yields performance comparable to that of a pilot model constructed via the exponential weighted average method.

\textbf{Pseudo-Gradient Computation (Lines \ref{line:gradient_computation_start}--\ref{line:gradient_compute_actual}):} Following the pilot stage, each new session \(s\) ($s > P$) initiates with a Pseudo-Gradient computation process to capture the current data distribution. We introduce an auxiliary model sequence \(\{\bm{w}_{\text{grad}}^{(s, v)}\}_{v=1}^{V+1}\), initialized using the pilot model: \(\bm{w}_{\text{grad}}^{(s, 1)} = \bm{w}_{\text{Pilot}}\). The system executes $V$ global rounds (Line \ref{line:gradient_computation_round}) on a sampled subset $\mathcal{A}_{\text{grad}}$, resulting in a final auxiliary model \(\bm{w}_{\text{grad}}^{(s, V+1)}\). The Pseudo-Gradient $G_s$ is defined as the difference (Line \ref{line:gradient_compute_actual}):
\begin{equation}
G_{s} \triangleq \bm{w}_{\text{grad}}^{(s, V+1)} - \bm{w}_{\text{grad}}^{(s, 1)} = \bm{w}_{\text{grad}}^{(s, V+1)} - \bm{w}_{\text{Pilot}}.
\end{equation}
This quantity is particularly useful for evaluating the similarity between different data distributions. It does not scale with sample count. We call \(G_s\) a Pseudo-Gradient because it measures the model displacement produced by a short auxiliary training run from the common pilot reference. It is used only to compare session data distributions; it is not assumed to be an unbiased estimator of \(\nabla F^{(s)}\). For each session following the pilot preparation stage, only one Pseudo-Gradient is retained.

\textbf{Initial Model Construction (Lines \ref{line:initial_model_construction_start}--\ref{line:initial_model_construction_end}):} If sufficient history exists (i.e., $s > P+1$), the system constructs a tailored initial model $\bm{w}_{\text{init}}^{(s,1)}$ before training begins. This is achieved using the saved models from previous sessions and the Pseudo-Gradients:
\begin{equation} \label{initial_model_construction}
    \bm{w}_{\text{init}}^{(s,1)}  \triangleq \sum\nolimits_{z = P+1}^{s-1} \mu_{s, z}\bm{w}^{(z, T+1)},
\end{equation}
where
\begin{equation}
    \mu_{s, z} = \frac{e^{-R \|G_{s} - G_{z}\|_2}}{\sum_{z = P+1}^{s-1} e^{-R \|G_{s} - G_{z}\|_2}}.
\end{equation}
The update rule in \eqref{initial_model_construction} defines a weighted average of the saved models \(\bm{w}^{(z, T+1)}\). A small difference between the past Pseudo-Gradient \(G_{z}\) and the current Pseudo-Gradient \(G_{s}\) suggests that the data distribution in the \(z\)-th session is similar to that of the current session \(s\). Consequently, a smaller \(\|G_{s} - G_{z}\|_2\) yields a larger \(\mu_{s, z}\), making the corresponding saved model a dominant component of the new initialization. The constant \(R\) in \(\mu_{s,z}\) controls the sensitivity to Pseudo-Gradient differences. When \(R = 0\), the resulting initial model reduces to a simple average of all saved models from the pilot preparation stage, irrespective of data similarity. In contrast, as \(R \rightarrow \infty\), the initial model converges to the single saved model corresponding to the most similar data distribution. As shown in Section \ref{sec:experiments} and Table \ref{tab:ablation_R}, our sensitivity analysis demonstrates that the performance of Algorithm \ref{alg:dynamic_initial_model_construction} is robust across a wide range of \(R\) values.

\textbf{Main Training (Line \ref{line:regular_training_start}--\ref{line:save_trained_model_after_pilot}):} The server samples participating devices $\mathcal{A}^{(s,t)}$ for the current round. A critical integration step occurs at the very start of the session (Lines \ref{line:using_initial_model_start}--\ref{line:using_initial_model}): if a dynamic initial model $\bm{w}_{\text{init}}^{(s,1)}$ was constructed ($s > P+1$), the current global model is explicitly updated to this warm start $\bm{w}^{(s,1)} \leftarrow \bm{w}_{\text{init}}^{(s,1)}$; otherwise, the session begins with the existing global model state. Devices then proceed with local training and model aggregation. Finally, the model obtained after the last round ($\bm{w}^{(s, T+1)}$) is saved to the repository (Line \ref{line:save_trained_model_after_pilot}) to facilitate future initializations.


\begin{figure*}[t]
\vspace{-4mm}
\begin{center}
\includegraphics[width=\linewidth]{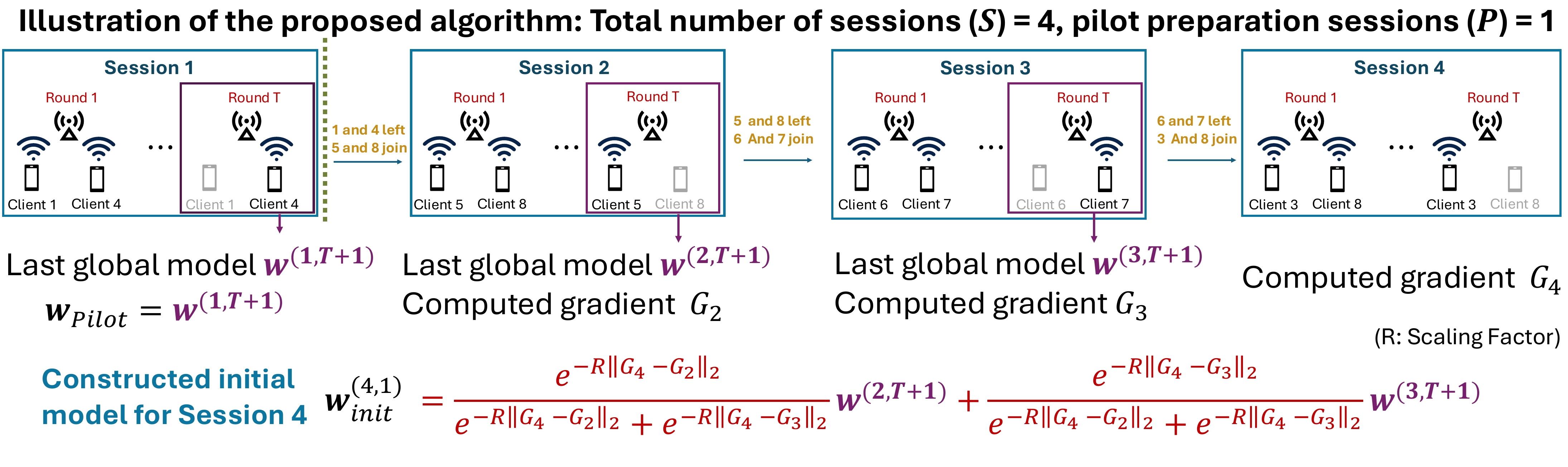}
\vspace{-8mm}
\caption{Illustration of Algorithm \ref{alg:dynamic_initial_model_construction}  when the number of total sessions \(S= 4\) and the number of pilot preparation sessions \(P =1\).}
\label{fig:illustration_algorithm}
\end{center}
\vspace{-4mm}
\end{figure*}

\section{Cost Analysis of Algorithm \ref{alg:dynamic_initial_model_construction}} We analyze the communication, computation, and storage costs of 
Algorithm~\ref{alg:dynamic_initial_model_construction} on a stage-by-stage 
basis and discuss strategies to mitigate these costs in this section.

\noindent\textbf{Pilot Model Construction}
\begin{itemize}[leftmargin=0.35cm]
  \item \emph{Computation:}  
    Each of the $P$ pilot sessions consists of $T$ global rounds. 
In round $t$ ($1 \le t \le T$) of session $s$, each participating device 
$k \in \mathcal{A}^{(s,t)}$ performs $e_k^{(s,t)}$ local epochs. 
The total number of local epochs executed across all pilot sessions is therefore
\[
\sum_{s=1}^{P} \sum_{t=1}^{T} \sum_{k \in \mathcal{A}^{(s,t)}} e_k^{(s,t)}.
\]
  \item \emph{Communication:}  
    There are $P \times T$ global rounds in total. In each round $t$ of session $s$, a set of 
$\lvert \mathcal{A}^{(s,t)} \rvert$ devices upload their locally trained models, 
and the server broadcasts the aggregated model. Consequently, the total number of 
transmissions consists of
\[
\sum_{s=1}^{P} \sum_{t=1}^{T} \lvert \mathcal{A}^{(s,t)} \rvert
\; \text{(device uploads)},  
\;
P T
\; \text{(server broadcasts)},
\]
  \item \emph{Storage:}  
    Let \(Z\) denote the model size. The algorithm temporarily stores \(P\) 
intermediate models (the final model of each pilot session), each of size \(Z\), and discards them after constructing 
the pilot model. The resulting temporary storage cost is therefore \(PZ\).
\end{itemize}

\noindent\textbf{Pseudo-Gradient Computation (for session $s$, $s\in\{P+1,\dots,S\}$)}
\begin{itemize}[leftmargin=0.35cm]
  \item \emph{Computation:}
    Starting from the pilot model, the server performs $V$ additional global
    rounds on a sampled subset $\mathcal{A}_{\text{grad}} \subseteq \mathcal{K}^{(s)}$. In each round $v \in \{1, \dots, V\}$, every device $k \in \mathcal{A}_{\text{grad}}$ runs 
    $e_k^{(s,1)}$ local epochs, yielding a total computational load of
    \[
      V \times \sum_{k \in \mathcal{A}_{\text{grad}}} e_k^{(s,1)}.
    \]

  \item \emph{Communication:}
    Across these $V$ rounds, each round incurs 
    $\lvert \mathcal{A}_{\text{grad}} \rvert$ device uploads and one server broadcast. 
    The total communication cost is therefore
    \[
      V \times \lvert \mathcal{A}_{\text{grad}} \rvert 
      \; \text{(device uploads)}, 
      \; 
      V 
      \; \text{(server broadcasts)}.
    \]

  \item \emph{Storage:}
  For each session \(s\), a single Pseudo-Gradient $G_s$ of size \(Z\) is stored.
  Thus, as sessions progress, the storage for Pseudo-Gradients grows linearly with the number of post-pilot sessions.
\end{itemize}

\noindent\textbf{Coefficient Calculation (for session \(s\in\{P+1,\dots,S\}\))}
\begin{itemize}[leftmargin=0.35cm]
  \item \emph{Computation:}  
    Computing softmax‐based coefficients over the stored Pseudo-Gradients from previous sessions incurs \(\mathcal{O}(s - (P+1))\) cost for vector norms and exponentials.
\end{itemize}

\noindent\textbf{Total Persistent Storage}
\begin{itemize}[leftmargin=0.35cm]
  \item \emph{Pilot model:} 
    A single model $\bm{w}_{\text{Pilot}}$ of size \(Z\) is retained; all intermediate checkpoints are discarded.
  \item \emph{Per post‐pilot session \(s\):} 
  A final model checkpoint $\bm{w}^{(s, T+1)}$ of size \(Z\) and a stored Pseudo-Gradient $G_s$ of size \(Z\).
  \item \emph{Overall growth:}
    The persistent storage grows linearly by \(2Z\) per post‐pilot session, 
    resulting in a total increase proportional to \((S - P)\).
\end{itemize}

\noindent\textbf{Justification for the Costs:}
The upfront cost of pilot training and Pseudo-Gradient computation is offset by accelerated adaptation to shifting device distributions, reducing the total number of global rounds needed to achieve the target accuracy.

\noindent\textbf{Mitigating Storage Cost:} To bound storage growth, one may:
\begin{itemize}[leftmargin=0.35cm]
  \item \emph{Limited history:} Retain only the most recent $H$ sessions' models/Pseudo-Gradients.
  \item \emph{Selective retention:} Store models/Pseudo-Gradients that maximize distributional diversity, preserving initialization quality with fewer saved models.
\end{itemize}
These strategies ensure manageable storage while maintaining algorithmic efficacy in dynamic federated learning settings.

\section{Experiments} \label{sec:experiments}

\textbf{Seamless Integration with Representative FL Algorithms:} To demonstrate the plug-and-play nature and broad applicability of our proposed 
initial model construction method, we integrate it with four representative and 
widely used FL algorithms: MOON~\cite{li2021modelmoon}, 
FedProx~\cite{li2020federated}, SCAFFOLD~\cite{karimireddy2020scaffold}, and 
FedACG~\cite{kim2024communication}. Notably, FedACG is the state-of-the-art 
method outperforming adaptive FL algorithms such as FedAvgM \cite{hsu2019measuring}, FedADAM \cite{reddi2020adaptive}, and FedDyn \cite{acar2021federated}. 
Demonstrating compatibility with FedACG highlights that our approach can be 
effectively combined with the strongest modern FL algorithms. Importantly, our method does not replace these algorithms, but rather complements them by addressing the critical challenge of inter-session 
transition. These algorithms were originally developed for static device sets 
within a single session. Our method enhances their performance in dynamic 
multi-session settings by providing a well-initialized global model tailored for 
each session. In all experiments, the pilot model (\(\bm{w}_{\text{Pilot}}\)) is 
fixed and reused across all sessions, taken from the last round of the first 
session. To construct the initialization for each subsequent session, we use only 
one global round (\(V = 1\)) to obtain the Pseudo-Gradient.

\textbf{Baselines:} To the best of our knowledge, there is no existing work addressing the inter-session initialization problem in a dynamic FL system. Therefore, we construct three intuitive yet strong baselines to benchmark our method:
\begin{itemize}
    \item \textbf{Previous:} Reuses the preceding session’s model for continued training without explicitly constructing a new initial model~\cite{mcmahan2017communication}.
    
    \item \textbf{Average:} Initializes models using the unweighted average of prior global models.
    
    \item \textbf{Continuous (image datasets only):} Uses Class-Semantic Relation Distillation Loss~\cite{dong2022federated}, anchoring the current model to the previous session’s model to preserve class relationships while adapting to new data.
\end{itemize}
 
\textbf{Task and Dataset:} We perform classification tasks on seven diverse benchmarks, including six image datasets (MNIST~\cite{lecun1998mnist}, Fashion-MNIST~\cite{xiao2017fashionmnist}, SVHN~\cite{netzer2011reading}, CIFAR-10~\cite{krizhevsky2009learning}, CIFAR-100~\cite{krizhevsky2009learning}, TinyImageNet~\cite{krizhevsky2009learning}) and the AG News text corpus~\cite{zhang2015character}. We show results for TinyImageNet and AG News in the main text, and the remainder in our online technical report \cite{chang2025federatedlearningdynamicclient}.

\textbf{Label Distribution and Evaluation Protocol:} In our main experiments, each session contains 100 active devices. The active device set changes only at session boundaries, and the setup allows a device that leaves to re-enter in a later non-consecutive session. Each device is assigned a fixed local training subset at the beginning of the experiment; therefore, when a device re-enters, it trains again on the same local samples. The train and test splits remain disjoint. To simulate controlled distribution shifts, the session-level label subsets are constructed from the class annotations available in the benchmark training split. We partition these labels according to the specified cross-session overlap, which ranges from zero overlap (disjoint session label sets) to high overlap (most labels shared across consecutive sessions). For the labels available in a session, samples are distributed among the active devices using a Dirichlet distribution parameterized by $\alpha$ \cite{mcmahan2017communication}. A smaller value of $\alpha$ produces a more heterogeneous label distribution across devices, while a larger value of $\alpha$ produces a more homogeneous distribution. For evaluation, we use the standard held-out test split of each dataset. At each round, the test set is filtered to include all test samples whose labels appear among the currently active devices, and the reported accuracy is computed on this active-session label subset. Thus, the reported accuracies measure adaptation to the current session's label distribution rather than accuracy over labels absent from that session.

\begin{table}[t]
    \caption{Simulation Parameters}
    \label{table:simulation_parameters_comm_comp}
    \centering
    \begin{tabular}{@{}ll@{}}
        \toprule
        \textbf{Parameter} & \textbf{Value} \\
        \midrule
        \multicolumn{2}{c}{\textit{Network Geometry \& Channel Model}} \\
        Carrier Frequency ($f_c$) & $3.5$ GHz \\
        Cell Radius ($R_{\text{cell}}$) & $250$ m \\
        Minimum Device Distance ($d_{\text{min}}$) & $10$ m \\
        BS Height ($h_{\text{BS}}$) & $10$ m \\
        User Terminal Height ($h_{\text{UT}}$) & $1.5$ m \\
        Environment Height ($h_E$) & $1.0$ m \\
        Shadowing Standard Deviation ($\sigma_{SF}$) & $4$ dB \\
        Device Velocity ($v_k$) & $10$ m/s ($36$ km/h) \\
        \midrule
        \multicolumn{2}{c}{\textit{Communication System}} \\
        System Bandwidth ($W$) & $100$ MHz \\
        Noise Power Spectral Density ($N_0$) & $-174$ dBm/Hz \\
        Server Transmit Power ($P_{\text{server}}^{\text{tx}}$) & $20$ W ($43$ dBm) \\
        Device Transmit Power ($P_k^{\text{tx}}$) & $200$ mW ($23$ dBm) \\
        Device Circuit Power ($P_k^{\text{rx}}$) & $100$ mW \\
        Time Slot Duration ($\tau$) & $0.5$ ms \\
        \midrule
        \multicolumn{2}{c}{\textit{Computation Model}} \\
        CPU Clock Frequency ($f_k$) & $1$ GHz \\
        Processor Efficiency ($\chi$) & $8$ FLOPs/cycle \\
        Effective Chipset Capacitance ($\xi_k$) & $10^{-28}$ \\
        \midrule
        \multicolumn{2}{c}{\textit{Model Specifications}} \\
        MNIST Parameters ($N_p(\text{\tiny MNIST})$) & $7,850$ \\
        MNIST Complexity ($\phi(\text{\tiny MNIST})$) & $2.00$ FLOPs/Param \\
        FMNIST Parameters ($N_p(\text{\tiny FMNIST})$) & $7,850$ \\
        FMNIST Complexity ($\phi(\text{\tiny FMNIST})$) & $2.00$ FLOPs/Param \\
        SVHN Parameters ($N_p(\text{\tiny SVHN})$) & $667,178$ \\
        SVHN Complexity ($\phi(\text{\tiny SVHN})$) & $91.04$ FLOPs/Param \\
        CIFAR-10 Parameters ($N_p(\text{\tiny C-10})$) & $11,573,066$ \\
        CIFAR-10 Complexity ($\phi(\text{\tiny C-10})$) & $6.50$ FLOPs/Param \\
        CIFAR-100 Parameters ($N_p(\text{\tiny C-100})$) & $11,596,196$ \\
        CIFAR-100 Complexity ($\phi(\text{\tiny C-100})$) & $6.50$ FLOPs/Param \\
        TinyImageNet Params ($N_p(\text{\tiny TinyImg})$) & $21,730,056$ \\
        TinyImageNet Complexity ($\phi(\text{\tiny TinyImg})$) & $27.69$ FLOPs/Param \\
        AG News Parameters ($N_p(\text{\tiny AGNews})$) & $9,157,804$ \\
        AG News Complexity ($\phi(\text{\tiny AGNews})$) & $2.63 \times 10^{-4}$ FLOPs/Param \\
        \bottomrule
    \end{tabular}
\end{table}

\textbf{Communication and Computation Model}:
To evaluate the real-world performance of Algorithm \ref{alg:dynamic_initial_model_construction}, we model the latency and energy constraints imposed by the wireless environment. We consider a Time Division Duplex (TDD) system to separate uplink and downlink transmission intervals. Within each interval, we assume an Orthogonal Frequency Division Multiple Access (OFDMA) scheme where devices are assigned orthogonal resource blocks (frequency channels), ensuring an interference-free environment.

\subsubsection{Channel Modeling}
The wireless channel between device $k$ and the server is characterized by both large-scale and small-scale fading. The total channel gain $g_k[n]$ at time slot $n$ is given by:
\begin{equation}
    |g_k[n]|^2 = \psi_k \cdot |h_k[n]|^2,
\end{equation}
where $\psi_k$ denotes the \textit{large-scale fading} coefficient and $h_k[n]$ represents the \textit{small-scale fading} component. 

For the large-scale fading models, we adopt the 3GPP Urban Micro (UMi) Street Canyon model defined in the technical report TR 38.901 \cite{3gpp38901}, which is the standard for 5G/6G simulations in dense urban environments. The large-scale fading coefficient $\psi_k$ captures the combined impact of path loss and shadowing.  Shadowing is modeled via a random variable $\xi$ following a zero-mean Gaussian distribution, denoted as $\xi \sim \mathcal{N}(0, \sigma_{SF}^2)$. For the UMi-LoS scenario, the standard deviation is defined as $\sigma_{SF} = 4 \text{ dB}$. Accordingly, the total path loss, encompassing both propagation loss and shadowing, is expressed as $PL_{\text{UMi-LoS}} + \xi$ where $PL_{\text{UMi-LoS}}$ is the Line-of-Sight (LoS) path loss for Urban Micro scenario.  
Consequently, the linear large-scale fading coefficient $\psi_k$ is obtained by converting the total path loss from the logarithmic scale:
\begin{equation}
    \psi_k = 10^{-(PL_{\text{UMi-LoS}} + \xi)/10}.
\end{equation}
We assume $\psi_k$ remains constant during a single communication round.

We consider a single-cell system where the server (BS) is positioned at the center of a circular coverage area. The device locations are modeled as a Poisson Point Process (PPP) with uniform areal density within the cell, conditioned on a cell radius of $R_{\text{cell}} = 250$ m and a minimum distance of $d_{\text{min}} = 10$ m to ensure the validity of the path loss model. To satisfy the uniform areal density requirement, the Euclidean 2D distance $d_{\text{2D}, k}$ of device $k$ from the BS follows the probability density function (PDF):
\begin{equation}
    f(d) = \frac{2d}{R_{\text{cell}}^2 - d_{\text{min}}^2}, \quad d_{\text{min}} \le d \le R_{\text{cell}}.
\end{equation}
The resulting 3D distance used in the path loss calculation is then given by $d_{\text{3D}, k} = \sqrt{d_{\text{2D}, k}^2 + (h_{\text{BS}} - h_{\text{UT}})^2}$.

The small-scale fading factor $h_k[n]$ is modeled as a first-order time-varying Gauss-Markov process:
\begin{equation}
    h_k[n] = \rho_k h_k[n-1] + \sqrt{1 - \rho_k^2} e_k[n],
\end{equation}
where $e_k[n] \sim \mathcal{CN}(0, 1)$ is an independent and identically distributed (i.i.d.) circularly symmetric complex Gaussian random variable. This term $e_k[n]$ introduces uncorrelated random fluctuations into the channel state. Consequently, the magnitude $|h_k[n]|$ follows a Rayleigh distribution \cite{amiri2020federated}, which effectively models a dense scattering wireless environment.

\subsubsection{Instantaneous Data Rate}
Based on the TDD assumption, the uplink and downlink channels share the same small-scale fading coefficient $h_k[n]$ within a time slot. The \textit{Uplink Instantaneous Data Rate} (from device $k$ to the server) at time slot $n$ is calculated as:
\begin{equation}
    R_k^{\text{UL}}[n] = W \log_2 \left( 1 + \frac{P_k^{\text{tx}} \psi_k |h_k[n]|^2}{N_0 W} \right),
\end{equation}
where $W$ denotes the allocated channel bandwidth per device (Hz), $P_k^{\text{tx}}$ is the transmission power of device $k$, and $N_0$ is the noise power spectral density.

Conversely, the \textit{Downlink Instantaneous Data Rate} (from the server to device $k$) is given by:
\begin{equation}
    R_k^{\text{DL}}[n] = W \log_2 \left( 1 + \frac{P_{\text{server}}^{\text{tx}} \psi_k |h_k[n]|^2}{N_0 W} \right),
\end{equation}
where $P_{\text{server}}^{\text{tx}}$ is the transmission power of the server. Note that typically $P_{\text{server}}^{\text{tx}} \gg P_k^{\text{tx}}$, resulting in higher downlink rates.

\subsubsection{Round-Trip Latency}
The total latency for a complete FL round involves the downlink transmission of the global model and the uplink transmission of the local model updates. Let $Z$ denote the size of the model parameters in bits. Since the data rate varies over time, the transmission time is calculated by accumulating the instantaneous rate over discrete time slots of duration $\tau$. The total round-trip communication latency $T_k^{\text{comm}}$ for device $k$ is: $T_k^{\text{comm}} = T_k^{\text{DL}} + T_k^{\text{UL}},$ where the downlink time $T_k^{\text{DL}}$ and uplink time $T_k^{\text{UL}}$ are the minimum durations required to satisfy:
\begin{equation}
    \sum_{n=1}^{T_k^{\text{DL}}/\tau} R_k^{\text{DL}}[n] \cdot \tau \geq Z \quad \text{and} \quad \sum_{n=1}^{T_k^{\text{UL}}/\tau} R_k^{\text{UL}}[n] \cdot \tau \geq Z.
\end{equation}

\subsubsection{Communication Energy Consumption}
We focus on the energy consumption of the devices, as they are typically battery-constrained. The total communication energy for device $k$, denoted as $E_k^{\text{comm}}$, comprises the energy required for receiving the global model (downlink) and transmitting the local update (uplink): $E_k^{\text{comm}} = E_k^{\text{DL}} + E_k^{\text{UL}}.$ The downlink energy consumption is determined by the receiving circuit power $P_k^{\text{rx}}$ and the downlink duration: $E_k^{\text{DL}} = P_k^{\text{rx}} \cdot T_k^{\text{DL}}.$
The uplink energy consumption is determined by the transmission power $P_k^{\text{tx}}$ and the uplink duration: $E_k^{\text{UL}} = P_k^{\text{tx}} \cdot T_k^{\text{UL}}.$
Note that typically $P_k^{\text{tx}} \gg P_k^{\text{rx}}$, making the uplink transmission the dominant factor in communication energy consumption.

\subsubsection{Local Model Computation}
The local computation latency is determined by the complexity of the specific neural network architecture $\mathcal{M}$ employed for the learning task and the device's processor efficiency. Let $N_p({\mathcal{M}})$ denote the number of trainable parameters in model $\mathcal{M}$. We define $\phi(\mathcal{M})$ as the computational density ratio, representing the number of floating point operations (FLOPs) required per parameter for a full forward and backward propagation (FLOPs/param). Consequently, the computational workload to process a single data sample is given by $N_p(\mathcal{M}) \cdot \phi(\mathcal{M})$ FLOPs.

To account for hardware capabilities such as instruction-level parallelism (e.g., SIMD instructions), let $\chi$ denote the processor efficiency, defined as the number of FLOPs the CPU can execute per clock cycle (FLOPs/cycle). Given that device $k$ performs $e_k^{(s,t)}$ local SGD iterations with a mini-batch size of $B^{(s,t)}$ during round $t$ of session $s$, the local computation time $T_k^{\text{comp}}$ is formulated as:
\begin{equation} \label{eq:local_computation_time}
    T_k^{\text{comp}} = \frac{N_p(\mathcal{M}) \cdot \phi(\mathcal{M}) \cdot e_k^{(s,t)} \cdot B^{(s,t)}}{f_k \cdot \chi},
\end{equation}
where $f_k$ is the CPU clock frequency of device $k$ (in Hz).

The energy consumption is modeled based on the effective chipset capacitance $\xi_k$ \cite{tran2019federated}. Since the required number of CPU cycles is inversely proportional to the processor efficiency $\chi$, the computation energy $E_k^{\text{comp}}$ is given by:
\begin{equation} \label{eq:local_computation_energy}
    E_k^{\text{comp}} = \xi_k \left( \frac{N_p(\mathcal{M}) \cdot \phi(\mathcal{M}) \cdot e_k^{(s,t)} \cdot B^{(s,t)}}{\chi} \right) (f_k)^2.
\end{equation}
Parameter values used for simulating communication and computation are listed in Table \ref{table:simulation_parameters_comm_comp}.

Unless otherwise stated, accuracy is evaluated on the held-out test samples whose labels are present in the active session.

\begin{table*}[!t]
\vspace{-4mm}
  \caption{Performance comparison of implementing Algorithm \ref{alg:dynamic_initial_model_construction} with MOON, FedProx, SCAFFOLD and FedACG under different label distributions for TinyImageNet dataset for a system with 100 devices.}
  \vspace{-4mm}
  \label{table:tinyimagenet_100_appendix}
  \begin{center}
  \begin{small}
  \resizebox{\textwidth}{!}{ 
  \begin{tabular}{c c c c c c c c c c c c}
    \toprule
    \multirow{2}{*}{\textbf{Dataset}}
      & \multirow{2}{*}{\textbf{Overlap}}
      & \multirow{2}{*}{\textbf{Dirichlet}\newline\boldmath$\,\alpha$}
      & \multirow{2}{*}{\textbf{FL Alg.}}
      & \multicolumn{4}{c}{\textbf{Second Session Transition}} 
      & \multicolumn{4}{c}{\textbf{Fourth Session Transition}} \\ 
    \cmidrule(lr){5-8} \cmidrule(lr){9-12}
      & & &
      & Proposed & Continuous & Previous & Average 
      & Proposed & Continuous & Previous & Average \\ 
    \midrule
    \midrule
    \multirow{16}{*}{TinyImageNet}
      & \multirow{8}{*}{\makecell{0.0 \\ (Non-Overlap)}}
      & \multirow{4}{*}{0.3}
        & MOON   
          & \textbf{65.98±0.30} & 56.16±0.26 & 58.58±0.51 & 58.15±0.33
          & \textbf{67.03±0.42} & 57.64±0.67 & 61.71±0.79 & 62.57±0.45 \\
      & & 
        & FedProx  
          & \textbf{65.33±0.30} & 56.03±0.31 & 63.03±0.36 & 61.15±0.38
          & \textbf{66.56±0.41} & 57.57±0.48 & 64.73±0.14 & 63.80±0.22 \\
      & & 
        & SCAFFOLD 
          & \textbf{67.51±0.17} & 52.43±0.92 & 60.19±0.26 & 61.66±0.24
          & \textbf{68.47±0.23} & 54.64±0.65 & 61.11±0.38 & 63.81±0.27 \\
      & & 
        & FedACG   
          & \textbf{66.54±0.19} & 56.05±0.34 & 32.99±3.50 & 55.69±1.47
          & \textbf{67.11±0.47} & 57.78±0.50 & 36.26±1.55 & 62.29±0.35 \\
      \cmidrule(lr){3-12}
      & & \multirow{4}{*}{0.7}
        & MOON   
          & \textbf{66.29±0.47} & 57.73±0.29 & 59.72±0.70 & 61.17±0.57
          & \textbf{67.77±0.32} & 58.79±0.42 & 62.64±0.55 & 64.33±0.33 \\
      & & 
        & FedProx  
          & \textbf{66.21±0.21} & 57.71±0.35 & 63.88±0.22 & 61.98±0.37
          & \textbf{67.26±0.16} & 58.75±0.46 & 65.25±0.39 & 64.40±0.21 \\
      & & 
        & SCAFFOLD 
          & \textbf{68.55±0.35} & 55.33±0.63 & 60.67±0.54 & 62.50±0.32
          & \textbf{69.23±0.41} & 56.46±0.41 & 61.12±0.85 & 64.71±0.28 \\
      & & 
        & FedACG   
          & \textbf{66.71±0.43} & 57.74±0.29 & 38.89±2.21 & 58.76±1.20
          & \textbf{67.84±0.28} & 58.78±0.41 & 31.27±21.23 & 63.96±0.30 \\
    \cmidrule(lr){2-12}
      & \multirow{8}{*}{\makecell{0.2 \\ (Slight Overlap)}}
      & \multirow{4}{*}{0.3}
        & MOON   
          & \textbf{64.99±0.66} & 58.67±0.35 & 60.88±0.51 & 60.79±0.29
          & \textbf{66.36±0.61} & 59.79±0.42 & 61.91±0.54 & 62.98±0.27 \\
      & & 
        & FedProx  
          & \textbf{65.48±0.43} & 58.64±0.26 & 63.75±0.17 & 60.47±0.35
          & \textbf{66.32±0.32} & 59.80±0.43 & 65.38±0.28 & 63.45±0.50 \\
      & & 
        & SCAFFOLD 
          & \textbf{67.58±0.46} & 56.77±0.49 & 60.84±0.45 & 62.22±0.38
          & \textbf{68.09±0.31} & 57.66±0.69 & 61.35±0.57 & 64.48±0.45 \\
      & & 
        & FedACG   
          & \textbf{66.33±0.70} & 58.65±0.35 & 57.24±0.51 & 58.62±0.42
          & \textbf{66.39±0.40} & 59.84±0.48 & 55.29±0.96 & 62.42±0.49 \\
      \cmidrule(lr){3-12}
      & & \multirow{4}{*}{0.7}
        & MOON   
          & \textbf{66.09±0.58} & 59.94±0.41 & 62.47±0.20 & 62.56±0.28
          & \textbf{67.03±0.39} & 60.69±0.52 & 63.18±0.49 & 64.20±0.22 \\
      & & 
        & FedProx  
          & \textbf{66.06±0.22} & 60.00±0.51 & 64.40±0.17 & 61.49±0.31
          & \textbf{67.06±0.31} & 60.58±0.54 & 65.46±0.22 & 64.06±0.38 \\
      & & 
        & SCAFFOLD 
          & \textbf{68.30±0.19} & 58.53±0.50 & 61.07±0.41 & 63.57±0.25
          & \textbf{68.83±0.33} & 58.76±0.41 & 61.16±0.28 & 65.35±0.40 \\
      & & 
        & FedACG   
          & \textbf{66.51±0.34} & 60.02±0.42 & 56.94±0.73 & 61.96±0.45
          & \textbf{67.06±0.12} & 60.58±0.44 & 56.75±0.72 & 63.95±0.36 \\
    \bottomrule
  \end{tabular}}
  \end{small}
  \end{center}
 \end{table*}

\begin{table*}[!t]
  \caption{Performance comparison of implementing Algorithm \ref{alg:dynamic_initial_model_construction} with MOON, FedProx, SCAFFOLD and FedACG under different label distributions for the AG News dataset with 100 devices.}
  \label{table:ag_news_100_appendix}
  \begin{center}
  \begin{small}
  \resizebox{0.9\textwidth}{!}{ 
  \begin{tabular}{c c c c c c c c c c c c}
    \toprule
    \multirow{2}{*}{\textbf{Dataset}}
      & \multirow{2}{*}{\textbf{Overlap}}
      & \multirow{2}{*}{\textbf{Dirichlet}\newline\boldmath$\,\alpha$}
      & \multirow{2}{*}{\textbf{FL Alg.}}
      & \multicolumn{3}{c}{\textbf{Second Session Transition}} 
      & \multicolumn{3}{c}{\textbf{Fourth Session Transition}} \\ 
    \cmidrule(lr){5-7} \cmidrule(lr){8-10}
      & & &
      & Proposed & Previous & Average 
      & Proposed & Previous & Average \\ 
    \midrule
    \midrule
    \multirow{16}{*}{AG News}
      & \multirow{8}{*}{\makecell{0.0 \\ (Non-Overlap)}}
      & \multirow{4}{*}{0.3}
        & MOON   
          & \textbf{73.94±1.43} & 14.04±0.29 & 14.22±0.23
          & \textbf{75.19±1.35} & 14.34±0.31 & 24.60±2.00 \\
      & & 
        & FedProx  
          & \textbf{73.89±1.43} & 14.03±0.28 & 14.20±0.21
          & \textbf{75.10±1.41} & 14.34±0.31 & 24.33±1.95 \\
      & & 
        & SCAFFOLD 
          & \textbf{66.46±2.18} & 12.66±0.36 & 12.86±0.18
          & \textbf{69.33±1.78} & 13.44±0.45 & 13.85±0.25 \\
      & & 
        & FedACG   
          & \textbf{76.27±1.28} & 14.52±0.30 & 16.95±1.06
          & \textbf{77.18±1.41} & 14.78±0.29 & 35.19±2.65 \\
      \cmidrule(lr){3-10}
      & & \multirow{4}{*}{0.7}
        & MOON   
          & \textbf{73.95±1.40} & 14.04±0.30 & 14.23±0.23
          & \textbf{74.83±1.50} & 14.34±0.32 & 24.59±1.99 \\
      & & 
        & FedProx  
          & \textbf{73.90±1.40} & 14.03±0.29 & 14.20±0.21
          & \textbf{74.75±1.56} & 14.34±0.31 & 24.32±1.93 \\
      & & 
        & SCAFFOLD 
          & \textbf{66.64±2.14} & 12.63±0.21 & 12.88±0.18
          & \textbf{68.99±3.04} & 13.24±0.16 & 13.89±0.26 \\
      & & 
        & FedACG   
          & \textbf{76.25±1.28} & 14.52±0.29 & 16.96±1.06
          & \textbf{76.88±1.55} & 14.78±0.29 & 35.19±2.64 \\
    \cmidrule(lr){2-10}
      & \multirow{8}{*}{\makecell{0.8 \\ (Significant Overlap)}}
      & \multirow{4}{*}{0.3}
        & MOON   
          & \textbf{59.15±0.66} & 51.98±0.85 & 53.36±0.85
          & \textbf{60.78±1.19} & 52.97±0.83 & 53.94±0.97 \\
      & & 
        & FedProx  
          & \textbf{59.05±0.62} & 51.96±0.83 & 53.34±0.85
          & \textbf{60.54±1.31} & 52.96±0.83 & 53.93±0.96 \\
      & & 
        & SCAFFOLD 
          & 47.26±0.62 & 48.35±0.39 & \textbf{48.58±1.07}
          & 49.11±1.25 & \textbf{50.09±0.52} & 49.48±0.96 \\
      & & 
        & FedACG   
          & \textbf{63.95±0.28} & 52.89±0.76 & 55.26±0.66
          & \textbf{65.28±0.90} & 53.83±0.57 & 55.82±0.51 \\
      \cmidrule(lr){3-10}
      & & \multirow{4}{*}{0.7}
        & MOON   
          & \textbf{59.20±0.67} & 51.99±0.83 & 53.36±0.85
          & \textbf{60.48±0.26} & 52.97±0.83 & 53.95±0.98 \\
      & & 
        & FedProx  
          & \textbf{59.09±0.65} & 51.97±0.83 & 53.34±0.84
          & \textbf{60.27±0.34} & 52.96±0.83 & 53.92±0.98 \\
      & & 
        & SCAFFOLD 
          & 47.37±0.61 & \textbf{48.87±0.21} & 48.73±1.07
          & 48.68±0.72 & \textbf{50.42±0.27} & 49.58±1.00 \\
      & & 
        & FedACG   
          & \textbf{63.95±0.29} & 52.92±0.75 & 55.26±0.66
          & \textbf{64.54±0.36} & 53.84±0.59 & 55.83±0.50 \\
    \bottomrule
  \end{tabular}}
  \end{small}
  \end{center}
 \end{table*}

\textbf{Performance Comparison and Analysis:} Table \ref{table:tinyimagenet_100_appendix} and Table \ref{table:ag_news_100_appendix} report the mean accuracy $\pm$ standard deviation across three random seeds for our proposed initialization strategy combined with MOON, FedProx, SCAFFOLD, and FedACG, compared to baselines on Tiny-ImageNet and AG News in the initial rounds following each session transition. Across diverse datasets and distribution scenarios, our method effectively mitigates accuracy degradation arising from device dynamics. Even in significant overlap settings where Previous and Average baselines are theoretically favored, our method retains a competitive advantage. Minor exceptions occur with SCAFFOLD due to its inherent variance reduction mechanism. Crucially, our scheme achieves the target accuracy with significantly fewer communication rounds. This effectively saves communication costs compared to baselines that require extensive retraining.

\begin{figure*}[!t]
\begin{center}
\includegraphics[width=\linewidth]{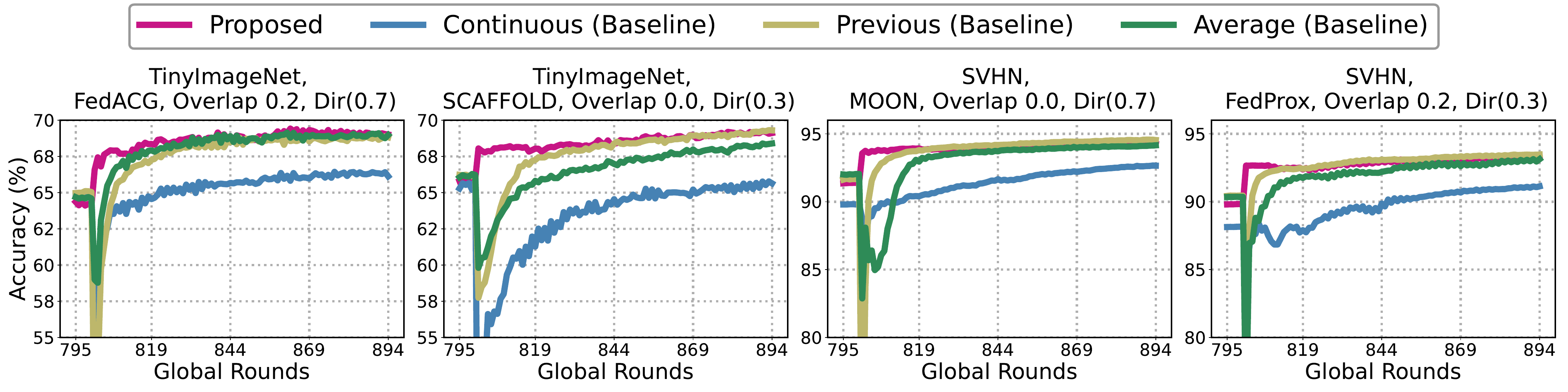}
\caption{Comparative performance analysis of the proposed algorithm against baseline FL methods across diverse datasets. The results validate that our scheme attains target accuracy with significantly fewer communication rounds, thereby effectively reducing communication costs under dynamic scenarios of device arrivals and departures.}
\label{fig:combined_subplots_1x4_with_legend_new}
\end{center}
\end{figure*}

These gains are achieved efficiently using only a single shared pilot model and one global round for Pseudo-Gradient computation per session. Figure \ref{fig:combined_subplots_1x4_with_legend_new} visualizes this efficiency. Baselines suffer severe and instantaneous accuracy degradation at session transitions and often plummet to near zero accuracy. In contrast, our method avoids such drops. By incurring the marginal cost of only one additional global round, our approach eliminates the need for the prolonged recovery phases observed in standard methods and offers a robust solution for dynamic FL systems.

\begin{table*}[!t]
  \caption{Comparison of Pilot Construction Methods For 9 Total Sessions with 3 Pilot Sessions: Average vs. Exponentially Weighted Average (EMA) for selected scenarios. The performance difference is negligible.}
  \vspace{-4mm}
  \label{tab:pilot_comparison_EMA}
  \begin{center}
  \begin{small}
  \resizebox{0.85\textwidth}{!}{ 
  \begin{tabular}{c c c c c c c c c}
    \toprule
    \multirow{2}{*}{\textbf{Dataset}}
      & \multirow{2}{*}{\textbf{Overlap}}
      & \multirow{2}{*}{\textbf{Dirichlet}\newline\boldmath$\,\alpha$}
      & \multirow{2}{*}{\textbf{FL Alg.}}
      & \multirow{2}{*}{\textbf{Pilot Construction}}
      & \multicolumn{4}{c}{\textbf{Average Accuracy for Each Transition (\%)}} \\
    \cmidrule(lr){6-9} 
      & & & & & \textbf{1st} & \textbf{2nd} & \textbf{3rd} & \textbf{4th} \\
    \midrule
    \midrule

    \multirow{2}{*}{AG News}
      & \multirow{2}{*}{0.0}
      & \multirow{2}{*}{0.7}
      & \multirow{2}{*}{SCAFFOLD}
      & Average
      & 60.92$\pm$1.17 & 66.63$\pm$2.14 & 64.11$\pm$0.67 & 69.43$\pm$2.64 \\
      & & & & EMA
      & 60.92$\pm$1.17 & 66.63$\pm$2.14 & 64.15$\pm$0.71 & 69.62$\pm$2.39 \\
    \cmidrule(lr){1-9}

    \multirow{2}{*}{SVHN}
      & \multirow{2}{*}{0.2}
      & \multirow{2}{*}{0.3}
      & \multirow{2}{*}{MOON}
      & Average
      & 89.69$\pm$0.78 & 93.28$\pm$0.17 & 91.26$\pm$0.38 & 93.81$\pm$0.12 \\
      & & & & EMA
      & 89.69$\pm$0.82 & 93.27$\pm$0.19 & 91.27$\pm$0.38 & 93.80$\pm$0.12 \\
    \cmidrule(lr){1-9}

    \multirow{2}{*}{CIFAR-100}
      & \multirow{2}{*}{0.2}
      & \multirow{2}{*}{0.7}
      & \multirow{2}{*}{FedACG}
      & Average
      & 49.54$\pm$0.53 & 53.50$\pm$0.48 & 53.28$\pm$0.47 & 55.53$\pm$0.70 \\
      & & & & EMA
      & 49.82$\pm$0.28 & 53.69$\pm$0.43 & 53.02$\pm$0.52 & 55.90$\pm$0.46 \\
    \cmidrule(lr){1-9}

    \multirow{2}{*}{\makecell{TinyImageNet}}
      & \multirow{2}{*}{0.0}
      & \multirow{2}{*}{0.3}
      & \multirow{2}{*}{FedProx}
      & Average
      & 62.02$\pm$0.38 & 65.88$\pm$0.28 & 63.93$\pm$0.22 & 67.18$\pm$0.23 \\
      & & & & EMA
      & 61.97$\pm$0.43 & 65.99$\pm$0.21 & 64.02$\pm$0.33 & 67.11$\pm$0.35 \\
    \bottomrule
  \end{tabular}}
  \end{small}
  \end{center}
\end{table*}

\textbf{Ablation Study (Different Pilot Construction Methods)}: To validate the robustness of our pilot construction strategy, we investigate whether a more sophisticated aggregation method improves the quality of the generated initial models. Specifically, we compare our proposed simple average (Eq. \ref{pilot_model}) against an Exponentially Weighted Moving Average (EMA) construction. While the simple average assigns uniform importance ($1/P$) to all pilot sessions, the EMA approach prioritizes the most recent global models. Under the EMA scheme, the pilot model is computed as
\begin{equation}
    \bm{w}_{\text{Pilot}}^{\text{EMA}} = \sum_{s=0}^{P-1} \gamma_s \bm{w}_s^{\text{last}},
\end{equation}
where the weighting coefficients $\gamma_s$ are determined by a decay factor $\beta \in [0, 1)$ as follows:
\begin{equation}
    \gamma_s = 
    \begin{cases} 
        \beta^{P-1}, & \text{if } s = 0 \text{ (oldest session),} \\
        (1 - \beta) \beta^{P-1-s}, & \text{if } 0 < s \le P-1.
    \end{cases}
\end{equation}
where $\sum_{s=0}^{P-1} \gamma_s = 1$. For example, with $P=3$ pilot sessions, the aggregated pilot model is computed as: $\bm{w}_{\text{Pilot}}^{\text{EMA}} = \beta^2 \bm{w}_0^{\text{last}} + \beta(1-\beta) \bm{w}_1^{\text{last}} + (1-\beta) \bm{w}_2^{\text{last}}.$ This pilot formulation theoretically adapts faster to recent trends in the data distribution.

Table \ref{tab:pilot_comparison_EMA} presents the accuracy comparison across four distinct datasets and FL algorithms. We observe that the performance difference between the two methods is negligible across all scenarios. This consistency implies that the specific weighting of historical models in the pilot phase does not significantly alter the effectiveness of the subsequently computed Pseudo-Gradients $G_s$. Consequently, we adopt the simple average in Algorithm \ref{alg:dynamic_initial_model_construction}, as it achieves competitive performance without introducing additional hyperparameters.

\begin{table*}[!t]
  \caption{Ablation Study on the Number of Pseudo-Gradient Computation Rounds ($V$). Comparison of accuracy (mean $\pm$ std) for the Second and Fourth Session Transitions. Using $V=1$ provides the best trade-off between performance and computational efficiency.}
  \vspace{-7mm}
  \label{table:ablation_V}
  \begin{center}
  \begin{small}
  \resizebox{1.0\textwidth}{!}{ 
  \begin{tabular}{c c c c c c c c c c}
    \toprule
    \multirow{2}{*}{\textbf{Dataset}}
      & \multirow{2}{*}{\textbf{Overlap}}
      & \multirow{2}{*}{\textbf{Dirichlet}\newline\boldmath$\,\alpha$}
      & \multirow{2}{*}{\textbf{FL Alg.}}
      & \multicolumn{3}{c}{\textbf{2nd Session Transition Accuracy (\%)}} 
      & \multicolumn{3}{c}{\textbf{4th Session Transition Accuracy (\%)}} \\
    \cmidrule(lr){5-7} \cmidrule(lr){8-10} 
      & & & & \textbf{\textit{V}=1} & \textbf{\textit{V}=2} & \textbf{\textit{V}=3} & \textbf{\textit{V}=1} & \textbf{\textit{V}=2} & \textbf{\textit{V}=3} \\
    \midrule
    \midrule

    AG News
      & 0.0
      & 0.7
      & SCAFFOLD
      & 66.64$\pm$2.14 & 66.62$\pm$2.18 & 66.63$\pm$2.16
      & 68.99$\pm$3.04 & 69.80$\pm$2.39 & 69.51$\pm$2.41 \\
    \cmidrule(lr){1-10}

    SVHN
      & 0.2
      & 0.3
      & MOON
      & 93.23$\pm$0.18 & 93.26$\pm$0.20 & 93.29$\pm$0.24
      & 93.86$\pm$0.19 & 93.89$\pm$0.21 & 93.86$\pm$0.17 \\
    \cmidrule(lr){1-10}

    CIFAR-100
      & 0.2
      & 0.7
      & FedACG
      & 53.52$\pm$0.14 & 52.91$\pm$0.60 & 53.00$\pm$0.79
      & 55.82$\pm$0.26 & 55.42$\pm$0.24 & 55.17$\pm$0.58 \\
    \cmidrule(lr){1-10}

    \makecell{TinyImageNet}
      & 0.0
      & 0.3
      & FedProx
      & 65.33$\pm$0.30 & 65.77$\pm$0.31 & 65.79$\pm$0.37
      & 65.56$\pm$0.41 & 66.96$\pm$0.22 & 67.14$\pm$0.22 \\
    \bottomrule
  \end{tabular}}
  \end{small}
  \end{center}
\end{table*}

\textbf{Ablation Study (Sensitivity Analysis of $V$)}:
This study evaluates how the number of Pseudo-Gradient computation rounds ($V$) affects the balance between performance stability and computational cost. While increasing $V$ can refine the session Pseudo-Gradient, the empirical results in Table \ref{table:ablation_V} demonstrate that setting $V > 1$ yields negligible improvements. Accuracy fluctuations across all datasets remain within standard deviations; for instance, SVHN exhibits a difference of less than $0.03\%$ between $V=1$ and $V=2$. Since computational and communication overheads scale linearly with $V$, we adopt $V=1$ to maximize efficiency without compromising accuracy.

\textbf{Sensitivity Analysis of Similarity Scaling Factor ($R$)}:
We evaluate the algorithm's robustness to the scaling factor $R$, which regulates the sensitivity of the weighting coefficients $\mu_{s,z}$ to Pseudo-Gradient differences. Table \ref{tab:ablation_R} summarizes performance across $R$ values tailored to the Pseudo-Gradient norms of each dataset (e.g., higher magnitudes for AG News). The results indicate that the algorithm is highly stable across variations in $R$. Accuracy fluctuations are statistically negligible; for instance, SVHN shows a variation of less than $0.1\%$, and the performance margins for AG News and CIFAR-100 largely overlap within standard deviations. This suggests that while $R$ must be set to an appropriate order of magnitude, precise fine-tuning is not necessary to achieve good performance.

\begin{table*}[!t]
  \caption{Sensitivity Analysis of the Similarity Scaling Factor ($R$). Comparison of accuracy (mean $\pm$ std) for the Second and Fourth Session Transitions. The algorithm demonstrates robustness to variations in $R$ within reasonable ranges.}
  \vspace{-7mm}
  \label{tab:ablation_R}
  \begin{center}
  \begin{small}
  \resizebox{1.0\textwidth}{!}{ 
  \begin{tabular}{c c c c c c c c c c}
    \toprule
    \multirow{2}{*}{\textbf{Dataset (Values of $R$)}}
      & \multirow{2}{*}{\textbf{Overlap}}
      & \multirow{2}{*}{\textbf{Dirichlet}\newline\boldmath$\,\alpha$}
      & \multirow{2}{*}{\textbf{FL Alg.}}
      & \multicolumn{3}{c}{\textbf{2nd Session Transition Accuracy (\%)}} 
      & \multicolumn{3}{c}{\textbf{4th Session Transition Accuracy (\%)}} \\
    \cmidrule(lr){5-7} \cmidrule(lr){8-10} 
      & & & & \textbf{Low \boldmath$R$} & \textbf{Med \boldmath$R$} & \textbf{High \boldmath$R$} & \textbf{Low \boldmath$R$} & \textbf{Med \boldmath$R$} & \textbf{High \boldmath$R$} \\
    \midrule
    \midrule

    \makecell{AG News \\ \scriptsize{($R \in \{800, 1000, 1200\}$)}}
      & 0.0
      & 0.7
      & SCAFFOLD
      & 66.64$\pm$2.14 & 66.64$\pm$2.14 & 66.64$\pm$2.14
      & 68.99$\pm$3.04 & 68.82$\pm$3.26 & 68.69$\pm$3.42 \\
    \cmidrule(lr){1-10}

    \makecell{SVHN \\ \scriptsize{($R \in \{10, 50, 100\}$)}}
      & 0.2
      & 0.3
      & MOON
      & 93.23$\pm$0.18 & 93.26$\pm$0.19 & 93.25$\pm$0.19
      & 93.86$\pm$0.19 & 93.88$\pm$0.21 & 93.89$\pm$0.23 \\
    \cmidrule(lr){1-10}

    \makecell{CIFAR-100 \\ \scriptsize{($R \in \{10, 50, 100\}$)}}
      & 0.2
      & 0.7
      & FedACG
      & 53.52$\pm$0.14 & 53.61$\pm$0.48 & 53.62$\pm$0.41
      & 55.82$\pm$0.26 & 55.91$\pm$1.03 & 55.57$\pm$1.25 \\
    \cmidrule(lr){1-10}

    \makecell{Tiny ImageNet \\ \scriptsize{($R \in \{10, 50, 100\}$)}}
      & 0.0
      & 0.3
      & FedProx
      & 65.33$\pm$0.30 & 65.99$\pm$0.23 & 66.01$\pm$0.35
      & 66.56$\pm$0.41 & 66.79$\pm$0.57 & 66.55$\pm$0.49 \\
    \bottomrule
  \end{tabular}}
  \end{small}
  \end{center}
\end{table*}

\textbf{Quantification of Performance Gain}: We quantify performance based on the model's ability to rapidly adapt with two metrics: (1) \textit{Time-to-Accuracy} and (2) \textit{Accumulated Accuracy Gain}.
\textit{Time-to-Accuracy (Adaptation Speed)} metric measures the latency of adaptation following the start of a new session. It is defined as the number of global rounds $t$ (within the current session $s$) required for an algorithm to recover to a specific percentage ($\rho$) of the peak accuracy achieved by the proposed scheme in that session. Let $Acc_{s}^{(\text{alg})}(t)$ denote the accuracy of an algorithm at the $t$-th global round of session $s$, and let $Acc^*_{s, \text{prop}}$ be the maximum accuracy achieved by the proposed method within session $s$. The Time-to-Accuracy $T_{\rho}$ is defined as:
\begin{equation}
    T_{\rho} = \min \{ t \in [1, T] \mid Acc_{s}^{(\text{alg})}(t) \ge \rho \cdot Acc^*_{s, \text{prop}} \}
\end{equation}

where $T$ is the total number of global rounds per session, and $\rho \in \{0.95, 0.97\}$. A lower $T_{\rho}$ indicates that the algorithm requires fewer global rounds (and consequently less latency and energy) to adapt to the new distribution. If the target is not reached within the session, $T_{\rho}$ is denoted as $\infty$.

While \textit{Time-to-Accuracy} measures convergence speed to a specific target, it fails to capture the stability or sustained magnitude of performance throughout the session. To address this, we define the \textit{Accumulated Accuracy Gain}, which quantifies the cumulative performance advantage of the proposed method over a baseline.

Calculated as the sum of accuracy differentials between the proposed method and a baseline $b$, the total gain $G_{\text{total}}$ for session $s$ is defined as:
\begin{equation}
    G_{\text{total}} = \sum_{t=1}^{T} \left( Acc_{s}^{(\text{prop})}(t) - Acc_{s}^{(b)}(t) \right)
\end{equation}
A positive $G_{\text{total}}$ indicates that the proposed method maintains a superior accuracy trajectory compared to the baseline across the session's global rounds.

Table \ref{tab:performance_metrics} presents the quantitative results for Time-to-Accuracy and Accumulated Accuracy Gain, highlighting the efficiency of the proposed dynamic initialization strategy. The most significant advantage is the substantial reduction in latency and energy consumption, computed using our communication and computation model. Across all scenarios, the proposed method demonstrates rapid adaptation, consistently recovering the target accuracy in just a single round. In the challenging TinyImageNet tasks (specifically under the SCAFFOLD algorithm), this efficiency contrasts sharply with the \textit{Continuous} baseline, which requires 150 rounds to reach the same threshold. This convergence gap translates into substantial resource savings; the proposed method consumes only 1.43 kJ of energy compared to the 214.33 kJ required by the Continuous baseline, representing a $150\times$ reduction in resource overhead.

Furthermore, the \textit{Accumulated Accuracy Gain} column confirms that the proposed method maintains a sustained performance advantage throughout the entire adaptation phase. This improvement is most pronounced in scenarios with zero overlap (e.g., SCAFFOLD on TinyImageNet), where our method yields a total accuracy gain of $+867.1\%$ over the Continuous baseline. Against the \textit{Previous} baseline, our method consistently achieves positive total gains, validating that the proposed weighted initialization strategy provides a superior starting point compared to simply reusing the last session's model.
\begin{table*}[!t]
\vspace{-4mm}
  \caption{Comparison of Time-to-Accuracy (95\% Target for TinyImageNet, 97\% Target for SVHN) and Accumulated Accuracy Gain.}
  \vspace{-4mm}
  \label{tab:performance_metrics}
  \begin{center}
  \resizebox{1.0\textwidth}{!}{
  \begin{tabular}{c c c c l  r r r r c}
    \toprule
    \multirow{2}{*}{\textbf{Dataset}} & 
    \multirow{2}{*}{\textbf{Overlap}} & 
    \multirow{2}{*}{\textbf{Dirichlet} $\bm{\alpha}$} & 
    \multirow{2}{*}{\textbf{FL Alg.}} & 
    \multirow{2}{*}{\textbf{Method}} & 
    \multicolumn{4}{c}{\textbf{Time-to-Accuracy (95\%/97\% Target)}} & 
    \textbf{Accumulated} \\
    \cmidrule(lr){6-9}
    & & & & & 
    \textbf{Rounds} & 
    \textbf{Latency (s)} & 
    \textbf{Energy (kJ)} & 
    \textbf{Speedup} & 
    \textbf{Accuracy Gain (\%)} \\
    \midrule

    \multirow{4}{*}{TinyImageNet} & 
    \multirow{4}{*}{0.2} & 
    \multirow{4}{*}{0.7} & 
    \multirow{4}{*}{FedACG} 
    & \textbf{Proposed} & \textbf{1} & \textbf{113.6} & \textbf{1.43} & \textbf{---} & \textbf{---} \\
    & & & & Previous (Baseline) & 11 & 1,249.4 & 15.72 & $11.0\times$ & $+139.6$ \\
    & & & & Continuous (Baseline) & 55 & 6,247.2 & 78.59 & $55.0\times$ & $+542.4$ \\
    & & & & Average (Baseline) & 7 & 795.1 & 10.00 & $7.0\times$ & $+61.1$ \\
    \cmidrule(lr){1-10}

    \multirow{4}{*}{TinyImageNet} & 
    \multirow{4}{*}{0.0} & 
    \multirow{4}{*}{0.3} & 
    \multirow{4}{*}{SCAFFOLD} 
    & \textbf{Proposed} & \textbf{1} & \textbf{113.6} & \textbf{1.43} & \textbf{---} & \textbf{---} \\
    & & & & Previous (Baseline) & 14 & 1,590.2 & 20.00 & $14.0\times$ & $+85.7$ \\
    & & & & Continuous (Baseline) & 150 & 17,037.7 & 214.33 & $150.0\times$ & $+867.1$ \\
    & & & & Average (Baseline) & 28 & 3,180.4 & 40.01 & $28.0\times$ & $+243.5$ \\
    \cmidrule(lr){1-10}

    \multirow{4}{*}{SVHN} & 
    \multirow{4}{*}{0.0} & 
    \multirow{4}{*}{0.7} & 
    \multirow{4}{*}{MOON} 
    & \textbf{Proposed} & \textbf{1} & \textbf{6.9} & \textbf{0.08} & \textbf{---} & \textbf{---} \\
    & & & & Previous (Baseline) & 5 & 34.4 & 0.39 & $5.0\times$ & $+11.5$ \\
    & & & & Continuous (Baseline) & 52 & 358.1 & 4.04 & $52.0\times$ & $+395.1$ \\
    & & & & Average (Baseline) & 14 & 96.4 & 1.09 & $14.0\times$ & $+116.1$ \\
    \cmidrule(lr){1-10}

    \multirow{4}{*}{SVHN} & 
    \multirow{4}{*}{0.2} & 
    \multirow{4}{*}{0.3} & 
    \multirow{4}{*}{FedProx} 
    & \textbf{Proposed} & \textbf{1} & \textbf{6.9} & \textbf{0.08} & \textbf{---} & \textbf{---} \\
    & & & & Previous (Baseline) & 4 & 27.5 & 0.31 & $4.0\times$ & $+8.0$ \\
    & & & & Continuous (Baseline) & 70 & 482.1 & 5.44 & $70.0\times$ & $+517.6$ \\
    & & & & Average (Baseline) & 10 & 68.9 & 0.78 & $10.0\times$ & $+115.0$ \\

    \bottomrule
  \end{tabular}
  }
  \end{center}
  \vspace{-4mm}
\end{table*}

\begin{figure}[!t]
    \centering
    \includegraphics[width=\columnwidth]{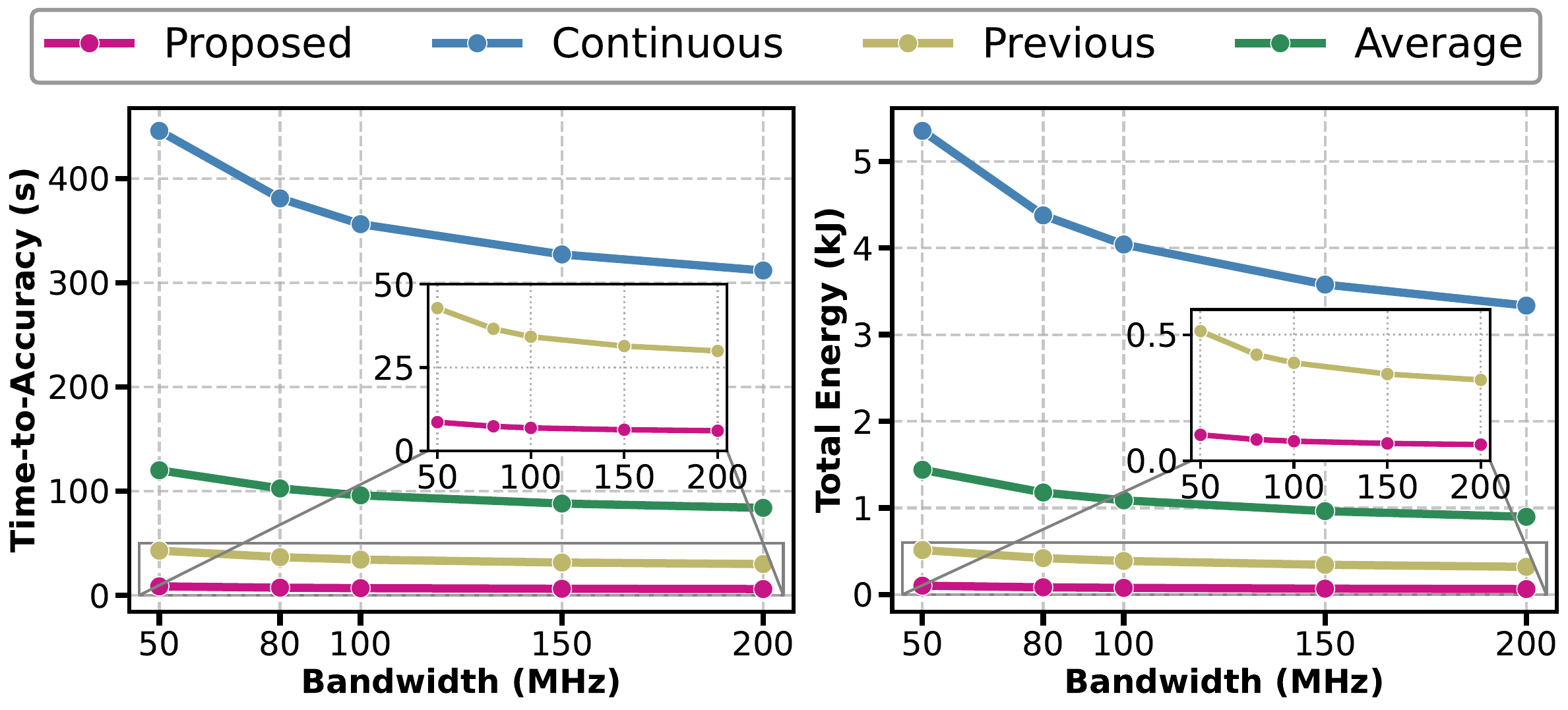}
    \caption{Impact of system bandwidth on Total Latency (left) and Total Energy Consumption (right) required to reach 97\% of the proposed scheme's peak accuracy on the SVHN dataset. The proposed method consistently achieves the lowest latency and energy consumption across all bandwidth configurations (50--200 MHz). Insets highlight the performance gap between the Proposed and Previous schemes at a finer scale.}
    \label{fig:svhn_bandwidth}
\end{figure}

\textbf{Impact of System Bandwidth}: We further investigate the robustness of the proposed framework to communication resource constraints by varying the system bandwidth from 50 MHz to 200 MHz. Figure \ref{fig:svhn_bandwidth} illustrates the total latency and energy consumption required to recover 97\% of the maximum accuracy for our proposed scheme on the SVHN dataset. The results demonstrate that our method consistently achieves the target accuracy with fewer communication rounds. Consequently, our approach effectively saves communication costs and yields the lowest resource overhead across all bandwidth configurations. While higher bandwidth naturally reduces transmission time for all methods, the proposed algorithm maintains a distinct efficiency advantage as highlighted by the insets comparing it to Previous baseline. This confirms that the efficiency gains are driven by the reduction in required training rounds and algorithmic convergence rather than being dependent on specific channel conditions.

\textbf{Other Experiments:} Our online technical report \cite{chang2025federatedlearningdynamicclient} provides supplementary results on MNIST, Fashion-MNIST, SVHN, CIFAR-10, and CIFAR-100 under additional device-scale and non-IID settings.

\section{Conclusion}
In this paper, we addressed the structural challenges inherent to Federated Learning in wireless edge networks, specifically those driven by dynamic device arrival and departure. We presented a theoretical convergence analysis that explicitly accounts for the shifting optimization goals associated with evolving device sets, quantifying the impact of gradient noise and data heterogeneity. Motivated by the initialization-related term in this analysis, we proposed a plug-and-play initialization strategy that accelerates adaptation by constructing a weighted average of historical models based on gradient similarity. This approach effectively bridges distributional shifts between sessions without requiring server-side knowledge of future device activities. Extensive simulations demonstrate that our method achieves convergence speedups typically an order of magnitude or more compared to baselines and substantial energy savings compared to state-of-the-art baselines. Future work will develop adaptive triggering mechanisms for session initialization to further reduce overhead in highly volatile environments.




\bibliographystyle{IEEEtran}
\bibliography{tnse}


 





\iftnseincludesupplement
\appendices
\onecolumn

\section{Communication and Computation Modeling Details}
\label{appendix:comm_comp_model}

To evaluate the real-world performance of Algorithm \ref{alg:dynamic_initial_model_construction} in a realistic wireless edge network, we model the latency and energy constraints imposed by the physical environment. We consider a Time Division Duplex (TDD) system to separate uplink and downlink transmission intervals. Within each interval, we assume an Orthogonal Frequency Division Multiple Access (OFDMA) scheme where devices are assigned orthogonal resource blocks, ensuring an interference-free environment.

\subsection{Channel Modeling}
The wireless channel between device $k$ and the server is characterized by both large-scale and small-scale fading. The total channel gain $g_k[n]$ at time slot $n$ is given by:
\begin{equation}
    |g_k[n]|^2 = \psi_k \cdot |h_k[n]|^2,
\end{equation}
where $\psi_k$ denotes the \textit{large-scale fading} coefficient and $h_k[n]$ represents the \textit{small-scale fading} component. 

\subsubsection{Large-Scale Fading}
We adopt the 3GPP Urban Micro (UMi) Street Canyon model defined in the technical report TR 38.901 \cite{3gpp38901}, the standard for 5G/6G simulations in dense urban environments. The coefficient $\psi_k$ captures the combined impact of path loss and shadowing. Shadowing is modeled via a random variable $\xi$ following a zero-mean Gaussian distribution, $\xi \sim \mathcal{N}(0, \sigma_{SF}^2)$. For the UMi-LoS scenario, the standard deviation is $\sigma_{SF} = 4 \text{ dB}$. The linear large-scale fading coefficient $\psi_k$ is obtained by converting the total path loss from the logarithmic scale:
\begin{equation}
    \psi_k = 10^{-(PL_{\text{UMi-LoS}} + \xi)/10}.
\end{equation}
We assume $\psi_k$ remains constant during a single communication round. The path loss $PL_{\text{UMi-LoS}}$ is divided into two segments based on a ``Break Point'' distance ($d'_{BP}$):
\begin{equation}
    PL_{\text{UMi-LoS}}[\text{dB}] = 
    \begin{cases} 
    PL_1, & \text{if } 10 \text{m} \le d_{\text{2D}} \le d'_{BP} \\ 
    PL_2, & \text{if } d'_{BP} < d_{\text{2D}} \le 5 \text{km} 
    \end{cases}
\end{equation}
where $d_{\text{2D}}$ is the 2D distance between the device and the Base Station (BS). The segment path losses are:
\begin{align}
    PL_1 &= 32.4 + 21 \log_{10}(d_{\text{3D}}) + 20 \log_{10}(f_c), \\
    PL_2 &= 32.4 + 40 \log_{10}(d_{\text{3D}}) + 20 \log_{10}(f_c) \notag \\
    &\quad - 9.5 \log_{10}\left( (d'_{BP})^2 + (h_{\text{BS}} - h_{\text{UT}})^2 \right),
\end{align}
where $f_c$ is the carrier frequency in GHz, $h_{\text{BS}}$ is the BS height, $h_{\text{UT}}$ is the user terminal height, and $d_{\text{3D}} = \sqrt{d_{\text{2D}}^2 + (h_{\text{BS}} - h_{\text{UT}})^2}$ is the 3D distance. 

Following TR 38.901 \cite{3gpp38901}, the break point distance is calculated using effective antenna heights to account for the environment height ($h_E = 1.0$ m). The effective heights are $h'_{\text{BS}} = h_{\text{BS}} - h_E$ and $h'_{\text{UT}} = h_{\text{UT}} - h_E.$ Consequently, the break point distance is $d'_{BP} = (4 h'_{\text{BS}} h'_{\text{UT}} f_c \times 10^9)/c,$ where $c = 3 \times 10^8$ m/s.

\subsubsection{Device Distribution}
We consider a single-cell system where the server is at the center of a circular coverage area. Device locations are modeled as a Poisson Point Process (PPP) with uniform areal density within the cell, conditioned on a radius $R_{\text{cell}} = 250$ m and a minimum distance $d_{\text{min}} = 10$ m. The Euclidean 2D distance $d_{\text{2D}, k}$ of device $k$ follows the probability density function (PDF):
\begin{equation}
    f(d) = \frac{2d}{R_{\text{cell}}^2 - d_{\text{min}}^2}, \quad d_{\text{min}} \le d \le R_{\text{cell}}.
\end{equation}

\subsubsection{Small-Scale Fading}
The factor $h_k[n]$ is modeled as a first-order time-varying Gauss-Markov process:
\begin{equation}
    h_k[n] = \rho_k h_k[n-1] + \sqrt{1 - \rho_k^2} e_k[n],
\end{equation}
where $e_k[n] \sim \mathcal{CN}(0, 1)$ is an i.i.d. circularly symmetric complex Gaussian random variable. The temporal correlation coefficient $\rho_k$ is determined by the zeroth-order Bessel function of the first kind, $J_0(\cdot)$, as $\rho_k = J_0(2\pi f_d \tau)$, where $\tau$ is the slot duration and $f_d$ is the maximum Doppler frequency, given by $f_d = v_k f_c \times 10^9/c$. To ensure the channel remains approximately constant within a slot ($\tau \ll T_c$), we calculate the coherence time $T_c \approx 0.423/f_d$ and select $\tau = 0.5$ ms.

\subsection{Instantaneous Data Rate and Latency}
Based on the TDD assumption, the \textit{Uplink Instantaneous Data Rate} (device $k$ to server) at time slot $n$ is:
\begin{equation}
    R_k^{\text{UL}}[n] = W \log_2 \left( 1 + \frac{P_k^{\text{tx}} \psi_k |h_k[n]|^2}{N_0 W} \right),
\end{equation}
where $W$ is the channel bandwidth per device, $P_k^{\text{tx}}$ is the transmission power of device $k$, and $N_0$ is the noise power spectral density. The \textit{Downlink Instantaneous Data Rate} $R_k^{\text{DL}}[n]$ is calculated similarly using the server's transmission power $P_{\text{server}}^{\text{tx}}$.

The total round-trip communication latency $T_k^{\text{comm}}$ for device $k$ to transmit model parameters of size $Z$ bits is:
\begin{equation}
    T_k^{\text{comm}} = T_k^{\text{DL}} + T_k^{\text{UL}},
\end{equation}
where $T_k^{\text{DL}}$ and $T_k^{\text{UL}}$ are the minimum durations satisfying $\sum_{n=1}^{T_k^{\text{DL}}/\tau} R_k^{\text{DL}}[n] \cdot \tau \geq Z$ and $\sum_{n=1}^{T_k^{\text{UL}}/\tau} R_k^{\text{UL}}[n] \cdot \tau \geq Z$, respectively.

\subsection{Energy Consumption and Computation}
\subsubsection{Communication Energy}
The total communication energy for device $k$ comprises the energy for receiving the global model and transmitting the local update:
\begin{equation}
    E_k^{\text{comm}} = P_k^{\text{rx}} \cdot T_k^{\text{DL}} + P_k^{\text{tx}} \cdot T_k^{\text{UL}},
\end{equation}
where $P_k^{\text{rx}}$ is the receiving circuit power. Typically $P_k^{\text{tx}} \gg P_k^{\text{rx}}$, making uplink transmission the dominant factor.

\subsubsection{Computation Latency and Energy}
Let $N_p({\mathcal{M}})$ denote the number of parameters in model $\mathcal{M}$ and $\phi(\mathcal{M})$ the computational density ratio (FLOPs/param). To account for hardware capabilities, let $\chi$ denote the processor efficiency (FLOPs/cycle). Given that device $k$ performs $e_k^{(s,t)}$ local SGD iterations with mini-batch size $B^{(s,t)}$ during round $t$ of session $s$, the local computation time $T_k^{\text{comp}}$ is:
\begin{equation}
    T_k^{\text{comp}} = \frac{N_p(\mathcal{M}) \cdot \phi(\mathcal{M}) \cdot e_k^{(s,t)} \cdot B^{(s,t)}}{f_k \cdot \chi},
\end{equation}
where $f_k$ is the CPU clock frequency. The computation energy is modeled via the effective chipset capacitance $\xi_k$:
\begin{equation}
    E_k^{\text{comp}} = \xi_k \left( \frac{N_p(\mathcal{M}) \cdot \phi(\mathcal{M}) \cdot e_k^{(s,t)} \cdot B^{(s,t)}}{\chi} \right) (f_k)^2.
\end{equation}

\section{Proof of Theorem \ref{theorem:upper_bound}}
\label{appendix:proof_upper_bound}

To streamline the derivation, we unify the session index $s \in \{1, \dots, S\}$ and the intra-session round index $t \in \{1, \dots, T\}$ from the main text into a single cumulative global round index $g \in \{1, \dots, G\}$, where $G = ST$. The mapping between these notations is given by $g = (s-1)T + t$. For instance, in a system where $T=10$, the 5th round of the 3rd session ($s=3, t=5$) corresponds to the cumulative index $g = (3-1)10 + 5 = 25$. 

Consequently, all parameters are mapped to this unified index $g$. For time-varying parameters such as the learning rate, we define $\eta^{(g)} \equiv \eta^{(s,t)}$. For parameters that remain constant throughout a session, such as the available device set $\mathcal{K}^{(s)}$, the notation $\mathcal{K}^{(g)}$ refers to the set associated with the unique session $s$ containing round $g$ (i.e., $\mathcal{K}^{(g)} \equiv \mathcal{K}^{(s)}$ holds for all $g$ such that $\lceil g/T \rceil = s$). Consistent with Theorem \ref{theorem:upper_bound}, the proof considers full participation of the currently available devices in each round, i.e., \(\mathcal{A}^{(g)}=\mathcal{K}^{(g)}\), and a common number of local steps \(e_k^{(g)}=e^{(g)}\) for all \(k\in\mathcal{K}^{(g)}\). We use \(\mathbb{E}_g[\cdot]\) to denote expectation over all mini-batch randomness in round \(g\), conditioned on the history up to the beginning of the round. Under this conditioning, the local iterates \(\bm{w}_k^{(g),e}\) for \(e\ge 1\) remain random because they depend on earlier mini-batch draws in the same round.

Define \(p_k^{(g)}\triangleq D_k^{(g)}/D^{(g)}\). Under full participation, the global update can be written as
\begin{equation} \label{different_global_model}
    \bm{w}^{(g+1)}-\bm{w}^{(g)}
    = \sum_{k\in\mathcal{K}^{(g)}} p_k^{(g)}\nabla \overline{F}_k^{(g)},
\end{equation}

where the accumulated local model displacement is
\begin{equation} \label{accumulated_gradient}
    \nabla \overline{F}_k^{(g)}
    =
    \bm{w}_k^{(g),e^{(g)}}-\bm{w}^{(g)}
    =
    -\eta^{(g)}
    \sum_{e=1}^{e^{(g)}}\frac{1}{B_k^{(g)}}
    \sum_{d\in\mathcal{B}_k^{(g),e}}
    \nabla_{\bm{w}}f(\bm{w}_k^{(g),e-1},d).
\end{equation}

For each local step \(e\), Assumption \ref{assumption:uniform_minibatch} gives the conditional unbiasedness relation
\begin{equation}
    \mathbb{E}_g\left[
    \frac{1}{B_k^{(g)}}\sum_{d\in\mathcal{B}_k^{(g),e}}
    \nabla_{\bm{w}}f(\bm{w}_k^{(g),e-1},d)
    \right]
    =
    \nabla F_k^{(g)}(\bm{w}_k^{(g),e-1}).
\end{equation}

We first derive a one-round descent inequality. By \(\beta\)-smoothness,
\begin{equation}\label{proof_smoothness}
    F^{(g)}(\bm{w}^{(g+1)})
    \leq
    F^{(g)}(\bm{w}^{(g)})
    + \left\langle \nabla F^{(g)}(\bm{w}^{(g)}),\bm{w}^{(g+1)}-\bm{w}^{(g)}\right\rangle
    + \frac{\beta}{2}\|\bm{w}^{(g+1)}-\bm{w}^{(g)}\|^2.
\end{equation}

Let
\begin{align}
G^{(g)}
&\triangleq \nabla F^{(g)}(\bm{w}^{(g)})
=\sum_{j\in\mathcal{K}^{(g)}}p_j^{(g)}
\nabla F_j^{(g)}(\bm{w}^{(g)}), \label{eq:proof_global_gradient}\\
U^{(g)}
&\triangleq
\sum_{k\in\mathcal{K}^{(g)}}p_k^{(g)}
\sum_{e=1}^{e^{(g)}}\frac{1}{B_k^{(g)}}\sum_{d\in\mathcal{B}_k^{(g),e}}
\nabla_{\bm w} f(\bm{w}_k^{(g),e-1},d),\\
H^{(g)}
&\triangleq
\sum_{k\in\mathcal{K}^{(g)}}p_k^{(g)}
\sum_{e=1}^{e^{(g)}}\nabla F_k^{(g)}(\bm{w}_k^{(g),e-1}).
\end{align}

Then \(\bm{w}^{(g+1)}-\bm{w}^{(g)}=-\eta^{(g)}U^{(g)}\) and \(\mathbb{E}_g[U^{(g)}]=\mathbb{E}_g[H^{(g)}]\). Substituting these relations into \eqref{proof_smoothness} yields the aggregate-first descent form
\begin{align}
\mathbb{E}_g[F^{(g)}(\bm{w}^{(g+1)})]
&\leq
F^{(g)}(\bm{w}^{(g)})
-\eta^{(g)}\mathbb{E}_g\langle G^{(g)},H^{(g)}\rangle
+\frac{\beta(\eta^{(g)})^2}{2}\mathbb{E}_g\|U^{(g)}\|^2.
\label{eq:aggregate_first_descent}
\end{align}

The comparison in \eqref{eq:aggregate_first_descent} is between the global gradient \(G^{(g)}\) and the aggregate full-batch local direction \(H^{(g)}\). This is the key correction relative to the original proof: we do not compare \(G^{(g)}\) separately with each fixed-client gradient \(\nabla F_k^{(g)}(\bm{w}^{(g)})\), so no artificial \(e_{\min}^{(g)}\)-dependent term is introduced.

Next, we bound the sampling part. Define the zero-mean mini-batch error at local step \(e\) on device \(k\) as
\begin{equation}
\delta_{k,e}^{(g)}
\triangleq
\frac{1}{B_k^{(g)}}\sum_{d\in\mathcal{B}_k^{(g),e}}
\nabla_{\bm w} f(\bm{w}_k^{(g),e-1},d)
-\nabla F_k^{(g)}(\bm{w}_k^{(g),e-1}).
\end{equation}

Then \(U^{(g)}-H^{(g)}=\sum_k p_k^{(g)}\sum_e\delta_{k,e}^{(g)}\). Using Jensen's inequality over the weights \(p_k^{(g)}\), the conditional unbiasedness of each \(\delta_{k,e}^{(g)}\), and the finite-population variance formula for sampling without replacement,
\begin{align}
\mathbb{E}_g\|U^{(g)}-H^{(g)}\|^2
&\leq
\sum_{k\in\mathcal{K}^{(g)}}p_k^{(g)}
\sum_{e=1}^{e^{(g)}}\mathbb{E}_g\|\delta_{k,e}^{(g)}\|^2 \nonumber\\
&\leq
\sum_{k\in\mathcal{K}^{(g)}}p_k^{(g)}
\sum_{e=1}^{e^{(g)}}
\left(1-\frac{B_k^{(g)}}{D_k^{(g)}}\right)
\frac{(\sigma_k^{(g),e-1})^2}{B_k^{(g)}},
\label{eq:proof_sampling_variance_raw}
\end{align}

where
\begin{align}
(\sigma_k^{(g),e-1})^2
&\triangleq
\frac{1}{D_k^{(g)}-1}
\sum_{i=1}^{D_k^{(g)}}
\left\|
\nabla_{\bm w}f(\bm{w}_k^{(g),e-1},d_{k,i}^{(g)})
-\nabla F_k^{(g)}(\bm{w}_k^{(g),e-1})
\right\|^2 .
\end{align}

Writing \(d_{k,i}^{(g)}=(\bm{x}_{k,i}^{(g)},y_{k,i}^{(g)})\), the variance term is bounded using Assumption \ref{assumption:local_data_variability}:
\begin{align}
(\sigma_k^{(g),e-1})^2
&=
\frac{1}{D_k^{(g)}-1}
\sum_{i=1}^{D_k^{(g)}}
\left\|
\frac{1}{D_k^{(g)}}\sum_{i'=1}^{D_k^{(g)}}
\Big(
\nabla_{\bm w}f(\bm{w}_k^{(g),e-1},d_{k,i}^{(g)})
-\nabla_{\bm w}f(\bm{w}_k^{(g),e-1},d_{k,i'}^{(g)})
\Big)
\right\|^2 \nonumber\\
&\leq
\frac{1}{(D_k^{(g)})^2}
\sum_{i=1}^{D_k^{(g)}}\sum_{i'=1}^{D_k^{(g)}}
\left\|
\nabla_{\bm w}f(\bm{w}_k^{(g),e-1},d_{k,i}^{(g)})
-\nabla_{\bm w}f(\bm{w}_k^{(g),e-1},d_{k,i'}^{(g)})
\right\|^2 \nonumber\\
&\leq
\frac{(\Theta^{(g)})^2}{(D_k^{(g)})^2}
\sum_{i=1}^{D_k^{(g)}}\sum_{i'=1}^{D_k^{(g)}}
\|\bm{x}_{k,i}^{(g)}-\bm{x}_{k,i'}^{(g)}\|^2 \nonumber\\
&=
\frac{2(\Theta^{(g)})^2}{D_k^{(g)}}
(\widetilde{\sigma}_k^{(g)})^2.
\label{eq:proof_sample_variance_bound}
\end{align}

The final equality follows from \(\sum_i(\bm{x}_{k,i}^{(g)}-\bm{\lambda}_k^{(g)})=0\), which implies \(\sum_{i,i'}\|\bm{x}_{k,i}^{(g)}-\bm{x}_{k,i'}^{(g)}\|^2=2D_k^{(g)}\sum_i\|\bm{x}_{k,i}^{(g)}-\bm{\lambda}_k^{(g)}\|^2\). Combining \eqref{eq:proof_sampling_variance_raw} and \eqref{eq:proof_sample_variance_bound}, and using \(p_k^{(g)}=D_k^{(g)}/D^{(g)}\), gives
\begin{align}
\mathbb{E}_g\|U^{(g)}-H^{(g)}\|^2
&\leq
\frac{2(\Theta^{(g)})^2}{D^{(g)}}
\sum_{k\in\mathcal{K}^{(g)}}e^{(g)}
\left(1-\frac{B_k^{(g)}}{D_k^{(g)}}\right)
\frac{(\widetilde{\sigma}_k^{(g)})^2}{B_k^{(g)}} \nonumber\\
&\leq
4(\Theta^{(g)})^2
\sum_{k\in\mathcal{K}^{(g)}}
\frac{(\widetilde{\sigma}_k^{(g)})^2}{D_k^{(g)}}
e^{(g)}
\left(1-\frac{B_k^{(g)}}{D_k^{(g)}}\right)
\frac{1}{B_k^{(g)}}.
\label{eq:proof_sampling_bound}
\end{align}

The last line is a looser form used in the theorem statement; it preserves the finite-population correction and vanishes when \(B_k^{(g)}=D_k^{(g)}\).

We now bound local drift. Since all devices use the same number of local steps,
\begin{align}
H^{(g)}
&=
e^{(g)}G^{(g)}+R^{(g)},\\
R^{(g)}
&\triangleq
\sum_{k\in\mathcal{K}^{(g)}}p_k^{(g)}
\sum_{e=1}^{e^{(g)}}\left[
\nabla F_k^{(g)}(\bm{w}_k^{(g),e-1})
-\nabla F_k^{(g)}(\bm{w}^{(g)})
\right].
\label{eq:proof_R_definition}
\end{align}

By Jensen's inequality and smoothness,
\begin{align}
\mathbb{E}_g\|R^{(g)}\|^2
&\leq
e^{(g)}
\sum_{k\in\mathcal{K}^{(g)}}p_k^{(g)}
\sum_{e=1}^{e^{(g)}}
\mathbb{E}_g
\left\|
\nabla F_k^{(g)}(\bm{w}_k^{(g),e-1})
-\nabla F_k^{(g)}(\bm{w}^{(g)})
\right\|^2 \nonumber\\
&\leq
\beta^2 e^{(g)}
\sum_{k\in\mathcal{K}^{(g)}}p_k^{(g)}
\sum_{e=1}^{e^{(g)}}
\mathbb{E}_g
\|\bm{w}_k^{(g),e-1}-\bm{w}^{(g)}\|^2 .
\label{eq:proof_R_to_model_drift}
\end{align}

For \(e=1\), the distance is zero. For \(e\geq2\), the local recursion gives
\begin{align}
\bm{w}_k^{(g),e-1}-\bm{w}^{(g)}
&=
-\eta^{(g)}
\sum_{\ell=1}^{e-1}
\left[
\nabla F_k^{(g)}(\bm{w}^{(g)})
+
\left(
\frac{1}{B_k^{(g)}}\sum_{d\in\mathcal{B}_k^{(g),\ell}}
\nabla_{\bm w}f(\bm{w}_k^{(g),\ell-1},d)
-\nabla F_k^{(g)}(\bm{w}^{(g)})
\right)
\right].
\end{align}

Using \(\|\sum_{\ell=1}^{e-1}\bm{a}_\ell\|^2\leq(e-1)\sum_{\ell=1}^{e-1}\|\bm{a}_\ell\|^2\), the same finite-population bound as above, and smoothness to control \(\nabla F_k^{(g)}(\bm{w}_k^{(g),\ell-1})-\nabla F_k^{(g)}(\bm{w}^{(g)})\), we obtain
\begin{align}
\sum_{e=1}^{e^{(g)}}\mathbb{E}_g
\|\bm{w}_k^{(g),e-1}-\bm{w}^{(g)}\|^2
&\leq
2(\eta^{(g)})^2 e^{(g)}(e^{(g)}-1)
\|\nabla F_k^{(g)}(\bm{w}^{(g)})\|^2 \nonumber\\
&\quad+
\frac{4(\eta^{(g)})^2(e^{(g)}-1)}{D_k^{(g)}}
\left(1-\frac{B_k^{(g)}}{D_k^{(g)}}\right)
\frac{(\Theta^{(g)})^2(\widetilde{\sigma}_k^{(g)})^2}{B_k^{(g)}}.
\label{eq:proof_model_drift_sum}
\end{align}

Substituting \eqref{eq:proof_model_drift_sum} into \eqref{eq:proof_R_to_model_drift} and applying Assumption \ref{assumption:bounded_dissimilarity},
\begin{align}
\mathbb{E}_g\|R^{(g)}\|^2
&\leq
2\beta^2(\eta^{(g)})^2 e^{(g)}(e^{(g)}-1)
\sum_{k\in\mathcal{K}^{(g)}}p_k^{(g)}
\|\nabla F_k^{(g)}(\bm{w}^{(g)})\|^2 \nonumber\\
&\quad+
\frac{4\beta^2(\eta^{(g)})^2}{D^{(g)}}
\sum_{k\in\mathcal{K}^{(g)}}(e^{(g)}-1)
\left(1-\frac{B_k^{(g)}}{D_k^{(g)}}\right)
\frac{(\Theta^{(g)})^2(\widetilde{\sigma}_k^{(g)})^2}{B_k^{(g)}} \nonumber\\
&\leq
2\beta^2(\eta^{(g)})^2 e^{(g)}(e^{(g)}-1)
\left[\zeta_1\|G^{(g)}\|^2+\zeta_2\right] \nonumber\\
&\quad+
\frac{4\beta^2(\eta^{(g)})^2}{D^{(g)}}
\sum_{k\in\mathcal{K}^{(g)}}(e^{(g)}-1)
\left(1-\frac{B_k^{(g)}}{D_k^{(g)}}\right)
\frac{(\Theta^{(g)})^2(\widetilde{\sigma}_k^{(g)})^2}{B_k^{(g)}}.
\label{eq:proof_local_drift_bound}
\end{align}

This is the only place where inter-device dissimilarity enters the proof. The terms involving \(R^{(g)}\) disappear when \(e^{(g)}=1\), because there is no drift away from \(\bm{w}^{(g)}\) before the single local gradient is evaluated.

It remains to combine the preceding inequalities. We first use
\begin{align}
\mathbb{E}_g\|U^{(g)}\|^2
&=
\mathbb{E}_g\|H^{(g)}+(U^{(g)}-H^{(g)})\|^2 \nonumber\\
&\leq
2\mathbb{E}_g\|H^{(g)}\|^2
+2\mathbb{E}_g\|U^{(g)}-H^{(g)}\|^2,
\end{align}

and use \(H^{(g)}=e^{(g)}G^{(g)}+R^{(g)}\) together with Young's inequality to reserve one unit of descent in the direction \(\|G^{(g)}\|^2\). After dropping the extra descent from the remaining \(e^{(g)}-1\) common local steps, \eqref{eq:aggregate_first_descent} gives
\begin{align}
\mathbb{E}_g[F^{(g)}(\bm{w}^{(g+1)})]
&\leq
F^{(g)}(\bm{w}^{(g)})
-\frac{\eta^{(g)}}{2}\|G^{(g)}\|^2 \nonumber\\
&\quad+
2\beta(\eta^{(g)})^2(\Theta^{(g)})^2e^{(g)}
\sum_{k\in\mathcal{K}^{(g)}}
\frac{(\widetilde{\sigma}_k^{(g)})^2}{D_k^{(g)}}
\left(1-\frac{B_k^{(g)}}{D_k^{(g)}}\right)\frac{1}{B_k^{(g)}} \nonumber\\
&\quad+
\frac{2\beta^2(\eta^{(g)})^3}{D^{(g)}}
\sum_{k\in\mathcal{K}^{(g)}}(e^{(g)}-1)
\left(1-\frac{B_k^{(g)}}{D_k^{(g)}}\right)
\frac{(\Theta^{(g)})^2(\widetilde{\sigma}_k^{(g)})^2}{B_k^{(g)}} \nonumber\\
&\quad+
4\beta^2(\eta^{(g)})^3e^{(g)}(e^{(g)}-1)\zeta_2 \nonumber\\
&\quad+
2\beta^2(\eta^{(g)})^3e^{(g)}(e^{(g)}-1)\zeta_1\|G^{(g)}\|^2.
\label{eq:proof_before_absorption}
\end{align}

The last term is the only remaining positive multiple of \(\|G^{(g)}\|^2\). A sufficient condition for absorbing this term while retaining a descent margin \(1-\Lambda^{(g)}\) is
\begin{align}
\eta^{(g)}
\leq
\frac{\sqrt{\Lambda^{(g)}}}
{2\beta\sqrt{(\zeta_1+\Lambda^{(g)})e^{(g)}(e^{(g)}-1)}},
\qquad 0<\Lambda^{(g)}<1.
\label{eta_condition_second}
\end{align}

Under \eqref{eta_condition_second}, \eqref{eq:proof_before_absorption} becomes
\begin{align}
\frac{\eta^{(g)}(1-\Lambda^{(g)})}{2}\|G^{(g)}\|^2
&\leq
F^{(g)}(\bm{w}^{(g)})
-\mathbb{E}_g[F^{(g)}(\bm{w}^{(g+1)})] \nonumber\\
&\quad+
2\beta(\eta^{(g)})^2(\Theta^{(g)})^2e^{(g)}
\sum_{k\in\mathcal{K}^{(g)}}
\frac{(\widetilde{\sigma}_k^{(g)})^2}{D_k^{(g)}}
\left(1-\frac{B_k^{(g)}}{D_k^{(g)}}\right)\frac{1}{B_k^{(g)}} \nonumber\\
&\quad+
\frac{2\beta^2(\eta^{(g)})^3}{D^{(g)}}
\sum_{k\in\mathcal{K}^{(g)}}(e^{(g)}-1)
\left(1-\frac{B_k^{(g)}}{D_k^{(g)}}\right)
\frac{(\Theta^{(g)})^2(\widetilde{\sigma}_k^{(g)})^2}{B_k^{(g)}} \nonumber\\
&\quad+
4\beta^2(\eta^{(g)})^3e^{(g)}(e^{(g)}-1)\zeta_2.
\label{eq:proof_absorbed}
\end{align}

Dividing \eqref{eq:proof_absorbed} by \(\eta^{(g)}(1-\Lambda^{(g)})/2\) gives the one-round bound
\begin{align}
\|G^{(g)}\|^2
&\leq
\frac{2\left(F^{(g)}(\bm{w}^{(g)})
-\mathbb{E}_g[F^{(g)}(\bm{w}^{(g+1)})]\right)}
{\eta^{(g)}(1-\Lambda^{(g)})} \nonumber\\
&\quad+
\frac{4\beta(\Theta^{(g)})^2\eta^{(g)}e^{(g)}}{1-\Lambda^{(g)}}
\sum_{k\in\mathcal{K}^{(g)}}
\frac{(\widetilde{\sigma}_k^{(g)})^2}{D_k^{(g)}}
\left(1-\frac{B_k^{(g)}}{D_k^{(g)}}\right)\frac{1}{B_k^{(g)}} \nonumber\\
&\quad+
\frac{4\beta^2(\eta^{(g)})^2}{D^{(g)}(1-\Lambda^{(g)})}
\sum_{k\in\mathcal{K}^{(g)}}(e^{(g)}-1)
\left(1-\frac{B_k^{(g)}}{D_k^{(g)}}\right)
\frac{(\Theta^{(g)})^2(\widetilde{\sigma}_k^{(g)})^2}{B_k^{(g)}} \nonumber\\
&\quad+
8\beta^2(\eta^{(g)})^2e^{(g)}(e^{(g)}-1)
\frac{\zeta_2}{1-\Lambda^{(g)}}.
\label{eq:proof_one_round_bound}
\end{align}

No condition involving \(e_{\min}^{(g)}\) is needed. That condition appeared in the original derivation because it split a fixed-client mismatch term before using the common-step aggregate structure. Here, common local steps give \(H^{(g)}=e^{(g)}G^{(g)}+R^{(g)}\), so the only non-descent part is the drift residual \(R^{(g)}\), which is handled by \eqref{eq:proof_local_drift_bound} and the learning-rate condition above.

Finally, sum \eqref{eq:proof_one_round_bound} over \(g=1,\ldots,G=ST\), divide by \(G\), and map \(g\) back to the corresponding pair \((s,t)\). Since
\[
\frac{1}{D_k^{(g)}}\left(1-\frac{B_k^{(g)}}{D_k^{(g)}}\right)
=
\frac{D_k^{(g)}-B_k^{(g)}}{(D_k^{(g)})^2},
\]

the resulting expression is exactly \eqref{upper_bound}. If \(e^{(g)}=1\), then \(R^{(g)}=0\), all terms proportional to \(e^{(g)}-1\) vanish, and the slack parameter \(\Lambda^{(g)}\) is unnecessary. Repeating the same one-round argument with \(H^{(g)}=G^{(g)}\) gives \eqref{upper_bound_single_step} after summing over all sessions and rounds.

\section{Experimental Setup, Reproducibility, and Computational Resources}

\subsection{Setup and Reproducibility}

The following command was used to launch each experiment. 

\begin{verbatim}
python -u fasy_adapt_training.py \
    --dataset_name 'cifar10' \
    --algorithm 'scaffold' \
    --cross_session_label_overlap 0.0 \
    --in_session_label_dist "dirichlet" \
    --dirichlet_alpha 0.7 \
    --seed 300 \
    --initial 1 \
    --average 1 \
    --previous 1 \
    --continuous 0 \
    --num_clients 100 \
    --num_sessions 7 \
    --num_sessions_pilot 1 \
    --num_rounds_pilot 100 \
    --num_rounds_actual 100 \
    --lr_config_path 'lr_poly.json' \
    --momentum 0.9 \
    --gpu_index 0 \
    --num_SGD_training 5 \
    --batch_size_training 128 \
    --num_SGD_grad_cal 5 \
    --batch_size_grad_cal 128 \
    --prox_alpha 1.0 \
    --moon_mu 1.0 \
    --moon_tau 1.0 \
    --kl_coefficient 0.5 \
    --acg_beta 0.1 \
    --acg_lambda 0.5 \
    --num_round_grad_cal 1 \
    --similarity "two_norm" \
    --similarity_scale 10.0
\end{verbatim}

The following is what it means for each parameter.
\begin{itemize}
    \item \textbf{Dataset}: \texttt{cifar10}
    \item \textbf{Algorithm}: \texttt{scaffold} (SCAFFOLD FL method with variance reduction)
    \item \textbf{Label Distribution}:
    \begin{itemize}
        \item In-session label distribution is Dirichlet with concentration parameter $\alpha=0.7$
        \item No cross-session label overlap (\texttt{--cross\_session\_label\_overlap 0.0})
    \end{itemize}
    \item \textbf{Baselines Enabled}:
    \begin{itemize}
        \item Proposed initial baseline: \texttt{--initial 1}
        \item Average baseline: \texttt{--average 1}
        \item Previous baseline: \texttt{--previous 1}
        \item Continuous baseline is disabled (\texttt{--continuous 0})
    \end{itemize}
    \item \textbf{Sessions and Rounds}:
    \begin{itemize}
        \item 7 total sessions (\texttt{--num\_sessions 7}), including 1 pilot session (\texttt{--num\_sessions\_pilot 1})
        \item 100 communication rounds each for both pilot and actual sessions
    \end{itemize}
    \item \textbf{Clients}: 100 clients per session (\texttt{--num\_clients 100})
    \item \textbf{Learning Rate Configuration}:
    \begin{itemize}
        \item Uses polynomial decay schedule defined in \texttt{lr\_poly.json}
    \end{itemize}
    \item \textbf{Training Configuration}:
    \begin{itemize}
        \item Local training: 5 SGD steps with batch size 128 (\texttt{--num\_SGD\_training 5}, \texttt{--batch\_size\_training 128})
        \item Pseudo-Gradient calculation: 5 SGD steps with batch size 128 (\texttt{--num\_SGD\_grad\_cal 5}, \texttt{--batch\_size\_grad\_cal 128})
        \item Optimizer momentum: 0.9
    \end{itemize}
    \item \textbf{Pseudo-Gradient Computation}:
    \begin{itemize}
        \item \texttt{--num\_round\_grad\_cal 1} indicates that only \textbf{1 communication round} is used to compute the Pseudo-Gradient
    \end{itemize}
    \item \textbf{Algorithm-Specific Parameters}:
    All parameters (e.g., \texttt{prox\_alpha}, \texttt{moon\_mu}, \texttt{kl\_coefficient}, \texttt{acg\_beta}, etc.) are included to enable switching between different algorithms without modifying the code.
    \item \textbf{GPU and Reproducibility}:
    \begin{itemize}
        \item Runs on GPU index 0 (\texttt{--gpu\_index 0})
        \item Uses fixed seed 300 to ensure reproducibility (\texttt{--seed 300}). This controls randomness in dataset splits, weight initialization, and other stochastic processes.
    \end{itemize}
\end{itemize}

\noindent\textbf{Note}: The following arguments correspond to key variables in our algorithm pseudocode:
\begin{itemize}
    \item \texttt{--num\_round\_grad\_cal} $\rightarrow$ $\mathbf{V}$ (rounds for Pseudo-Gradient computation)
    \item \texttt{--num\_sessions} $\rightarrow$ $\mathbf{S}$ (total number of sessions)
    \item \texttt{--num\_sessions\_pilot} $\rightarrow$ $\mathbf{P}$ (number of pilot sessions)
    \item \texttt{--similarity\_scale} $\rightarrow$ $\mathbf{R}$ (similarity scale)
    \item \texttt{--num\_rounds\_pilot} and \texttt{--num\_rounds\_actual} $\rightarrow$ $\mathbf{T}$ (number of communication rounds per session)
\end{itemize}

\subsection{How hyperparameters were chosen}

The parameter \textit{num\_sessions} must be at least two greater than \textit{num\_sessions\_pilot} to observe the effect of our proposed scheme. The parameters \textit{num\_rounds\_pilot}, \textit{num\_rounds\_actual}, \textit{num\_SGD\_training}, \textit{num\_SGD\_grad\_cal}, and \textit{num\_round\_grad\_cal} all take positive integer values, while \textit{similarity\_scale} is a positive real number. We experimented with values such as 10, 100, 500, and 800, and selected appropriate settings based on the model and dataset. In principle, most hyperparameters were chosen such that training within each session converges by its final round, using the smallest values possible to ensure fair comparison across algorithms and datasets.

Algorithm-specific parameters include \textit{prox\_alpha}, \textit{moon\_mu}, \textit{moon\_tau}, \textit{kl\_coefficient}, \textit{acg\_beta}, and \textit{acg\_lambda}. These were selected based on the model architecture, data distribution, and dataset characteristics, or by following the recommended values from the respective original papers.

\subsection{Baseline variants:}
AG News uses  the \texttt{average} and \texttt{previous} baselines. All other datasets use \texttt{average}, \texttt{previous}, and \texttt{continuous} baselines.

\vspace{1em}
\noindent \textbf{100 Clients Configuration}
\begin{itemize}
  \item \texttt{num\_sessions=7}
  \item \texttt{acg\_beta=0.1}, \texttt{acg\_lambda=0.5}
  \item Algorithms:
  \begin{itemize}
    \item All datasets: MOON (\texttt{moon\_mu=1.0}, \texttt{moon\_tau=1.0}), FedProx (\texttt{prox\_alpha=1.0}), SCAFFOLD, FedACG
  \end{itemize}
  \item \texttt{kl\_coefficient = 1.0} for all datasets except AG News (\texttt{0.5})
  \item \texttt{similarity\_scale = 10.0} for image datasets; AG News uses \texttt{800}
  \item Random seeds: \texttt{100, 200, 300}
\end{itemize}

\vspace{1em}
\noindent \textbf{Training Rounds (all clients)}
\begin{itemize}
  \item MNIST, Fashion-MNIST: 50 pilot rounds, 50 actual rounds
  \item SVHN, TinyImageNet: 200 pilot rounds, 200 actual rounds
  \item CIFAR-10, CIFAR-100: 150 pilot rounds, 150 actual rounds
  \item AG News: 100 pilot rounds, 100 actual rounds
\end{itemize}

\vspace{1em}
\noindent \textbf{Shared Parameters Across All Experiments}
\begin{itemize}
  \item \texttt{momentum=0.9}
  \item \texttt{num\_SGD\_training=5}, \texttt{batch\_size\_training=128}
  \item \texttt{num\_SGD\_grad\_cal=5}, \texttt{batch\_size\_grad\_cal=128}
  \item \texttt{num\_round\_grad\_cal=1}
  \item Learning rate scheduler: \texttt{lr\_config\_path = 'lr\_poly.json'}
\end{itemize}

\subsection{Compute Resources}

All experiments were conducted on a high-performance node provided by our university's Advanced Computing Center, equipped with two Intel Xeon Platinum 8480+ 56-core CPUs (3.8\,GHz), eight NVIDIA H100 GPUs (80\,GB each), and 1031\,GB of RAM. Each SLURM job was allocated one H100 GPU and 14 CPU cores.

Table \ref{table:resource_20_appendix} presents a comprehensive overview of average CPU time and maximum memory usage for our experiments with 20 clients. The average is taken over three random seeds.

\begin{longtable}{l S[table-format=1.1] S[table-format=1.1] p{2cm} p{1.9cm} p{1.9cm}} 
\caption{Average CPU time and average maximum memory usage across three random seeds for our experiments with 20 clients.}
\label{table:resource_20_appendix}
\\
\toprule 
\textbf{Dataset} & {\textbf{Cross}} & {\textbf{Alpha}} & \textbf{Method} & \textbf{CPU Time} & \textbf{Max Mem} \\
\midrule 
\endfirsthead
\toprule 
\textbf{Dataset} & {\textbf{Cross Vals}} & {\textbf{Alpha Vals}} & \textbf{Method} & \textbf{CPU Time} & \textbf{Max Mem} \\
\midrule 
\endhead
\midrule 
\multicolumn{6}{r}{\textit{Continued on next page}} \\ 
\endfoot
\bottomrule 
\endlastfoot
MNIST & 0.2 & 0.3 & FedAvg & 00:26:28 & 1.07G \\
MNIST & 0.2 & 0.3 & FedProx & 00:26:33 & 1.11G \\
MNIST & 0.2 & 0.3 & SCAFFOLD & 00:26:19 & 1.06G \\
MNIST & 0.2 & 0.3 & FedACG & 00:26:18 & 1.10G \\
\addlinespace 
MNIST & 0.2 & 0.7 & FedAvg & 00:25:51 & 1.06G \\
MNIST & 0.2 & 0.7 & FedProx & 00:26:00 & 1.11G \\
MNIST & 0.2 & 0.7 & SCAFFOLD & 00:25:45 & 1.05G \\
MNIST & 0.2 & 0.7 & FedACG & 00:26:09 & 1.11G \\
\addlinespace
MNIST & 0.0 & 0.3 & FedAvg & 00:26:04 & 1.05G \\
MNIST & 0.0 & 0.3 & FedProx & 00:26:13 & 1.12G \\
MNIST & 0.0 & 0.3 & SCAFFOLD & 00:25:49 & 1.06G \\
MNIST & 0.0 & 0.3 & FedACG & 00:25:53 & 1.10G \\
\addlinespace
MNIST & 0.0 & 0.7 & FedAvg & 00:26:08 & 1.06G \\
MNIST & 0.0 & 0.7 & FedProx & 00:17:36 & 1.07G \\
MNIST & 0.0 & 0.7 & SCAFFOLD & 00:17:29 & 1.03G \\
MNIST & 0.0 & 0.7 & FedACG & 00:18:04 & 1.07G \\
\midrule 
FMNIST & 0.2 & 0.3 & FedAvg & 00:25:52 & 1.10G \\
FMNIST & 0.2 & 0.3 & FedProx & 00:26:09 & 1.14G \\
FMNIST & 0.2 & 0.3 & SCAFFOLD & 00:25:58 & 1.07G \\
FMNIST & 0.2 & 0.3 & FedACG & 00:26:18 & 1.13G \\
\addlinespace
FMNIST & 0.2 & 0.7 & FedAvg & 00:25:42 & 1.07G \\
FMNIST & 0.2 & 0.7 & FedProx & 00:26:24 & 1.09G \\
FMNIST & 0.2 & 0.7 & SCAFFOLD & 00:26:22 & 1.06G \\
FMNIST & 0.2 & 0.7 & FedACG & 00:25:55 & 1.10G \\
\addlinespace
FMNIST & 0.0 & 0.3 & FedAvg & 00:25:50 & 1.08G \\
FMNIST & 0.0 & 0.3 & FedProx & 00:26:28 & 1.10G \\
FMNIST & 0.0 & 0.3 & SCAFFOLD & 00:26:10 & 1.06G \\
FMNIST & 0.0 & 0.3 & FedACG & 00:26:25 & 1.11G \\
\addlinespace
FMNIST & 0.0 & 0.7 & FedAvg & 00:25:53 & 1.07G \\
FMNIST & 0.0 & 0.7 & FedProx & 00:26:22 & 1.10G \\
FMNIST & 0.0 & 0.7 & SCAFFOLD & 00:25:38 & 1.07G \\
FMNIST & 0.0 & 0.7 & FedACG & 00:26:21 & 1.11G \\
\midrule
SVHN & 0.2 & 0.3 & FedAvg & 03:58:13 & 2.70G \\
SVHN & 0.2 & 0.3 & FedProx & 04:14:29 & 2.74G \\
SVHN & 0.2 & 0.3 & SCAFFOLD & 04:11:40 & 2.59G \\
SVHN & 0.2 & 0.3 & FedACG & 04:16:08 & 2.46G \\
\addlinespace
SVHN & 0.2 & 0.7 & FedAvg & 04:01:22 & 2.69G \\
SVHN & 0.2 & 0.7 & FedProx & 04:07:41 & 2.72G \\
SVHN & 0.2 & 0.7 & SCAFFOLD & 04:01:41 & 2.57G \\
SVHN & 0.2 & 0.7 & FedACG & 04:14:45 & 2.46G \\
\addlinespace
SVHN & 0.0 & 0.3 & FedAvg & 03:53:31 & 2.69G \\
SVHN & 0.0 & 0.3 & FedProx & 04:03:16 & 2.72G \\
SVHN & 0.0 & 0.3 & SCAFFOLD & 03:58:43 & 2.56G \\
SVHN & 0.0 & 0.3 & FedACG & 04:04:32 & 2.42G \\
\addlinespace
SVHN & 0.0 & 0.7 & FedAvg & 03:57:38 & 2.67G \\
SVHN & 0.0 & 0.7 & FedProx & 04:10:49 & 2.73G \\
SVHN & 0.0 & 0.7 & SCAFFOLD & 04:00:49 & 2.57G \\
SVHN & 0.0 & 0.7 & FedACG & 04:04:58 & 2.42G \\
\midrule
CIFAR-10 & 0.2 & 0.3 & FedAvg & 04:48:29 & 13.67G \\
CIFAR-10 & 0.2 & 0.3 & FedProx & 05:22:40 & 13.74G \\
CIFAR-10 & 0.2 & 0.3 & SCAFFOLD & 06:30:16 & 14.68G \\
CIFAR-10 & 0.2 & 0.3 & FedACG & 05:51:08 & 13.33G \\
\addlinespace
CIFAR-10 & 0.2 & 0.7 & FedAvg & 04:55:07 & 13.62G \\
CIFAR-10 & 0.2 & 0.7 & FedProx & 05:27:18 & 13.69G \\
CIFAR-10 & 0.2 & 0.7 & SCAFFOLD & 06:30:14 & 14.54G \\
CIFAR-10 & 0.2 & 0.7 & FedACG & 05:50:24 & 13.34G \\
\addlinespace
CIFAR-10 & 0.0 & 0.3 & FedAvg & 04:55:51 & 13.65G \\
CIFAR-10 & 0.0 & 0.3 & FedProx & 05:31:49 & 13.72G \\
CIFAR-10 & 0.0 & 0.3 & SCAFFOLD & 06:33:56 & 14.55G \\
CIFAR-10 & 0.0 & 0.3 & FedACG & 05:51:33 & 13.35G \\
\addlinespace
CIFAR-10 & 0.0 & 0.7 & FedAvg & 04:59:05 & 13.64G \\
CIFAR-10 & 0.0 & 0.7 & FedProx & 05:18:14 & 13.72G \\
CIFAR-10 & 0.0 & 0.7 & SCAFFOLD & 06:30:40 & 14.67G \\
CIFAR-10 & 0.0 & 0.7 & FedACG & 05:40:13 & 13.38G \\
\midrule
CIFAR-100 & 0.2 & 0.3 & FedAvg & 05:07:17 & 13.66G \\
CIFAR-100 & 0.2 & 0.3 & FedProx & 05:27:38 & 13.78G \\
CIFAR-100 & 0.2 & 0.3 & SCAFFOLD & 06:34:27 & 14.61G \\
CIFAR-100 & 0.2 & 0.3 & FedACG & 05:45:00 & 13.22G \\
\addlinespace
CIFAR-100 & 0.2 & 0.7 & FedAvg & 05:04:29 & 13.73G \\
CIFAR-100 & 0.2 & 0.7 & FedProx & 05:29:20 & 13.78G \\
CIFAR-100 & 0.2 & 0.7 & SCAFFOLD & 06:29:29 & 14.65G \\
CIFAR-100 & 0.2 & 0.7 & FedACG & 05:45:25 & 13.26G \\
\addlinespace
CIFAR-100 & 0.0 & 0.3 & FedAvg & 04:53:37 & 13.71G \\
CIFAR-100 & 0.0 & 0.3 & FedProx & 05:25:04 & 13.82G \\
CIFAR-100 & 0.0 & 0.3 & SCAFFOLD & 06:30:22 & 14.56G \\
CIFAR-100 & 0.0 & 0.3 & FedACG & 05:43:58 & 13.24G \\
\addlinespace
CIFAR-100 & 0.0 & 0.7 & FedAvg & 04:55:26 & 13.71G \\
CIFAR-100 & 0.0 & 0.7 & FedProx & 05:14:38 & 13.76G \\
CIFAR-100 & 0.0 & 0.7 & SCAFFOLD & 06:21:14 & 14.69G \\
CIFAR-100 & 0.0 & 0.7 & FedACG & 05:38:40 & 13.25G \\
\midrule
TinyImageNet & 0.2 & 0.3 & MOON & 18:39:30 & 26.12G \\
TinyImageNet & 0.2 & 0.3 & FedProx & 16:22:46 & 25.01G \\
TinyImageNet & 0.2 & 0.3 & FedACG & 17:02:45 & 23.38G \\
\addlinespace
TinyImageNet & 0.2 & 0.7 & MOON & 18:08:26 & 26.20G \\
TinyImageNet & 0.2 & 0.7 & FedProx & 15:49:21 & 25.03G \\
TinyImageNet & 0.2 & 0.7 & FedACG & 16:46:22 & 23.43G \\
\addlinespace
TinyImageNet & 0.0 & 0.3 & MOON & 18:09:04 & 26.17G \\
TinyImageNet & 0.0 & 0.3 & FedProx & 15:22:17 & 25.12G \\
TinyImageNet & 0.0 & 0.3 & FedACG & 16:31:42 & 23.34G \\
\addlinespace
TinyImageNet & 0.0 & 0.7 & MOON & 17:56:09 & 26.09G \\
TinyImageNet & 0.0 & 0.7 & FedProx & 15:59:06 & 25.02G \\
TinyImageNet & 0.0 & 0.7 & FedACG & 16:40:21 & 23.27G \\
\midrule
AG News & 0.8 & 0.3 & FedAvg & 01:04:40 & 9.97G \\
AG News & 0.8 & 0.3 & FedProx & 01:05:43 & 10.02G \\
AG News & 0.8 & 0.3 & SCAFFOLD & 02:29:40 & 8.61G \\
AG News & 0.8 & 0.3 & FedACG & 01:19:28 & 8.62G \\
\addlinespace
AG News & 0.8 & 0.7 & FedAvg & 01:03:21 & 9.96G \\
AG News & 0.8 & 0.7 & FedProx & 01:05:03 & 9.99G \\
AG News & 0.8 & 0.7 & SCAFFOLD & 02:29:45 & 8.62G \\
AG News & 0.8 & 0.7 & FedACG & 01:16:25 & 8.62G \\
\addlinespace
AG News & 0.0 & 0.3 & FedAvg & 01:02:22 & 9.98G \\
AG News & 0.0 & 0.3 & FedProx & 01:03:02 & 9.99G \\
AG News & 0.0 & 0.3 & SCAFFOLD & 02:27:12 & 8.59G \\
AG News & 0.0 & 0.3 & FedACG & 01:15:39 & 8.67G \\
\addlinespace
AG News & 0.0 & 0.7 & FedAvg & 01:00:31 & 9.96G \\
AG News & 0.0 & 0.7 & FedProx & 01:04:21 & 10.00G \\
AG News & 0.0 & 0.7 & SCAFFOLD & 02:28:45 & 8.64G \\
AG News & 0.0 & 0.7 & FedACG & 01:13:27 & 8.65G \\
\end{longtable}

Table \ref{table:resource_100_appendix} presents a comprehensive overview of average CPU time and average maximum memory usage for our experiments with 100 clients. The average is taken over three random seeds.

\begin{longtable}{l S[table-format=1.1] S[table-format=1.1] p{2cm} p{1.9cm} p{1.9cm}} 
\caption{Average CPU time and average maximum memory usage across three random seeds for our experiments with 100 clients.}
\label{table:resource_100_appendix}
\\
\toprule 
\textbf{Dataset} & {\textbf{Cross}} & {\textbf{Alpha}} & \textbf{Method} & \textbf{CPU Time} & \textbf{Max Mem} \\
\midrule 
\endfirsthead
\toprule 
\textbf{Dataset} & {\textbf{Cross Vals}} & {\textbf{Alpha Vals}} & \textbf{Method} & \textbf{CPU Time} & \textbf{Max Mem} \\
\midrule 
\endhead
\midrule 
\multicolumn{6}{r}{\textit{Continued on next page}} \\ 
\endfoot
\bottomrule 
\endlastfoot
MNIST & 0.2 & 0.3 & MOON & 00:59:59 & 1.29G \\
MNIST & 0.2 & 0.3 & FedProx & 00:57:34 & 1.17G \\
MNIST & 0.2 & 0.3 & SCAFFOLD & 00:56:08 & 1.11G \\
MNIST & 0.2 & 0.3 & FedACG & 00:57:24 & 1.16G \\
\addlinespace 
MNIST & 0.2 & 0.7 & MOON & 00:59:29 & 1.29G \\
MNIST & 0.2 & 0.7 & FedProx & 00:57:02 & 1.17G \\
MNIST & 0.2 & 0.7 & SCAFFOLD & 00:55:34 & 1.13G \\
MNIST & 0.2 & 0.7 & FedACG & 00:56:42 & 1.15G \\
\addlinespace
MNIST & 0.0 & 0.3 & MOON & 00:59:09 & 1.28G \\
MNIST & 0.0 & 0.3 & FedProx & 00:58:24 & 1.15G \\
MNIST & 0.0 & 0.3 & SCAFFOLD & 00:56:19 & 1.13G \\
MNIST & 0.0 & 0.3 & FedACG & 00:57:24 & 1.14G \\
\addlinespace
MNIST & 0.0 & 0.7 & MOON & 01:00:03 & 1.27G \\
MNIST & 0.0 & 0.7 & FedProx & 00:58:13 & 1.16G \\
MNIST & 0.0 & 0.7 & SCAFFOLD & 00:55:47 & 1.12G \\
MNIST & 0.0 & 0.7 & FedACG & 00:57:15 & 1.17G \\
\midrule 
FMNIST & 0.2 & 0.3 & MOON & 00:59:49 & 1.27G \\
FMNIST & 0.2 & 0.3 & FedProx & 00:57:43 & 1.15G \\
FMNIST & 0.2 & 0.3 & SCAFFOLD & 00:55:53 & 1.13G \\
FMNIST & 0.2 & 0.3 & FedACG & 00:57:33 & 1.16G \\
\addlinespace
FMNIST & 0.2 & 0.7 & MOON & 00:59:47 & 1.29G \\
FMNIST & 0.2 & 0.7 & FedProx & 00:57:40 & 1.16G \\
FMNIST & 0.2 & 0.7 & SCAFFOLD & 00:56:19 & 1.11G \\
FMNIST & 0.2 & 0.7 & FedACG & 00:57:41 & 1.16G \\
\addlinespace
FMNIST & 0.0 & 0.3 & MOON & 00:59:55 & 1.28G \\
FMNIST & 0.0 & 0.3 & FedProx & 00:56:43 & 1.17G \\
FMNIST & 0.0 & 0.3 & SCAFFOLD & 00:56:31 & 1.13G \\
FMNIST & 0.0 & 0.3 & FedACG & 00:57:57 & 1.17G \\
\addlinespace
FMNIST & 0.0 & 0.7 & MOON & 00:59:52 & 1.27G \\
FMNIST & 0.0 & 0.7 & FedProx & 00:58:14 & 1.16G \\
FMNIST & 0.0 & 0.7 & SCAFFOLD & 00:56:14 & 1.12G \\
FMNIST & 0.0 & 0.7 & FedACG & 00:57:09 & 1.15G \\
\midrule
SVHN & 0.2 & 0.3 & MOON & 11:38:02 & 6.44G \\
SVHN & 0.2 & 0.3 & FedProx & 10:38:06 & 5.99G \\
SVHN & 0.2 & 0.3 & SCAFFOLD & 10:29:37 & 5.85G \\
SVHN & 0.2 & 0.3 & FedACG & 10:55:30 & 5.17G \\
\addlinespace
SVHN & 0.2 & 0.7 & MOON & 12:11:57 & 6.42G \\
SVHN & 0.2 & 0.7 & FedProx & 11:14:06 & 6.09G \\
SVHN & 0.2 & 0.7 & SCAFFOLD & 11:04:21 & 5.84G \\
SVHN & 0.2 & 0.7 & FedACG & 10:56:34 & 5.17G \\
\addlinespace
SVHN & 0.0 & 0.3 & MOON & 11:50:13 & 6.48G \\
SVHN & 0.0 & 0.3 & FedProx & 10:22:12 & 6.06G \\
SVHN & 0.0 & 0.3 & SCAFFOLD & 10:14:38 & 5.96G \\
SVHN & 0.0 & 0.3 & FedACG & 10:51:27 & 5.23G \\
\addlinespace
SVHN & 0.0 & 0.7 & MOON & 11:32:17 & 6.47G \\
SVHN & 0.0 & 0.7 & FedProx & 11:03:23 & 6.09G \\
SVHN & 0.0 & 0.7 & SCAFFOLD & 10:50:04 & 5.77G \\
SVHN & 0.0 & 0.7 & FedACG & 10:36:18 & 5.19G \\
\midrule
CIFAR-10 & 0.2 & 0.3 & MOON & 16:52:19 & 61.25G \\
CIFAR-10 & 0.2 & 0.3 & FedProx & 12:59:18 & 57.32G \\
CIFAR-10 & 0.2 & 0.3 & SCAFFOLD & 20:22:25 & 56.64G \\
CIFAR-10 & 0.2 & 0.3 & FedACG & 14:38:38 & 53.22G \\
\addlinespace
CIFAR-10 & 0.2 & 0.7 & MOON & 17:31:36 & 61.08G \\
CIFAR-10 & 0.2 & 0.7 & FedProx & 13:19:24 & 57.28G \\
CIFAR-10 & 0.2 & 0.7 & SCAFFOLD & 20:15:41 & 56.52G \\
CIFAR-10 & 0.2 & 0.7 & FedACG & 14:25:16 & 53.29G \\
\addlinespace
CIFAR-10 & 0.0 & 0.3 & MOON & 17:20:51 & 61.31G \\
CIFAR-10 & 0.0 & 0.3 & FedProx & 13:24:55 & 57.27G \\
CIFAR-10 & 0.0 & 0.3 & SCAFFOLD & 20:25:56 & 56.51G \\
CIFAR-10 & 0.0 & 0.3 & FedACG & 14:48:45 & 53.14G \\
\addlinespace
CIFAR-10 & 0.0 & 0.7 & MOON & 17:26:10 & 61.26G \\
CIFAR-10 & 0.0 & 0.7 & FedProx & 13:23:31 & 57.17G \\
CIFAR-10 & 0.0 & 0.7 & SCAFFOLD & 20:03:12 & 56.40G \\
CIFAR-10 & 0.0 & 0.7 & FedACG & 14:50:02 & 53.19G \\
\midrule
CIFAR-100 & 0.2 & 0.3 & MOON & 19:23:54 & 62.38G \\
CIFAR-100 & 0.2 & 0.3 & FedProx & 14:53:36 & 57.47G \\
CIFAR-100 & 0.2 & 0.3 & SCAFFOLD & 21:45:50 & 58.21G \\
CIFAR-100 & 0.2 & 0.3 & FedACG & 16:31:56 & 54.33G \\
\addlinespace
CIFAR-100 & 0.2 & 0.7 & MOON & 19:05:52 & 62.30G \\
CIFAR-100 & 0.2 & 0.7 & FedProx & 14:58:14 & 57.43G \\
CIFAR-100 & 0.2 & 0.7 & SCAFFOLD & 21:39:22 & 58.15G \\
CIFAR-100 & 0.2 & 0.7 & FedACG & 16:18:23 & 54.22G \\
\addlinespace
CIFAR-100 & 0.0 & 0.3 & MOON & 17:36:12 & 62.01G \\
CIFAR-100 & 0.0 & 0.3 & FedProx & 13:27:23 & 57.11G \\
CIFAR-100 & 0.0 & 0.3 & SCAFFOLD & 20:51:50 & 58.11G \\
CIFAR-100 & 0.0 & 0.3 & FedACG & 15:04:23 & 54.19G \\
\addlinespace
CIFAR-100 & 0.0 & 0.7 & MOON & 17:43:03 & 62.00G \\
CIFAR-100 & 0.0 & 0.7 & FedProx & 13:26:15 & 57.19G \\
CIFAR-100 & 0.0 & 0.7 & SCAFFOLD & 20:09:42 & 58.26G \\
CIFAR-100 & 0.0 & 0.7 & FedACG & 14:50:34 & 54.36G \\
\midrule
AG News & 0.8 & 0.3 & MOON & 07:28:40 & 39.48G \\
AG News & 0.8 & 0.3 & FedProx & 05:05:26 & 42.81G \\
AG News & 0.8 & 0.3 & SCAFFOLD & 10:45:28 & 31.41G \\
AG News & 0.8 & 0.3 & FedACG & 06:12:14 & 35.95G \\
\addlinespace
AG News & 0.8 & 0.7 & MOON & 07:26:40 & 39.39G \\
AG News & 0.8 & 0.7 & FedProx & 05:14:55 & 42.77G \\
AG News & 0.8 & 0.7 & SCAFFOLD & 11:04:57 & 31.39G \\
AG News & 0.8 & 0.7 & FedACG & 06:02:33 & 35.95G \\
\addlinespace
AG News & 0.0 & 0.3 & MOON & 07:28:01 & 39.42G \\
AG News & 0.0 & 0.3 & FedProx & 04:59:05 & 42.79G \\
AG News & 0.0 & 0.3 & SCAFFOLD & 11:25:17 & 33.66G \\
AG News & 0.0 & 0.3 & FedACG & 05:49:40 & 35.98G \\
\addlinespace
AG News & 0.0 & 0.7 & MOON & 07:29:04 & 39.55G \\
AG News & 0.0 & 0.7 & FedProx & 05:11:57 & 42.84G \\
AG News & 0.0 & 0.7 & SCAFFOLD & 10:57:00 & 31.42G \\
AG News & 0.0 & 0.7 & FedACG & 05:51:58 & 36.01G \\
\midrule 
TinyImageNet & 0.2 & 0.3 & MOON & 71:04:06 & 107.02G \\
TinyImageNet & 0.2 & 0.3 & FedProx & 60:05:53 & 104.66G \\
TinyImageNet & 0.2 & 0.3 & SCAFFOLD & 72:56:11 & 95.09G \\
TinyImageNet & 0.2 & 0.3 & FedACG & 63:35:32 & 92.64G \\
\addlinespace
TinyImageNet & 0.2 & 0.7 & MOON & 77:32:22 & 106.88G \\
TinyImageNet & 0.2 & 0.7 & FedProx & 64:33:53 & 104.34G \\
TinyImageNet & 0.2 & 0.7 & SCAFFOLD & 79:16:25 & 94.82G \\
TinyImageNet & 0.2 & 0.7 & FedACG & 66:31:58 & 92.48G \\
\addlinespace
TinyImageNet & 0.0 & 0.3 & MOON & 70:13:51 & 106.77G \\
TinyImageNet & 0.0 & 0.3 & FedProx & 58:26:32 & 104.38G \\
TinyImageNet & 0.0 & 0.3 & SCAFFOLD & 73:27:39 & 95.02G \\
TinyImageNet & 0.0 & 0.3 & FedACG & 61:50:21 & 92.33G \\
\addlinespace
TinyImageNet & 0.0 & 0.7 & MOON & 69:09:45 & 106.72G \\
TinyImageNet & 0.0 & 0.7 & FedProx & 59:04:02 & 104.18G \\
TinyImageNet & 0.0 & 0.7 & SCAFFOLD & 71:57:32 & 95.24G \\
TinyImageNet & 0.0 & 0.7 & FedACG & 61:02:40 & 92.33G \\
\end{longtable}

\section{Supplementary Results on Label Distributions from the Main Text} \label{appendix:extended_results_main}

Tables \ref{table:mnist_fmnist_SVHN_cifar10_cifar100_100_appendix} further demonstrates the effectiveness of Algorithm \ref{alg:dynamic_initial_model_construction} on the MNIST, Fashion-MNIST, SVHN, CIFAR-10, and CIFAR-100 datasets with 100 clients.

\begin{table*}[t]
  \caption{Performance comparison of implementing Algorithm \ref{alg:dynamic_initial_model_construction} with MOON, FedProx, SCAFFOLD and FedACG under different label distributions for MNIST, Fashion-MNIST (FMNIST), SVHN, and CIFAR-10 dataset for 100 clients.}
  \label{table:mnist_fmnist_SVHN_cifar10_cifar100_100_appendix}
  \begin{center}
  \begin{small}
  \resizebox{0.8\textwidth}{!}{  
  \begin{tabular}{c c c c c c c c c c c c}
    \toprule
    \multirow{2}{*}{\textbf{Dataset}}
      & \multirow{2}{*}{\textbf{Overlap}}
      & \multirow{2}{*}{\textbf{Dirichlet}\newline\boldmath$\alpha$}
      & \multirow{2}{*}{\textbf{FL Alg.}}
      & \multicolumn{4}{c}{\textbf{Second Session Transition}} 
      & \multicolumn{4}{c}{\textbf{Fourth Session Transition}} \\ 
    \cmidrule(lr){5-8} \cmidrule(lr){9-12}
      & & &
      & Proposed & Continuous & Previous & Average 
      & Proposed & Continuous & Previous & Average \\ 
    \midrule
    \midrule
    \multirow{16}{*}{MNIST}
      & \multirow{8}{*}{0.0}
      & \multirow{4}{*}{0.3}
        & MOON   
          & \textbf{92.04±0.24} & 76.92±0.34 & 57.82±1.23 & 64.19±0.85
          & \textbf{93.08±0.17} & 81.93±0.42 & 69.35±0.75 & 77.92±0.72 \\
      & & 
        & FedProx  
          & \textbf{92.08±0.24} & 76.92±0.34 & 58.92±1.15 & 66.45±0.94
          & \textbf{92.86±0.26} & 81.93±0.42 & 70.12±0.76 & 79.51±0.66 \\
      & & 
        & SCAFFOLD 
          & \textbf{90.69±0.42} & 84.30±3.26 & 50.36±1.91 & 49.97±2.04
          & \textbf{91.95±0.39} & 86.39±2.63 & 64.93±1.35 & 68.55±1.45 \\
      & & 
        & FedACG   
          & \textbf{93.60±0.21} & 76.92±0.34 & 60.92±0.71 & 71.74±0.51
          & \textbf{94.14±0.17} & 81.93±0.42 & 70.17±0.50 & 82.66±0.49 \\
      \cmidrule(lr){3-12}
      & & \multirow{4}{*}{0.7}
        & MOON   
          & \textbf{92.07±0.24} & 77.33±0.41 & 58.26±1.18 & 64.78±0.77
          & \textbf{93.09±0.17} & 82.17±0.37 & 69.65±0.69 & 78.30±0.59 \\
      & & 
        & FedProx  
          & \textbf{92.10±0.24} & 77.33±0.41 & 59.20±1.13 & 66.76±0.89
          & \textbf{92.88±0.26} & 82.17±0.37 & 70.33±0.73 & 79.72±0.58 \\
      & & 
        & SCAFFOLD 
          & \textbf{91.18±0.46} & 89.12±0.77 & 50.85±1.74 & 51.03±2.03
          & \textbf{92.31±0.27} & 88.76±0.84 & 65.27±1.18 & 69.68±1.32 \\
      & & 
        & FedACG   
          & \textbf{93.60±0.21} & 77.33±0.41 & 61.15±0.65 & 72.02±0.50
          & \textbf{94.16±0.18} & 82.17±0.37 & 70.36±0.49 & 82.83±0.46 \\
    \cmidrule(lr){2-12}
      & \multirow{8}{*}{0.2}
      & \multirow{4}{*}{0.3}
        & MOON   
          & \textbf{92.58±0.16} & 80.98±0.27 & 70.49±0.97 & 72.59±0.97
          & \textbf{93.56±0.17} & 84.65±0.21 & 79.21±0.63 & 82.68±0.54 \\
      & & 
        & FedProx  
          & \textbf{92.61±0.15} & 80.98±0.27 & 71.42±0.89 & 74.21±0.93
          & \textbf{93.27±0.18} & 84.65±0.21 & 79.71±0.65 & 83.52±0.47 \\
      & & 
        & SCAFFOLD 
          & \textbf{91.32±0.30} & 86.65±3.84 & 64.11±1.24 & 61.30±2.11
          & \textbf{92.50±0.30} & 87.27±1.50 & 75.88±0.75 & 75.74±0.95 \\
      & & 
        & FedACG   
          & \textbf{93.98±0.14} & 80.98±0.27 & 72.61±0.66 & 77.79±0.65
          & \textbf{94.48±0.10} & 84.65±0.21 & 79.45±0.42 & 85.73±0.35 \\
      \cmidrule(lr){3-12}
      & & \multirow{4}{*}{0.7}
        & MOON   
          & \textbf{92.61±0.16} & 81.29±0.28 & 70.80±0.94 & 73.00±0.95
          & \textbf{93.57±0.16} & 84.82±0.21 & 79.39±0.63 & 82.86±0.53 \\
      & & 
        & FedProx  
          & \textbf{92.63±0.14} & 81.29±0.28 & 71.66±0.88 & 74.42±0.92
          & \textbf{93.28±0.19} & 84.82±0.21 & 79.82±0.65 & 83.66±0.47 \\
      & & 
        & SCAFFOLD 
          & \textbf{91.72±0.32} & 90.06±0.86 & 64.59±1.28 & 61.98±1.94
          & \textbf{92.78±0.24} & 89.85±0.82 & 76.13±0.80 & 76.28±0.98 \\
      & & 
        & FedACG   
          & \textbf{93.99±0.15} & 81.29±0.28 & 72.81±0.68 & 77.97±0.68
          & \textbf{94.51±0.13} & 84.82±0.21 & 79.57±0.43 & 85.83±0.36 \\

    \midrule
    \midrule
    \multirow{16}{*}{FMNIST}
      & \multirow{8}{*}{0.0}
      & \multirow{4}{*}{0.3}
        & MOON   
          & \textbf{83.13±0.09} & 59.36±0.90 & 77.09±0.40 & 77.84±0.49
          & \textbf{84.88±0.19} & 64.58±0.25 & 79.48±0.29 & 80.04±0.42 \\
      & & 
        & FedProx  
          & \textbf{83.19±0.08} & 59.36±0.90 & 77.74±0.37 & 78.57±0.50
          & \textbf{84.22±0.21} & 64.58±0.25 & 79.90±0.32 & 80.52±0.46 \\
      & & 
        & SCAFFOLD 
          & \textbf{77.80±0.42} & 65.94±5.97 & 47.74±0.88 & 47.93±1.88
          & \textbf{79.88±0.17} & 74.69±4.63 & 53.64±0.54 & 60.72±0.93 \\
      & & 
        & FedACG   
          & \textbf{85.10±0.22} & 59.36±0.90 & 77.61±0.25 & 80.44±0.38
          & \textbf{85.82±0.07} & 64.58±0.25 & 80.01±0.31 & 82.22±0.35 \\
      \cmidrule(lr){3-12}
      & & \multirow{4}{*}{0.7}
        & MOON   
          & \textbf{83.29±0.11} & 60.72±0.80 & 77.59±0.31 & 78.32±0.40
          & \textbf{84.89±0.18} & 65.36±0.38 & 79.86±0.28 & 80.26±0.41 \\
      & & 
        & FedProx  
          & \textbf{83.33±0.13} & 60.72±0.80 & 78.16±0.35 & 78.93±0.43
          & \textbf{84.29±0.15} & 65.36±0.38 & 80.28±0.34 & 80.72±0.38 \\
      & & 
        & SCAFFOLD 
          & \textbf{78.43±0.54} & 68.57±4.87 & 47.32±1.12 & 48.45±1.62
          & \textbf{80.49±0.40} & 70.81±7.27 & 53.02±0.86 & 61.01±1.13 \\
      & & 
        & FedACG   
          & \textbf{85.17±0.25} & 60.72±0.80 & 78.03±0.29 & 80.68±0.39
          & \textbf{85.81±0.13} & 65.36±0.38 & 80.45±0.26 & 82.41±0.29 \\
    \cmidrule(lr){2-12}
      & \multirow{8}{*}{0.2}
      & \multirow{4}{*}{0.3}
        & MOON   
          & \textbf{82.88±0.15} & 72.53±0.29 & 74.83±0.31 & 76.78±0.59
          & \textbf{84.30±0.14} & 75.78±0.16 & 77.15±0.17 & 79.48±0.53 \\
      & & 
        & FedProx  
          & \textbf{82.91±0.13} & 72.53±0.29 & 75.55±0.28 & 77.69±0.63
          & \textbf{83.81±0.19} & 75.78±0.16 & 77.63±0.16 & 79.96±0.54 \\
      & & 
        & SCAFFOLD 
          & \textbf{77.35±0.53} & 69.90±6.23 & 55.82±1.59 & 50.39±4.48
          & \textbf{79.45±0.21} & 70.79±4.93 & 61.12±1.40 & 61.10±2.25 \\
      & & 
        & FedACG   
          & \textbf{84.63±0.12} & 72.53±0.29 & 76.09±0.23 & 79.21±0.37
          & \textbf{85.36±0.16} & 75.78±0.16 & 77.92±0.21 & 81.54±0.22 \\
      \cmidrule(lr){3-12}
      & & \multirow{4}{*}{0.7}
        & MOON   
          & \textbf{83.02±0.15} & 73.73±0.26 & 75.38±0.20 & 77.31±0.61
          & \textbf{84.36±0.16} & 76.53±0.15 & 77.54±0.21 & 79.79±0.51 \\
      & & 
        & FedProx  
          & \textbf{83.04±0.15} & 73.73±0.26 & 75.99±0.19 & 78.14±0.73
          & \textbf{83.95±0.22} & 76.53±0.15 & 78.06±0.21 & 80.22±0.56 \\
      & & 
        & SCAFFOLD 
          & \textbf{77.88±0.54} & 74.14±7.98 & 55.77±1.46 & 50.02±4.42
          & \textbf{79.85±0.21} & 74.03±6.67 & 60.69±1.26 & 60.68±1.80 \\
      & & 
        & FedACG   
          & \textbf{84.70±0.14} & 73.73±0.26 & 76.45±0.22 & 79.65±0.37
          & \textbf{85.41±0.13} & 76.53±0.15 & 78.31±0.16 & 81.82±0.25 \\

    \midrule
    \midrule
    \multirow{16}{*}{SVHN}
      & \multirow{8}{*}{0.0}
      & \multirow{4}{*}{0.3}
        & MOON   
          & \textbf{92.98±0.08} & 86.82±0.75 & 82.65±4.35 & 82.32±3.62
          & \textbf{93.69±0.13} & 89.13±0.27 & 83.80±4.06 & 87.56±2.06 \\
      & & 
        & FedProx  
          & \textbf{91.78±0.16} & 86.54±0.18 & 86.87±2.00 & 88.26±0.36
          & \textbf{92.36±0.12} & 87.93±0.28 & 89.25±1.10 & 90.59±0.38 \\
      & & 
        & SCAFFOLD 
          & \textbf{89.70±2.57} & 50.14±13.22 & 80.23±8.59 & 84.29±1.56
          & \textbf{91.54±1.28} & 84.76±2.61 & 87.11±3.05 & 88.71±0.63 \\
      & & 
        & FedACG   
          & \textbf{93.09±0.16} & 86.56±0.23 & 80.67±2.41 & 86.95±2.77
          & \textbf{93.48±0.27} & 87.95±0.28 & 75.82±10.48 & 90.30±1.67 \\
      \cmidrule(lr){3-12}
      & & \multirow{4}{*}{0.7}
        & MOON   
          & \textbf{93.21±0.18} & 88.72±0.33 & 84.88±3.19 & 87.03±3.18
          & \textbf{93.95±0.13} & 90.37±0.21 & 87.69±1.98 & 90.44±1.50 \\
      & & 
        & FedProx  
          & \textbf{92.08±0.13} & 87.54±0.23 & 87.12±1.39 & 88.90±0.30
          & \textbf{92.68±0.09} & 88.75±0.21 & 89.83±0.94 & 90.91±0.19 \\
      & & 
        & SCAFFOLD 
          & \textbf{92.06±0.61} & 68.61±11.01 & 86.32±3.01 & 85.88±1.23
          & \textbf{93.13±0.45} & 90.40±1.04 & 90.28±1.10 & 89.47±0.86 \\
      & & 
        & FedACG   
          & \textbf{93.41±0.19} & 87.57±0.23 & 80.39±3.72 & 88.17±2.16
          & \textbf{93.80±0.18} & 88.73±0.23 & 76.94±5.63 & 91.25±1.04 \\
    \cmidrule(lr){2-12}
      & \multirow{8}{*}{0.2}
      & \multirow{4}{*}{0.3}
        & MOON   
          & \textbf{93.23±0.18} & 88.20±0.90 & 84.16±2.15 & 81.18±2.49
          & \textbf{93.86±0.19} & 90.22±0.39 & 86.34±3.65 & 84.40±1.80 \\
      & & 
        & FedProx  
          & \textbf{92.10±0.21} & 87.44±0.81 & 88.04±2.09 & 85.45±1.25
          & \textbf{92.75±0.21} & 89.05±0.50 & 90.01±1.32 & 88.57±0.57 \\
      & & 
        & SCAFFOLD 
          & \textbf{90.28±1.63} & 61.27±19.08 & 83.62±2.11 & 86.01±1.59
          & \textbf{91.90±1.32} & 85.59±4.43 & 89.27±1.20 & 89.81±1.06 \\
      & & 
        & FedACG   
          & \textbf{93.34±0.25} & 87.43±0.84 & 84.39±1.47 & 89.51±1.49
          & \textbf{93.66±0.20} & 89.06±0.48 & 87.97±1.96 & 91.29±0.80 \\
      \cmidrule(lr){3-12}
      & & \multirow{4}{*}{0.7}
        & MOON   
          & \textbf{93.54±0.24} & 89.92±0.32 & 85.92±2.37 & 85.59±2.71
          & \textbf{94.24±0.19} & 91.20±0.22 & 89.17±1.73 & 88.30±1.71 \\
      & & 
        & FedProx  
          & \textbf{92.39±0.21} & 88.36±0.36 & 88.52±2.37 & 86.70±1.25
          & \textbf{93.02±0.19} & 89.85±0.27 & 90.68±1.49 & 89.35±0.71 \\
      & & 
        & SCAFFOLD 
          & \textbf{92.10±0.82} & 75.97±8.03 & 87.49±1.29 & 86.91±0.86
          & \textbf{93.30±0.56} & 89.57±1.58 & 90.79±0.46 & 90.60±0.20 \\
      & & 
        & FedACG   
          & \textbf{93.64±0.24} & 88.38±0.36 & 84.67±2.49 & 90.33±1.14
          & \textbf{93.97±0.23} & 89.88±0.27 & 87.79±1.84 & 91.85±0.63 \\

    \midrule
    \midrule
    \multirow{16}{*}{CIFAR-10}
      & \multirow{8}{*}{0.0}
      & \multirow{4}{*}{0.3}
        & MOON   
          & \textbf{61.29±1.22} & 57.89±0.82 & 48.09±0.39 & 59.17±1.51
          & \textbf{64.73±0.71} & 63.34±0.53 & 53.91±0.92 & 61.77±1.53 \\
      & & 
        & FedProx  
          & \textbf{61.40±1.09} & 57.96±0.68 & 45.94±0.76 & 59.18±1.37
          & \textbf{64.06±0.75} & 63.38±0.57 & 51.16±0.56 & 61.91±1.31 \\
      & & 
        & SCAFFOLD 
          & 68.44±0.73 & \textbf{68.93±1.91} & 53.67±0.64 & 65.15±1.10
          & 70.49±0.59 & \textbf{72.52±1.55} & 55.75±0.87 & 67.62±0.95 \\
      & & 
        & FedACG   
          & 61.88±2.33 & 57.88±0.75 & 44.12±1.29 & \textbf{64.33±1.53}
          & 64.05±2.01 & 63.33±0.40 & 45.60±1.78 & \textbf{65.57±1.50} \\
      \cmidrule(lr){3-12}
      & & \multirow{4}{*}{0.7}
        & MOON   
          & \textbf{66.28±0.51} & 61.39±0.73 & 49.03±1.55 & 63.70±0.46
          & \textbf{69.46±0.45} & 67.11±0.51 & 57.95±1.08 & 66.00±0.53 \\
      & & 
        & FedProx  
          & \textbf{66.30±0.64} & 61.54±0.71 & 46.50±1.17 & 64.01±0.67
          & \textbf{69.13±0.78} & 67.09±0.44 & 54.31±1.17 & 66.56±0.54 \\
      & & 
        & SCAFFOLD 
          & \textbf{72.95±0.68} & 71.91±0.81 & 57.37±0.84 & 69.29±0.86
          & \textbf{75.12±0.81} & 76.43±0.38 & 59.80±0.57 & 71.71±0.89 \\
      & & 
        & FedACG   
          & \textbf{70.35±0.80} & 61.46±0.74 & 47.91±1.35 & 68.17±0.50
          & \textbf{73.23±0.60} & 67.09±0.41 & 49.10±1.29 & 69.73±0.33 \\
    \cmidrule(lr){2-12}
      & \multirow{8}{*}{0.2}
      & \multirow{4}{*}{0.3}
        & MOON   
          & \textbf{54.39±1.21} & 51.64±0.93 & 51.54±1.02 & 53.65±0.92
          & \textbf{59.17±1.25} & 57.26±0.93 & 56.34±0.66 & 56.85±0.84 \\
      & & 
        & FedProx  
          & \textbf{55.26±1.80} & 51.74±0.88 & 50.40±0.99 & 53.29±0.80
          & \textbf{59.97±1.44} & 57.42±0.74 & 54.30±0.96 & 56.91±0.68 \\
      & & 
        & SCAFFOLD 
          & \textbf{67.35±0.65} & 63.67±3.23 & 52.91±1.16 & 62.49±0.88
          & \textbf{69.36±0.73} & 68.02±2.04 & 54.60±1.12 & 65.57±0.90 \\
      & & 
        & FedACG   
          & \textbf{59.33±2.61} & 51.48±0.88 & 48.80±1.07 & 55.33±0.99
          & \textbf{62.01±2.09} & 57.31±0.67 & 49.07±2.10 & 59.35±0.73 \\
      \cmidrule(lr){3-12}
      & & \multirow{4}{*}{0.7}
        & MOON   
          & \textbf{61.78±1.59} & 57.64±0.52 & 53.14±1.17 & 58.49±0.82
          & \textbf{66.10±1.53} & 64.19±0.45 & 59.72±0.74 & 60.73±0.73 \\
      & & 
        & FedProx  
          & \textbf{63.81±1.31} & 57.73±0.50 & 52.35±0.95 & 58.70±0.71
          & \textbf{66.36±1.13} & 64.24±0.44 & 58.05±0.72 & 61.36±0.66 \\
      & & 
        & SCAFFOLD 
          & \textbf{71.33±0.81} & 69.56±0.60 & 58.65±0.59 & 67.46±0.45
          & 73.51±0.74 & \textbf{74.41±1.15} & 61.05±0.85 & 70.39±0.38 \\
      & & 
        & FedACG   
          & \textbf{66.27±1.49} & 57.67±0.56 & 51.04±0.52 & 61.61±0.39
          & \textbf{69.04±1.37} & 64.20±0.48 & 53.47±0.73 & 64.66±0.37 \\
    \midrule
    \midrule
    \multirow{16}{*}{CIFAR-100}
      & \multirow{8}{*}{0.0}
      & \multirow{4}{*}{0.3}
        & MOON    
          & \textbf{38.74±0.52} & 32.35±0.52 & 37.40±0.61 & 36.09±0.58
          & 42.21±0.55 & 38.76±0.46 & \textbf{43.42±0.85} & 39.70±0.49 \\
      & & 
        & FedProx   
          & \textbf{38.38±0.46} & 32.21±0.46 & 35.87±1.13 & 35.41±0.56
          & 41.94±0.40 & 38.83±0.68 & \textbf{42.04±1.14} & 39.23±0.54 \\
      & & 
        & SCAFFOLD 
          & \textbf{45.03±0.34} & 46.10±0.80 & 44.21±0.65 & 42.32±0.56
          & 48.32±0.76 & 50.89±0.50 & \textbf{49.94±0.34} & 45.49±0.50 \\
      & & 
        & FedACG    
          & \textbf{47.07±0.37} & 32.30±0.19 & 26.86±0.57 & 41.86±0.80
          & \textbf{50.10±0.46} & 38.90±0.36 & 38.18±0.86 & 45.26±0.55 \\
    \cmidrule(lr){3-12}
      & & \multirow{4}{*}{0.7}
        & MOON    
          & \textbf{41.25±0.65} & 35.54±0.52 & 39.22±1.20 & 38.10±0.64
          & \textbf{44.99±0.89} & 41.91±0.49 & 44.65±1.87 & 41.43±0.54 \\
      & & 
        & FedProx   
          & \textbf{41.11±0.36} & 35.62±0.53 & 37.07±1.67 & 37.70±0.65
          & \textbf{44.59±0.54} & 42.02±0.49 & 42.71±1.32 & 41.60±0.72 \\
      & & 
        & SCAFFOLD 
          & \textbf{47.60±0.43} & 47.85±0.46 & 46.91±0.58 & 44.16±0.59
          & 50.71±0.70 & 51.37±0.44 & \textbf{52.50±0.81} & 47.79±0.42 \\
      & & 
        & FedACG    
          & \textbf{49.57±0.40} & 35.58±0.48 & 28.44±0.62 & 44.35±0.94
          & \textbf{52.30±0.78} & 42.13±0.41 & 39.32±1.57 & 47.63±0.51 \\
    \cmidrule(lr){2-12}
      & \multirow{8}{*}{0.2}
      & \multirow{4}{*}{0.3}
        & MOON    
          & 44.46±0.65 & 39.95±0.81 & \textbf{44.49±0.64} & 35.92±1.05
          & 47.35±0.75 & 45.93±0.27 & \textbf{49.99±0.81} & 39.61±0.47 \\
      & & 
        & FedProx   
          & 44.18±1.00 & 40.27±0.38 & \textbf{45.57±0.99} & 38.69±0.36
          & 47.40±0.90 & 45.92±0.28 & \textbf{50.22±0.34} & 42.32±0.47 \\
      & & 
        & SCAFFOLD 
          & \textbf{49.40±0.53} & 49.20±0.90 & 47.35±1.00 & 43.93±0.43
          & \textbf{52.22±0.38} & 51.88±0.55 & 52.19±1.35 & 47.31±0.41 \\
      & & 
        & FedACG    
          & \textbf{51.73±0.22} & 39.95±0.83 & 39.87±1.36 & 43.93±1.12
          & \textbf{53.64±0.19} & 45.62±0.70 & 46.71±0.92 & 48.03±1.22 \\
    \cmidrule(lr){3-12}
      & & \multirow{4}{*}{0.7}
        & MOON    
          & 47.27±0.41 & 43.14±0.31 & \textbf{44.98±0.29} & 37.88±0.66
          & 50.41±0.33 & 48.62±0.40 & \textbf{51.39±0.64} & 41.56±0.58 \\
      & & 
        & FedProx   
          & 46.87±0.33 & 43.05±0.51 & \textbf{47.72±0.19} & 40.45±0.30
          & 49.57±0.52 & 48.67±0.38 & \textbf{52.18±0.29} & 44.14±0.52 \\
      & & 
        & SCAFFOLD 
          & \textbf{51.28±0.74} & 51.47±0.46 & 51.13±0.39 & 47.94±0.47
          & 54.08±0.94 & 54.61±0.38 & \textbf{56.20±0.20} & 51.45±0.53 \\
      & & 
        & FedACG    
          & \textbf{53.52±0.14} & 42.91±0.44 & 41.16±1.25 & 47.39±0.54
          & \textbf{55.82±0.26} & 48.58±0.29 & 48.99±0.75 & 51.11±0.48 \\
    \bottomrule
  \end{tabular}}
  \end{small}
  \end{center}
\end{table*}

\section{Additional Results for Alternative Label Distributions} \label{appendix:alternative_distribution}
In addition to the label distribution we used in the main text, we also evaluate Algorithm \ref{alg:dynamic_initial_model_construction} using the following distribution. 
\begin{itemize}[leftmargin=0.35cm]
    \item \textbf{Two-Shard} \cite{mcmahan2017communication, hsu2019measuring, li2020federated, fallah2020personalized, karimireddy2020scaffold}\textbf{:} For datasets with 10 labels, each label is divided into two equally-sized shards, and each client receives two different label shards. For datasets with 100 labels, the labels are divided into 10 non-overlapping batches, and each batch is split into two shards, which are then assigned to two randomly selected clients. Each client ends up with data from two labels for 10-label datasets, or 20 labels for 100-label datasets .
    \item \textbf{Half:} Half the clients have one half of the labels, and the rest have the other half. Each client has all the labels from their assigned half (i.e., 5 labels for 10-label datasets or 50 labels for 100-label datasets), and the data is evenly distributed, meaning that the amount of data for each label is equally divided among the clients.
    \item \textbf{Partial-Overlap:} Two sets of labels are selected, each containing 60\% of the total labels, with a 20\% overlap between them. Each client in the first set has 6 labels for 10-label datasets (or 60 labels for 100-label datasets), and each client in the second set has a similar distribution. The overlapping labels are split between the two halves of clients, with half of the data for overlapping labels going to the first client set and the other half to the second set. The non-overlapping labels are assigned to the clients within each set, and the data corresponding to these labels is evenly distributed across the clients, meaning that each client receives an approximately equal share of the data for their assigned labels.
    \item \textbf{Distinct:} Each client is assigned a unique set of labels. For datasets with 10 labels, each client receives 1 unique label, while for datasets with 100 labels, each client receives 10 unique labels.
\end{itemize}

Figure \ref{fig:combined_subplots_4x2_with_legend} highlights selected scenarios and illustrates the advantages of the proposed algorithm for all FL algorithms during periods of pronounced performance drops or gains due to data distribution shifts. The algorithm prioritizes models trained on past distributions similar to the current one, enabling the initial model to adapt more quickly and consistently outperform baseline approaches. These results underscore the robustness and effectiveness of the proposed approach across diverse FL algorithms, datasets, models, and dynamic data distributions.

\begin{figure*}[t]
\vskip 0.2in
\begin{center}
\includegraphics[width=\linewidth]{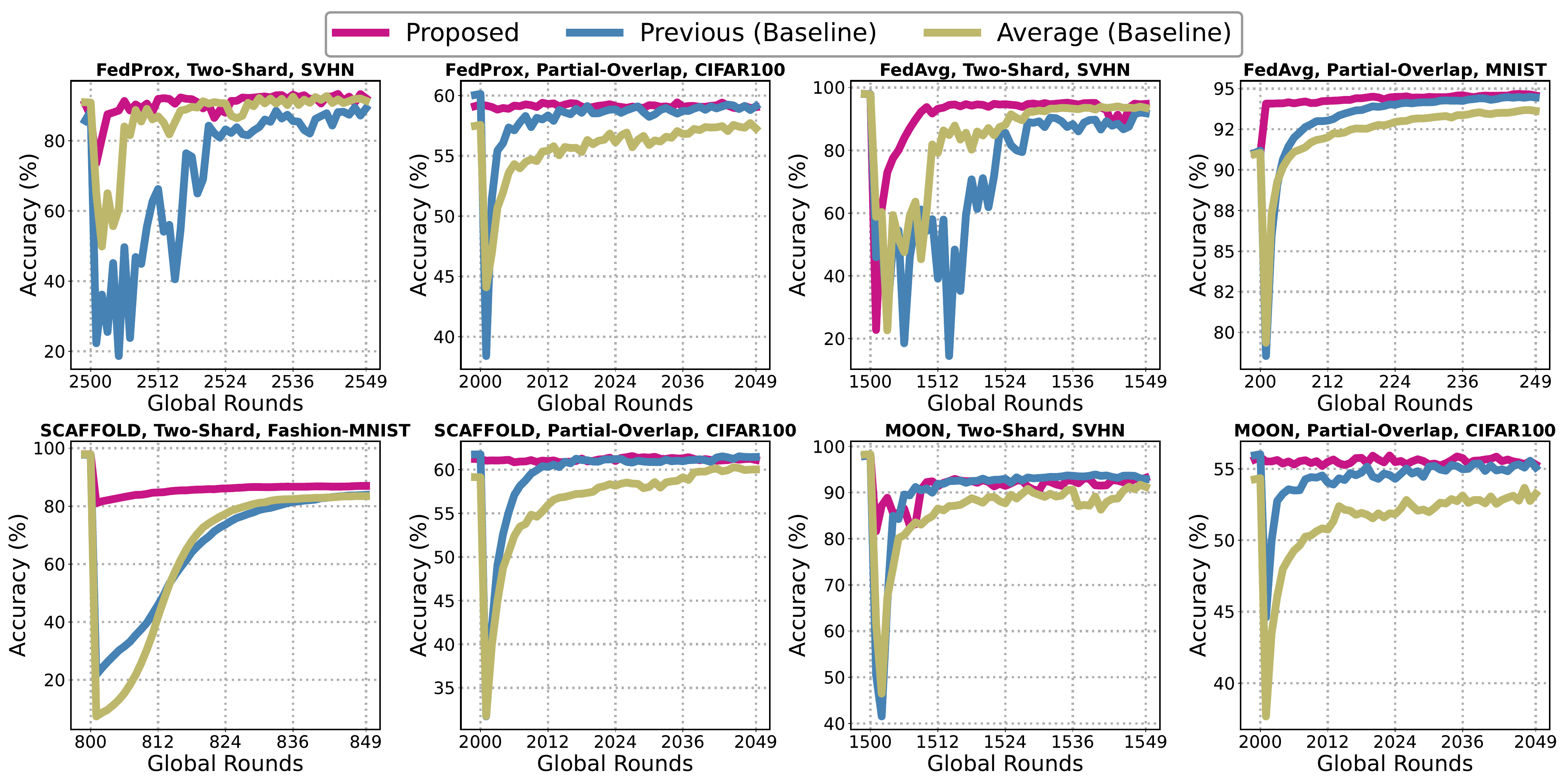}
\caption{Performance comparison of the proposed algorithm implemented with four FL algorithms under ``Two-Shard'' and ``Partial-Overlap'' label distributions across selected datasets. The results demonstrate the robustness of our proposed scheme to dynamic data distributions caused by client arrivals and departures.}
\label{fig:combined_subplots_4x2_with_legend}
\end{center}
\vskip -0.2in
\end{figure*}

Table \ref{table:first10} compares the average accuracy of the proposed algorithm across all FL algorithms during the first 10 global rounds after three data distribution shifts under the ``Two-Shard'' and ``Partial-Overlap'' label distributions. The results across datasets, label distributions, and models show that the proposed algorithm mitigates performance degradation from dynamic client arrivals and departures. Moreover, it accelerates performance recovery when the current data distribution aligns with a previously observed one.

\begin{table*}[ht!]
    \caption{Performance comparison of implementing Algorithm \ref{alg:dynamic_initial_model_construction} with FedProx, FedAvg, SCAFFOLD and MOON under different label distributions and datasets. Performance is measured across 3 transitions for each dataset.}
    \label{table:first10}
    \vskip 0.15in
    \begin{center}
    \resizebox{\textwidth}{!}{  
    \begin{tabular}{c c c >{\bfseries}c c c >{\bfseries}c c c >{\bfseries}c c c}
        \toprule
        \multirow{2}{*}{\centering \textbf{FL Algorithm}} & \multirow{2}{*}{\centering \textbf{Label Distribution}} & \multirow{2}{*}{\textbf{Dataset (Model)}} & \multicolumn{3}{c}{\textbf{1st Transition}} & \multicolumn{3}{c}{\textbf{2nd Transition}} & \multicolumn{3}{c}{\textbf{3rd Transition}} \\
        \cmidrule(lr){4-6} \cmidrule(lr){7-9} \cmidrule(lr){10-12}
        & & & \textnormal{Proposed} & \textnormal{Previous} & \textnormal{Average} & \textnormal{Proposed} & \textnormal{Previous} & \textnormal{Average} & \textnormal{Proposed} & \textnormal{Previous} & \textnormal{Average} \\
        \midrule
        \multirow{10}{*}{\textbf{FedProx}} 
        & \multirow{5}{*}{\textbf{Two-Shard}} 
        & MNIST (MLP) & 98.45 & 96.4 & 98.24 & 94.77 & 90.72 & 93.19 & 98.56 & 96.53 & 98.4 \\
        & & Fashion-MNIST (MLP) & 92.04 & 90.31 & 88.89 & 97.59 & 96.38 & 97.33 & 95.39 & 93.54 & 94.58 \\
        & & SVHN (CNN) & 86.52 & 40.13 & 72.7 & 97.44 & 84.53 & 95.08 & 95.8 & 55.16 & 83.08 \\
        & & CIFAR10 (ResNet18) & 41.64 & 35.32 & 39.68 & 46.26 & 35.62 & 44.62 & 82.58 & 58.34 & 78.3 \\
        & & CIFAR100 (ResNet18) & 58.46 & 56.68 & 51.07 & 49.28 & 46.52 & 47.7 & 52.73 & 48.35 & 51.49 \\
        \cmidrule{2-12}
        & \multirow{5}{*}{\textbf{Partial-Overlap}} 
        & MNIST (MLP) & 62.88 & 56.16 & 56.23 & 64.27 & 57.14 & 53.96 & 63.53 & 57.41 & 56.81 \\
        & & Fashion-MNIST (MLP) & 87.06 & 83.79 & 86.78 & 88.06 & 85.52 & 84.78 & 87.2 & 84.33 & 86.89 \\
        & & SVHN (CNN) & 92.4 & 76.09 & 91.72 & 93.64 & 68.67 & 92.35 & 92.51 & 71.79 & 92.02 \\
        & & CIFAR10 (ResNet18) & 81.5 & 77.05 & 77.31 & 86.99 & 83.88 & 84.97 & 81.69 & 79.84 & 77.62 \\
        & & CIFAR100 (ResNet18) & 59.09 & 54.94 & 52.23 & 59.22 & 55.09 & 56.13 & 59.34 & 56.11 & 53.34 \\
        \midrule
        \multirow{10}{*}{\textbf{FedAvg}} 
        & \multirow{5}{*}{\textbf{Two-Shard}} 
        & MNIST (MLP) & 98.45 & 96.4 & 98.24 & 94.77 & 90.72 & 93.19 & 98.56 & 96.53 & 98.4 \\
        & & Fashion-MNIST (MLP) & 92.04 & 90.31 & 88.89 & 97.59 & 96.38 & 97.33 & 95.39 & 93.54 & 94.58 \\
        & & SVHN (CNN) & 76.53 & 51.03 & 56.65 & 83.45 & 39.58 & 72.65 & 95.87 & 42.97 & 85.15 \\
        & & CIFAR10 (ResNet18) & 41.98 & 32.77 & 41.05 & 51.01 & 36.57 & 48.57 & 82.51 & 64.03 & 81.04 \\
        & & CIFAR100 (ResNet18) & 47.88 & 44.29 & 46.87 & 49.63 & 46.04 & 47.64 & 52.18 & 48.43 & 51.17 \\
        \cmidrule{2-12}
        & \multirow{5}{*}{\textbf{Partial-Overlap}} 
        & MNIST (MLP) & 93.81 & 89.65 & 89.33 & 91.11 & 87.46 & 90.35 & 94.16 & 90.1 & 90.67 \\
        & & Fashion-MNIST (MLP) & 87.58 & 85.06 & 83.04 & 86.77 & 83.88 & 86.49 & 87.64 & 85.59 & 84.16 \\
        & & SVHN (CNN) & 92.27 & 79.11 & 89.94 & 93.67 & 82.01 & 91.46 & 92.48 & 77.2 & 91.47 \\
        & & CIFAR10 (ResNet18) & 85.87 & 82.35 & 83.84 & 79.94 & 77.1 & 75.31 & 85.87 & 82.9 & 84.41 \\
        & & CIFAR100 (ResNet18) & 56.94 & 53.29 & 54.31 & 57.25 & 54.17 & 51.71 & 57.35 & 54.41 & 54.6 \\
        \midrule
        \multirow{10}{*}{\textbf{SCAFFOLD}} 
         & \multirow{5}{*}{\textbf{Two-Shard}} 
        & MNIST (MLP) & 97.97 & 51.69 & 96.73 & 92.27 & 42.16 & 44.86 & 98.32 & 54.78 & 97.26 \\
        & & Fashion-MNIST (MLP) & 84.24 & 36.59 & 22.94 & 96.89 & 51.82 & 96.18 & 92.29 & 52.41 & 45.45 \\
        & & SVHN (CNN) & 79.23 & 69.2 & 47.94 & 95.32 & 90.76 & 73.11 & 93.96 & 88.37 & 73.25 \\
        & & CIFAR10 (ResNet18) & 46.63 & 10.74 & 23.31 & 92.39 & 38.32 & 90.59 & 56.11 & 12.66 & 24.98 \\
        & & CIFAR100 (ResNet18) & 21.63 & 5.61 & 9.59 & 35.75 & 21.3 & 19.42 & 26.18 & 7.69 & 11.98 \\
        \cmidrule{2-12}
        & \multirow{5}{*}{\textbf{Partial-Overlap}} 
        & MNIST (MLP) & 85.72 & 61.6 & 71.39 & 89.66 & 77.03 & 53.64 & 87.1 & 69.45 & 72.93 \\
        & & Fashion-MNIST (MLP) & 86.67 & 62.41 & 74.68 & 87.54 & 63.48 & 68.97 & 87.64 & 65.2 & 74.91 \\
        & & SVHN (CNN) & 90.33 & 84.02 & 73.88 & 91.9 & 89.05 & 80.91 & 91.33 & 85.83 & 74.68 \\
        & & CIFAR10 (ResNet18) & 72.64 & 37.95 & 42.06 & 79.12 & 39.31 & 69.88 & 74.52 & 40.29 & 49.57 \\
        & & CIFAR100 (ResNet18) & 61.03 & 52.56 & 48.94 & 61.16 & 53.54 & 58.66 & 60.79 & 53.99 & 53.2 \\
        \midrule
        \multirow{10}{*}{\textbf{MOON}} 
        & \multirow{5}{*}{\textbf{Two-Shard}} 
        & MNIST (MLP) & 98.15 & 96.54 & 98.04 & 98.38 & 96.55 & 98.24 & 98.52 & 96.48 & 98.39 \\
        & & Fashion-MNIST (MLP) & 92.08 & 90.28 & 88.84 & 97.42 & 96.47 & 97.31 & 95.36 & 93.54 & 94.53 \\
        & & SVHN (CNN) & 87.0 & 78.61 & 75.72 & 69.51 & 64.57 & 54.88 & 93.49 & 90.2 & 89.87 \\
        & & CIFAR10 (ResNet18) & 44.03 & 40.53 & 39.42 & 51.31 & 43.46 & 41.82 & 85.0 & 71.45 & 73.54 \\
        & & CIFAR100 (ResNet18) & 57.94 & 56.05 & 44.18 & 45.8 & 43.81 & 44.71 & 62.06 & 60.56 & 53.54 \\
        \cmidrule{2-12}
       & \multirow{5}{*}{\textbf{Partial-Overlap}} 
        & MNIST (MLP) & 93.85 & 89.65 & 89.33 & 91.1 & 87.64 & 90.35 & 94.12 & 90.2 & 90.67 \\
        & & Fashion-MNIST (MLP) & 87.6 & 85.06 & 83.04 & 86.66 & 83.66 & 86.49 & 88.01 & 85.53 & 84.16 \\
        & & SVHN (CNN) & 92.96 & 90.41 & 88.39 & 91.74 & 87.97 & 90.77 & 93.18 & 91.25 & 90.25 \\
        & & CIFAR10 (ResNet18) & 85.53 & 83.59 & 84.67 & 79.28 & 77.57 & 76.04 & 79.37 & 78.44 & 76.11 \\
        & & CIFAR100 (ResNet18) & 55.52 & 52.58 & 47.77 & 55.27 & 52.22 & 52.7 & 55.72 & 53.49 & 49.01 \\
        \bottomrule
    \end{tabular}}
\end{center}
\vskip -0.1in
\end{table*}

\begin{table}[!ht]
    \caption{Performance comparison of FedProx, FedAvg, SCAFFOLD, and MOON under ``Half'' and ``Distinct'' label distributions across various datasets. Performance is evaluated over three transitions for each dataset.}
    \label{table:first10_2}
    \vskip 0.15in
    \begin{center}
    \resizebox{\textwidth}{!}{  
    \begin{tabular}{c c c >{\bfseries}c c c >{\bfseries}c c c >{\bfseries}c c c}
        \toprule
        \multirow{2}{*}{\centering \textbf{FL Algorithm}} & \multirow{2}{*}{\centering \textbf{Label Distribution}} & \multirow{2}{*}{\textbf{Dataset (Model)}} & \multicolumn{3}{c}{\textbf{1st Transition}} & \multicolumn{3}{c}{\textbf{2nd Transition}} & \multicolumn{3}{c}{\textbf{3rd Transition}} \\
        \cmidrule(lr){4-6} \cmidrule(lr){7-9} \cmidrule(lr){10-12}
        & & & Proposed & Previous & Average & Proposed & Previous & Average & Proposed & Previous & Average \\
        \midrule
        \multirow{8}{*}{\textbf{FedProx}} 
        & \multirow{5}{*}{\textbf{Half}} 
        & MNIST (MLP) & 96.02 & 90.12 & 92.45 & 96.29 & 90.56 & 93.59 & 93.25 & 88.17 & 92.45 \\
        & & Fashion-MNIST (MLP) & 90.97 & 86.02 & 90.17 & 86.62 & 85.06 & 84.2 & 91.29 & 86.67 & 90.26 \\
        & & SVHN (CNN) & 93.23 & 27.0 & 92.18 & 91.64 & 17.15 & 90.98 & 93.53 & 69.5 & 92.75 \\
        & & CIFAR10 (ResNet18) & 86.32 & 48.83 & 83.59 & 90.38 & 54.74 & 87.64 & 87.68 & 58.41 & 84.62 \\
        & & CIFAR100 (ResNet18) & 62.88 & 56.16 & 56.23 & 64.27 & 57.14 & 53.96 & 63.53 & 57.41 & 56.81 \\
        \cmidrule{2-12}
        & \multirow{3}{*}{\textbf{Distinct}} 
        & MNIST (MLP) & 98.05 & 92.75 & 97.71 & 95.98 & 88.77 & 93.96 & 94.11 & 85.66 & 90.54 \\
        & & Fashion-MNIST (MLP) & 95.5 & 92.4 & 95.0 & 95.8 & 90.59 & 94.56 & 94.02 & 87.36 & 90.73 \\
        & & SVHN (CNN) & 92.25 & 45.18 & 67.51 & 87.91 & 38.45 & 63.85 & 92.24 & 42.7 & 60.68 \\
        \midrule
        \multirow{8}{*}{\textbf{FedAvg}} 
        & \multirow{5}{*}{\textbf{Half}} 
        & MNIST (MLP) & 96.03 & 90.12 & 92.45 & 92.56 & 87.83 & 92.45 & 96.21 & 90.56 & 93.59 \\
        & & Fashion-MNIST (MLP) & 86.24 & 84.54 & 82.54 & 90.29 & 86.12 & 89.91 & 90.5 & 86.62 & 90.1 \\
        & & SVHN (CNN) & 91.63 & 71.49 & 88.78 & 93.54 & 51.98 & 91.07 & 91.98 & 42.34 & 90.55 \\
        & & CIFAR10 (ResNet18) & 90.19 & 70.15 & 88.08 & 86.23 & 77.41 & 81.67 & 90.03 & 73.55 & 89.25 \\
        & & CIFAR100 (ResNet18) & 57.41 & 52.95 & 54.37 & 57.09 & 53.7 & 51.07 & 57.48 & 53.38 & 54.44 \\
        \cmidrule{2-12}
       & \multirow{3}{*}{\textbf{Distinct}} 
        & MNIST (MLP) & 97.97 & 93.03 & 97.57 & 95.72 & 89.19 & 93.65 & 94.04 & 86.11 & 90.4 \\
        & & Fashion-MNIST (MLP) & 95.81 & 90.66 & 94.44 & 93.88 & 87.5 & 90.56 & 95.8 & 92.0 & 94.86 \\
        & & SVHN (CNN) & 94.99 & 49.88 & 80.62 & 91.9 & 38.94 & 58.53 & 87.21 & 32.71 & 66.47 \\
        \midrule
        \multirow{7}{*}{\textbf{SCAFFOLD}} 
         & \multirow{5}{*}{\textbf{Half}} 
        & MNIST (MLP) & 85.53 & 29.8 & 82.93 & 92.58 & 57.81 & 30.18 & 89.08 & 51.43 & 84.39 \\
        & & Fashion-MNIST (MLP) & 77.18 & 36.71 & 28.55 & 83.35 & 65.31 & 77.91 & 79.51 & 40.82 & 42.19 \\
        & & SVHN (CNN) & 91.66 & 86.87 & 76.06 & 89.86 & 84.79 & 80.25 & 93.04 & 87.31 & 87.18 \\
        & & CIFAR10 (ResNet18) & 78.63 & 17.35 & 34.76 & 84.62 & 19.59 & 79.32 & 80.13 & 20.86 & 49.7 \\
        & & CIFAR100 (ResNet18) & 28.92 & 9.97 & 18.74 & 31.11 & 13.3 & 6.44 & 34.23 & 13.84 & 21.62 \\
        \cmidrule{2-12}
        & \multirow{2}{*}{\textbf{Distinct}} 
        & MNIST (MLP) & 96.95 & 50.5 & 95.35 & 93.08 & 22.24 & 60.96 & 91.36 & 21.06 & 22.78 \\
        & & Fashion-MNIST (MLP) & 94.08 & 32.05 & 91.66 & 93.28 & 13.81 & 53.5 & 91.72 & 29.1 & 33.07 \\
        \midrule
        \multirow{7}{*}{\textbf{MOON}} 
        & \multirow{5}{*}{\textbf{Half}} 
        & MNIST (MLP) & 96.06 & 90.16 & 92.45 & 92.64 & 87.78 & 92.45 & 96.29 & 90.46 & 93.59 \\
        & & Fashion-MNIST (MLP) & 86.12 & 84.43 & 82.54 & 90.33 & 85.79 & 89.91 & 86.31 & 84.96 & 83.75 \\
        & & SVHN (CNN) & 93.64 & 89.35 & 89.95 & 92.11 & 87.3 & 91.18 & 93.63 & 90.54 & 91.37 \\
        & & CIFAR10 (ResNet18) & 89.0 & 85.77 & 88.26 & 84.89 & 81.08 & 81.53 & 89.16 & 86.67 & 88.51 \\
        & & CIFAR100 (ResNet18) & 54.9 & 52.27 & 53.8 & 55.95 & 52.75 & 48.99 & 55.05 & 52.63 & 54.07 \\
        \cmidrule{2-12}
        & \multirow{2}{*}{\textbf{Distinct}} 
        & MNIST (MLP) & 95.67 & 88.67 & 93.56 & 94.07 & 85.7 & 90.2 & 95.63 & 90.65 & 94.32 \\
        & & Fashion-MNIST (MLP) & 95.76 & 90.14 & 94.4 & 93.86 & 86.83 & 90.06 & 95.66 & 91.76 & 94.85 \\
        \bottomrule
    \end{tabular}}
\end{center}
\vskip -0.1in
\end{table}

Table \ref{table:first10_2} presents a comparative analysis of the average accuracy achieved by the proposed algorithm across all federated learning (FL) algorithms during the first 10 global rounds following three shifts in data distribution under the ``Half'' and ``Distinct'' label distributions.

Table \ref{table:tinyimagenet_fedavg} presents the results of applying Table \ref{alg:dynamic_initial_model_construction} with FedAvg for 20 clients on the TinyImageNet dataset \cite{le2015tiny}. The baseline method used is ``Average.’’ These results highlight the effectiveness of Table \ref{alg:dynamic_initial_model_construction} in handling scenarios with a larger client population and a more complex dataset using a more advanced model ResNet34. 

\begin{table}[t]
    \caption{Performance of Algorithm \ref{alg:dynamic_initial_model_construction} with FedAvg under ``Half'' and ``Partial-Overlap'' label distributions for the TinyImageNet dataset with 20 clients. Performance is measured across three transitions for each dataset.}
    \label{table:tinyimagenet_fedavg}
    \begin{center}
    \resizebox{\textwidth}{!}{  
    \begin{tabular}{c c c c c c c c}
        \toprule
        \multirow{2}{*}{\centering \textbf{Label Distribution}} & \multirow{2}{*}{\textbf{Dataset (Model)}} & \multicolumn{2}{c}{\textbf{1st Transition}} & \multicolumn{2}{c}{\textbf{2nd Transition}} & \multicolumn{2}{c}{\textbf{3rd Transition}} \\
        \cmidrule(lr){3-4} \cmidrule(lr){5-6} \cmidrule(lr){7-8}
        & & \textbf{Proposed} & Previous & \textbf{Proposed} & Previous & \textbf{Proposed} & Previous \\
        \midrule
        Half & TinyImageNet (ResNet34) & \bfseries 80.72 & 72.81 & \bfseries 77.77 & 72.24 & \bfseries 80.43 & 73.42 \\
        Partial-Overlap & TinyImageNet (ResNet34) & \bfseries 92.27 & 79.11 & \bfseries 93.67 & 82.01 & \bfseries 92.48 & 77.2 \\
        \bottomrule
    \end{tabular}
    }
    \end{center}
\end{table}

Table \ref{table:100clients_first10_partial_overlap} shows the results for 100 clients for Algorithm \ref{alg:dynamic_initial_model_construction} integrated with FedProx, FedAvg, SCAFFOLD, and MOON on the MNIST and Fashion-MNIST datasets, using 100 clients under the Partial-Overlap label distribution. Performance is evaluated over three transitions for each dataset.

\begin{table}[t]
    \caption{Performance comparison of Algorithm \ref{alg:dynamic_initial_model_construction} integrated with FedProx, FedAvg, SCAFFOLD, and MOON on the MNIST and Fashion-MNIST datasets, utilizing 100 clients for FedProx, FedAvg, and MOON, and 50 clients for SCAFFOLD under the Partial-Overlap label distribution. Performance is evaluated across three transitions for each dataset.}
    \label{table:100clients_first10_partial_overlap}
    \begin{center}
    \resizebox{\textwidth}{!}{  
    \begin{tabular}{c c >{\bfseries}c c c >{\bfseries}c c c >{\bfseries}c c c}
        \toprule
        \multirow{2}{*}{\centering \textbf{FL Algorithm}} & \multirow{2}{*}{\textbf{Dataset (Model)}} & \multicolumn{3}{c}{\textbf{1st Transition}} & \multicolumn{3}{c}{\textbf{2nd Transition}} & \multicolumn{3}{c}{\textbf{3rd Transition}} \\
        \cmidrule(lr){3-5} \cmidrule(lr){6-8} \cmidrule(lr){9-11}
        & & Proposed & Previous & Average & Proposed & Previous & Average & Proposed & Previous & Average \\
        \midrule
        \multirow{2}{*}{\textbf{FedProx}} 
        & MNIST (MLP) & 94.09 & 89.53 & 89.35 & 91.36 & 87.4 & 90.45 & 94.33 & 89.92 & 90.78 \\
        & Fashion-MNIST (MLP) & 87.88 & 85.13 & 83.37 & 87.14 & 83.86 & 86.76 & 88.08 & 85.57 & 84.65 \\ 
        \midrule
        \multirow{2}{*}{\textbf{FedAvg}} 
        & MNIST (MLP) & 90.86 & 86.15 & 90.25 & 94.09 & 89.53 & 89.35 & 91.36 & 87.4 & 90.45 \\
        & Fashion-MNIST (MLP) & 86.55 & 82.78 & 86.4 & 87.88 & 85.13 & 83.37 & 87.14 & 83.86 & 86.76 \\
        \midrule
        \textbf{SCAFFOLD} 
        & MNIST (MLP) & 64.82 & 33.15 & 52.67 & 82.67 & 62.47 & 30.18 & 74.04 & 50.77 & 59.66 \\
        \midrule
        \multirow{2}{*}{\textbf{MOON}} 
        & MNIST (MLP) & 91.36 & 87.4 & 90.45 & 94.33 & 89.92 & 90.78 & 91.55 & 87.87 & 90.55 \\
        & Fashion-MNIST (MLP) & 87.14 & 83.86 & 86.75 & 88.07 & 85.57 & 84.65 & 87.31 & 84.34 & 86.9 \\
        \bottomrule
    \end{tabular}}
\end{center}
\end{table}

Figure \ref{fig:combined_subplots_4x4_with_legend} highlights the performance of Algorithm \ref{alg:dynamic_initial_model_construction}  in more selected scenarios and illustrates the advantages of the proposed algorithm for all FL algorithms during periods of pronounced performance drops or gains due to data distribution shifts. The algorithm prioritizes models trained on past distributions that share similarities with the current one, enabling the initial model to adapt more quickly to the new distribution and consistently outperforming baseline approaches. These results underscore the robustness and effectiveness of the proposed approach across diverse FL algorithms, datasets, models, and dynamic data distributions.

\begin{figure}[ht]
\vskip 0.2in
\begin{center}
\includegraphics[width=\linewidth]{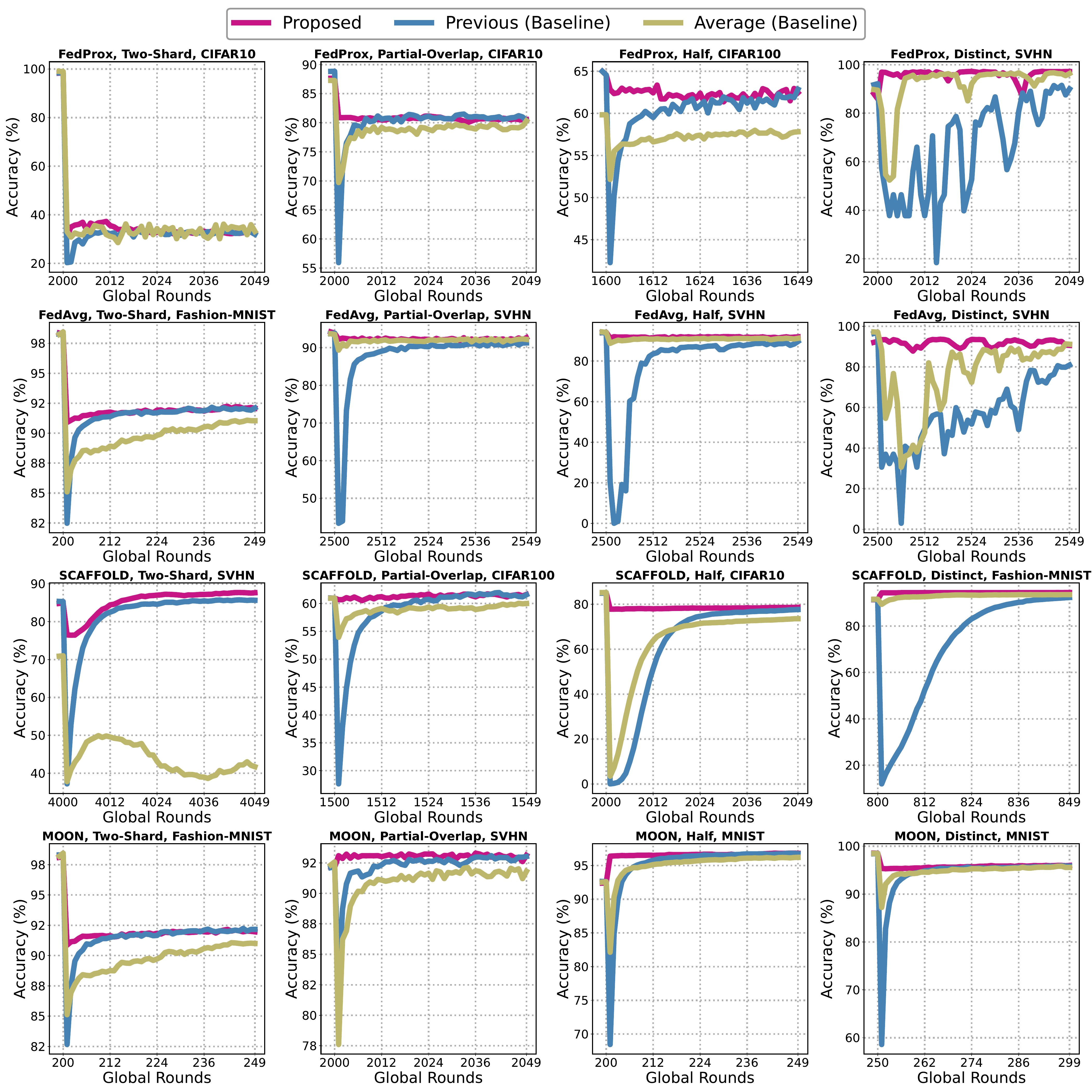}
\caption{Performance comparison of the proposed algorithm implemented with four FL algorithms under all label distributions across more selected datasets. The results demonstrate the robustness of our proposed scheme to dynamic data distributions caused by client arrivals and departures.}
\label{fig:combined_subplots_4x4_with_legend}
\end{center}
\vskip -0.2in
\end{figure}

\fi

\end{document}